\documentclass[10pt,journal,compsoc]{IEEEtran}
\usepackage{times}
\usepackage{epsfig}
\usepackage{graphicx}
\usepackage{amsmath}
\usepackage{amssymb}

\usepackage{caption}
\usepackage{subcaption}
\usepackage{epsfig}
\usepackage{amsmath}
\usepackage{amssymb}
\usepackage{booktabs}
\usepackage{rotating}
\usepackage{algorithm} 
\usepackage{algpseudocode}
\usepackage{multirow}
\usepackage{color}

\usepackage{enumitem}
\usepackage{enumerate}

\newif\ifshowcomments
\showcommentstrue

\ifshowcomments
\newcommand{\TODO}[1]{{\color{red}{[TODO: #1]}}}
\newcommand{\revised}[1]{{\color[rgb]{0.2,0.7,0.2}{#1}}}
\newcommand{\lzhu}[1]{{\color[rgb]{0.7,0.7,0}{#1}}}
\newcommand{\phil}[1]{{\color[rgb]{0.9,0.1,0.1}{#1}}}
\newcommand{\xwhu}[1]{{\color[rgb]{0.9,0.2,0.9}{#1}}}

\else
\newcommand{\TODO}[1]{}
\newcommand{\revised}[1]{}
\newcommand{\lzhu}[1]{}
\newcommand{\phil}[1]{}
\fi

\ifCLASSOPTIONcompsoc
  \usepackage[nocompress]{cite}
\else
  \usepackage{cite}
\fi
\ifCLASSINFOpdf
\else
\fi
\begin{document}
%
\title{Direction-aware Spatial Context Features\\ for Shadow Detection and Removal}
%
%
%
%

\author{Xiaowei~Hu,~\IEEEmembership{Student Member,~IEEE},
	Chi-Wing~Fu,~\IEEEmembership{Member,~IEEE,} \\
	Lei~Zhu,
	Jing~Qin,~\IEEEmembership{Member,~IEEE,}
	and~Pheng-Ann~Heng,~\IEEEmembership{Senior Member,~IEEE}
    \IEEEcompsocitemizethanks{
	\IEEEcompsocthanksitem X. Hu and L. Zhu are with the Department of Computer Science and Engineering, The Chinese University of Hong Kong.
	\IEEEcompsocthanksitem J. Qin is with Centre for Smart Health, School of Nursing, The Hong Kong Polytechnic University.
	\IEEEcompsocthanksitem C.-W. Fu and P.-A. Heng are with the Department of Computer Science and Engineering, The Chinese University of Hong Kong and Guangdong Provincial Key Laboratory of Computer Vision and Virtual Reality Technology, Shenzhen Institutes of Advanced Technology, Chinese Academy of Sciences, China.
	
	\IEEEcompsocthanksitem A preliminary version of this work was accepted for presentation in CVPR 2018~\cite{hu2018direction}. The source code is publicly available at https://xw-hu.github.io/.
\IEEEcompsocthanksitem Chi-Wing Fu and Lei Zhu are the co-corresponding authors of this work.
}
}

\markboth{IEEE Transactions on Pattern Analysis and Machine Intelligence}%
{Shell \MakeLowercase{\textit{et al.}}: Bare Demo of IEEEtran.cls for Computer Society Journals}
\IEEEtitleabstractindextext{%
\begin{abstract}
Shadow detection and shadow removal are fundamental and challenging tasks, requiring an understanding of the global image semantics.
This paper presents a novel deep neural network design for shadow detection and removal by analyzing the spatial image context in a direction-aware manner.
To achieve this, we first formulate the direction-aware attention mechanism in a spatial recurrent neural network (RNN) by introducing attention weights when aggregating spatial context features in the RNN.
By learning these weights through training, we can recover direction-aware spatial context (DSC) for detecting and removing shadows.
This design is developed into the DSC module and embedded in a convolutional neural network (CNN) to learn the DSC features at different levels.
Moreover, we design a weighted cross entropy loss to make effective the training for shadow detection and further adopt the network for shadow removal by using a Euclidean loss function and formulating a color transfer function to address the color and luminosity inconsistencies in the training pairs.
We employed two shadow detection benchmark datasets and two shadow removal benchmark datasets, and performed various experiments to evaluate our method.
Experimental results show that our method performs favorably against the state-of-the-art methods for both shadow detection and shadow removal.
%
\end{abstract}

\begin{IEEEkeywords}
Shadow detection, shadow removal, spatial context features, deep neural network.
\end{IEEEkeywords}}

\maketitle

\IEEEdisplaynontitleabstractindextext

%
\IEEEpeerreviewmaketitle


\IEEEraisesectionheading{\section{Introduction}
\label{sec::introduction}}

\IEEEPARstart{S}{hadow} is a monocular visual cue for perceiving depth and geometry.
On the one hand, knowing the shadow location allows us to obtain the lighting direction~\cite{lalonde2009estimating}, camera parameters~\cite{junejo2008estimating}, and scene geometry~\cite{okabe2009attached,karsch2011rendering}.
On the other hand, the presence of shadows could, however, deteriorate the performance of many computer vision tasks, e.g., object detection and tracking~\cite{cucchiara2003detecting,nadimi2004physical}.
Hence, shadow detection and shadow removal have long been fundamental problems in computer vision research.


Early approaches detect and remove shadows by developing physical models to analyze the statistics of color and illumination~\cite{salvador2004cast,panagopoulos2011illumination,tian2016new,finlayson2006removal,finlayson2002removing,liu2008texture,nadimi2004physical,finlayson2009entropy,wu2007natural}.
%
However, these approaches are built on assumptions that may not be correct in complex cases~\cite{khan2016automatic}.
%
To distill the knowledge from real images, the data-driven approach learns and understands shadows by using hand-crafted features~\cite{huang2011characterizes,lalonde2010detecting,zhu2010learning,guo2013paired,gryka2015learning}, or by learning the features using deep neural networks~\cite{khan2016automatic,khan2014automatic,vicente2016large,wang2018stacked}.
While the state-of-the-art methods are already able to detect shadows with an accuracy of 87\% to 90\%~\cite{vicente2016large,zhu2010learning} and recover most shadow regions~\cite{wang2018stacked}, they may misunderstand black objects as shadows and produce various artifacts, since they fail to analyze and understand the global image semantics; see Sections~\ref{sec:experiments} \&~\ref{sec:experiments_removal} for quantitative and qualitative comparison results.


\begin{figure}[!t]
	\centering
	\includegraphics[width=0.8\linewidth]{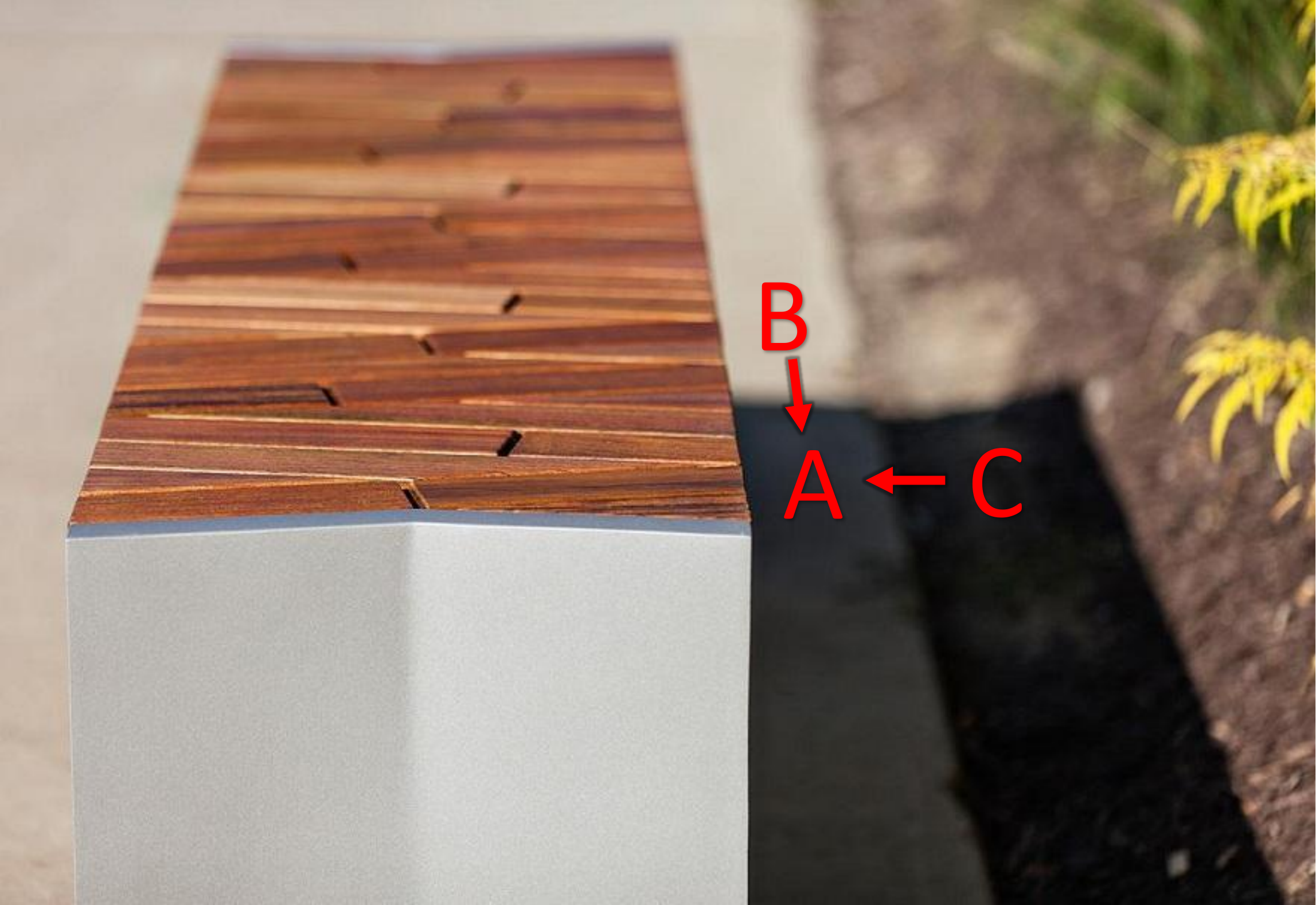}
	\vspace*{-1mm}
	\caption{In this example image, region B would give a stronger indication that A is a shadow compared to region C.
	This motivates us to analyze the global image context in a direction-aware manner for detecting and removing shadows.}
	\label{fig:motivation}
	\vspace*{-2mm}
\end{figure}

To recognize and remove shadows requires exploiting the global image semantics, as shown very recently by V. Nguyen~\emph{et al.}~\cite{nguyen2017shadow} for shadow detection and L. Qu~\emph{et al.}~\cite{qu2017deshadownet} for shadow removal.
%
To this end, we propose to analyze the image context in a {\em direction-aware\/} manner, since shadows are typically recognized by comparing with the surroundings.
Taking region A in Figure~\ref{fig:motivation} as an example, comparing it with regions B and C, region B would give a stronger indication (than region C) that A is a shadow.
Hence, spatial context in different directions would give different amount of contributions in suggesting the presence of shadows.

To capture the differences between image/spatial context in various directions,
we design the {\em direction-aware spatial context} (DSC) module, or {\em DSC module\/} for short, in a deep neural network, where we first aggregate the global image context by adopting a spatial recurrent neural network (RNN) in four principal directions, and then formulate a direction-aware attention mechanism in the RNN to learn the attention weights for each direction.
Hence, we can obtain the spatial context in a direction-aware manner.
Further, we embed multiple copies of DSC module in the convolutional neural network to learn the DSC features in different layers (scales), and combine these features with the convolutional features to predict a shadow mask for each layer.
%
%
After that, we fuse the predictions from different layers into the final shadow detection result with the weighted cross entropy loss to optimize the network.

To further adopt the network for shadow removal, we replace the shadow masks with the shadow-free images as the ground truth, and use a Euclidean loss between the training pairs (images with and without shadows) to predict the shadow-free images.
In addition, due to variations in camera exposure and environmental lighting, the training pairs may have inconsistent colors and luminosity; such inconsistencies can be observed in existing shadow removal datasets such as SRD~\cite{qu2017deshadownet} and ISTD~\cite{wang2018stacked}.
To this end, we formulate a transfer function to adjust the shadow-free ground truth images and use the adjusted ground truth images to train the network, so that our shadow removal network can produce shadow-free images that are more faithful to the input test images.

We summarize the major contributions of this work below:
\begin{itemize}[]

\item
First, we design a novel attention mechanism in a spatial RNN and construct the DSC module to learn the spatial context in a direction-aware manner.

\item
Second, we develop a new network for shadow detection by adopting multiple DSC modules to learn the direction-aware spatial context in different layers and by designing a weighted cross entropy loss to balance the detection accuracy in shadow and non-shadow regions.

\item
Third, we further adopt the network for shadow removal by formulating a Euclidean loss and training the network with color-compensated shadow-free images, which are produced through a color transfer function.


\item
Lastly, we evaluate our method on several benchmark datasets on shadow detection and shadow removal, and compare it with the state-of-the-art methods.
Experimental results show that our network performs favorably against the previous methods for both tasks; see Sections~\ref{sec:experiments} \& \ref{sec:experiments_removal} for quantitative and qualitative comparison results.

\end{itemize}

\section{Related Work}
\label{sec:related_work}

In this section, we focus on discussing works on single-image shadow detection and removal.


\vspace*{3mm}
\noindent
{\bf Shadow detection.} \
Traditionally, single-image shadow detection methods~\cite{salvador2004cast,panagopoulos2011illumination,tian2016new} exploit physical models of illumination and color.
This approach, however, tends to produce satisfactory results only for wide dynamic range images~\cite{lalonde2010detecting,nguyen2017shadow}.
Another approach learns shadow properties using hand-crafted features based on annotated shadow images.
It first describes image regions by feature descriptors then classifies the regions into shadow and non-shadow regions.
Features like color~\cite{lalonde2010detecting,vicente2015leave,guo2011single,vicente2018leave}, texture~\cite{zhu2010learning,vicente2015leave,guo2011single,vicente2018leave}, edge~\cite{lalonde2010detecting,zhu2010learning,huang2011characterizes}, and T-junction~\cite{lalonde2010detecting} are commonly used for shadow detection followed by classifiers like decision tree~\cite{lalonde2010detecting,zhu2010learning} and SVM~\cite{guo2011single,huang2011characterizes,vicente2015leave,vicente2018leave}.
However, since hand-crafted features have limited capability in describing shadows, this approach often fails for complex cases.


\begin{figure*} [htbp]
	\centering
	\includegraphics[width=0.92\linewidth]{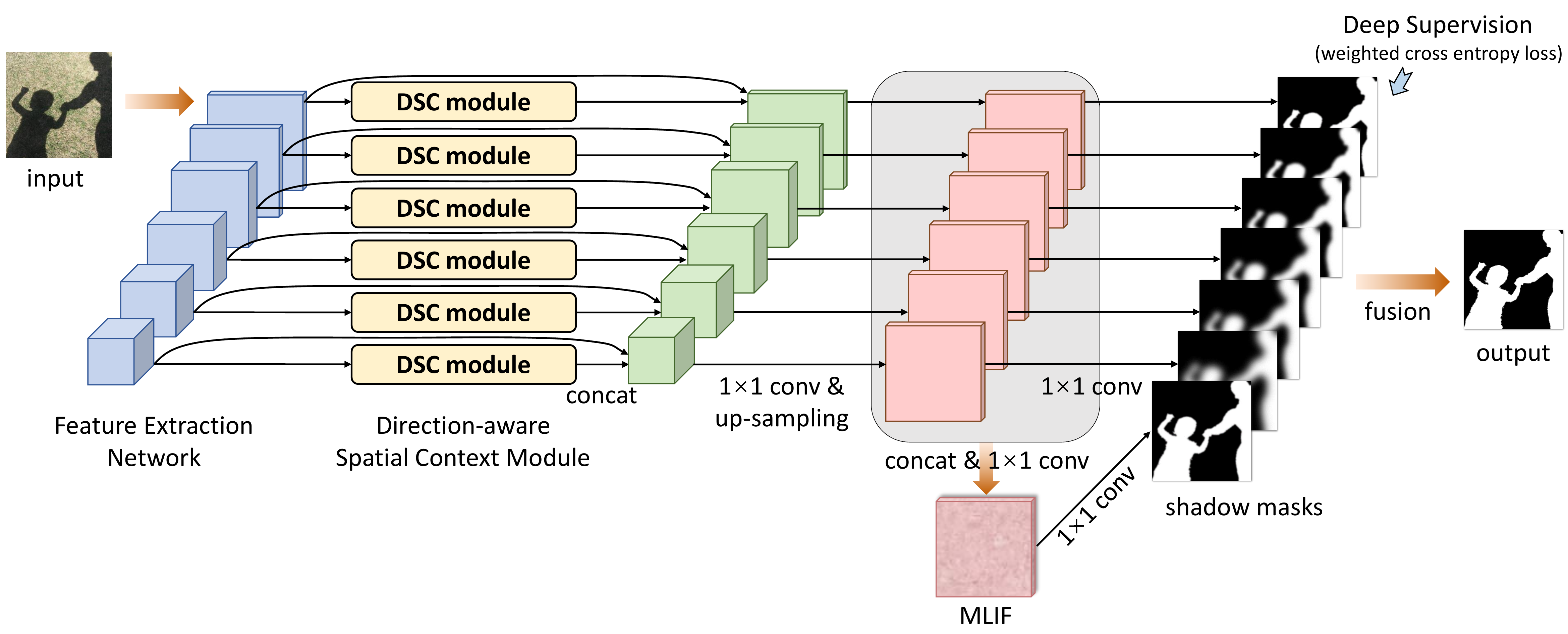}
	\caption{The schematic illustration of the overall shadow detection network:
(i) we extract features in different scales over the CNN layers from the input image;
(ii) we embed a DSC module (see Figure~\ref{fig:context}) to generate direction-aware spatial context (DSC) features for each layer;
(iii) we concatenate the DSC features with convolutional features at each layer and upsample the concatenated feature maps to the size of the input image;
%
%
(iv) we combine the upsampled feature maps into the multi-level integrated features (MLIF), predict a shadow mask based on the features for each layer using the deep supervision mechanism in~\cite{lee2015deeply}, and fuse the resulting shadow masks;
and
(v) in the testing process, we compute the mean shadow mask over the MLIF layer and the fusion layer, and use the conditional random field~\cite{krahenbuhl2011efficient} to further refine the detection result.
See Section~\ref{subsec:3.3} for how we adopt this network for shadow removal.}
	\label{fig:arc}
\end{figure*}


Convolutional neural networks (CNN) have been shown to be powerful tools for learning features to detect shadows, with results outperforming previous approaches, as when large data is available.
Khan et al.~\cite{khan2014automatic} used multiple CNNs to learn features in superpixels and along object boundaries, and fed the output features to a conditional random field to locate shadows.
Shen et al.~\cite{shen2015shadow} presented a deep structured shadow edge detector and employed structured labels to improve the local consistency of the predicted shadow map.
Vicente et al.~\cite{vicente2016large} trained a stacked-CNN using a large dataset with noisy annotations.
They minimized the sum of squared leave-one-out errors for image clusters to recover the annotations, and trained two CNNs to detect shadows.

Recently, Hosseinzadeh et al.~\cite{hosseinzadeh2017fast} detected shadows using a patch-level CNN and a shadow prior map computed from hand-crafted features.
Nguyen et al.~\cite{nguyen2017shadow} designed scGAN with a sensitivity parameter to adjust weights in loss functions.
Though the shadow detection accuracy keeps improving on the benchmarks~\cite{zhu2010learning,vicente2016large}, existing methods may still misrecognize black objects as shadows and miss unobvious shadows.
The
recent work by Nguyen et al.~\cite{nguyen2017shadow} emphasized the importance of reasoning global semantics for detecting shadows.
%
Beyond this work, we further consider the directional variance when analyzing the spatial context.
Experimental results show that our method further outperforms~\cite{nguyen2017shadow} on the benchmarks; see Section~\ref{sec:experiments_removal}.




\vspace*{3mm}
\noindent
{\bf Shadow removal.} \
Early works remove shadows by developing physical models deduced from the process of image formation~\cite{finlayson2006removal,finlayson2002removing,liu2008texture,nadimi2004physical,finlayson2009entropy,wu2007natural,baba2004shadow}.
However, these approaches are not effective to describe the shadows in complex real scenes~\cite{khan2016automatic}.
Afterwards, statistical learning methods were developed for shadow removal based on hand-crafted features (e.g., intensity~\cite{gong2014interactive,gryka2015learning,guo2013paired}, color~\cite{guo2013paired}, texture~\cite{guo2013paired}, gradient~\cite{gryka2015learning}), which lack high-level semantic knowledge for discovering shadows.



Lately, features learned by the convolutional neural networks (CNNs) are widely used for shadow removal.
Khan et al.~\cite{khan2016automatic} applied multiple CNNs to learn to detect shadows, and formulated a Bayesian model to extract shadow matte and remove shadows in a single image.
Very recently, Qu et al.~\cite{qu2017deshadownet} presented an architecture to remove shadows in an end-to-end manner.
The method applied three embedding networks (global localization network, semantic modeling network, and appearance modeling network) to extract features in three levels.
Wang et al.~\cite{wang2018stacked} designed two conditional generative adversarial networks in one framework to detect and remove shadows simultaneously.

However, shadow removal is a challenging task.
As pointed out by Qu et al.~\cite{qu2017deshadownet} and Wang et al.~\cite{wang2018stacked}, shadow removal needs a global view of the image to achieve global consistency in the prediction results.
However, existing methods may still fail to reasonably restore the shadow regions and mistakenly change the colors in the non-shadow regions.
In this work, we analyze the global spatial context in a direction-aware manner and formulate a color compensation mechanism to adjust the pixel colors and luminosity by considering the non-shadow regions between the training pairs in the current benchmark datasets~\cite{qu2017deshadownet,wang2018stacked}.
Experimental results show the effectiveness of our method over the state-of-the-art methods, both qualitatively and quantitatively.


\vspace*{3mm}
\noindent
{\bf Intrinsic images.} \
Another relevant topic is the intrinsic image decomposition, which aims to take away the illumination from the input and produce an image that contains only the reflectance.
To resolve the problem, early methods~\cite{shen2008intrinsic,bousseau2009user,shen2011intrinsic,zhao2012closed,chen2013simple,bell2014intrinsic} use various hand-crafted features to formulate constraints for extracting valid solutions; please refer to~\cite{barron2015shape} for a detailed review.
With deep neural networks, techniques have shifted towards data-driven methods with CNNs.
Narihira et al.~\cite{narihira2015direct} presented a method named as ``Direct intrinsics,'' which was an early attempt that employs a multi-layer CNN to directly transform an image into shading and reflectance.
Later, Kim et al.~\cite{kim2016unified} predicted the depth and other intrinsic components using a joint CNN with shared intermediate layers.
More recently, Lettry et al.~\cite{lettry2018darn} presented the DARN network that employs a discriminator network and an adversarial training scheme to enhance the performance of the generator network, while
Cheng et al.~\cite{cheng2018intrinsic} designed a scale-space network to generate the intrinsic images.
%


\vspace*{3mm}   
\noindent
This work extends our earlier work~\cite{hu2018direction} in three aspects.
First, we adopt the shadow detection network with the DSC features to remove shadows by re-designing the outputs and formulating different loss functions to train the network.
Second, we show that the pixel colors and luminosity in training pairs (shadow images and shadow-free images) of existing shadow removal datasets may not be consistent.
To this end, we formulate a color compensation mechanism and use a transfer function to make consistent the pixel colors in ground truth images before training our shadow removal network.
Third, we perform more experiments to evaluate the design of our networks for shadow detection and for shadow removal by considering more benchmark datasets and measuring the time performance, and show how our shadow removal network outperforms the best existing methods for shadow removal.



\section{Methodology}
\label{sec:method}

\begin{figure*} [tp]
	\centering
	\includegraphics[width=0.98\linewidth]{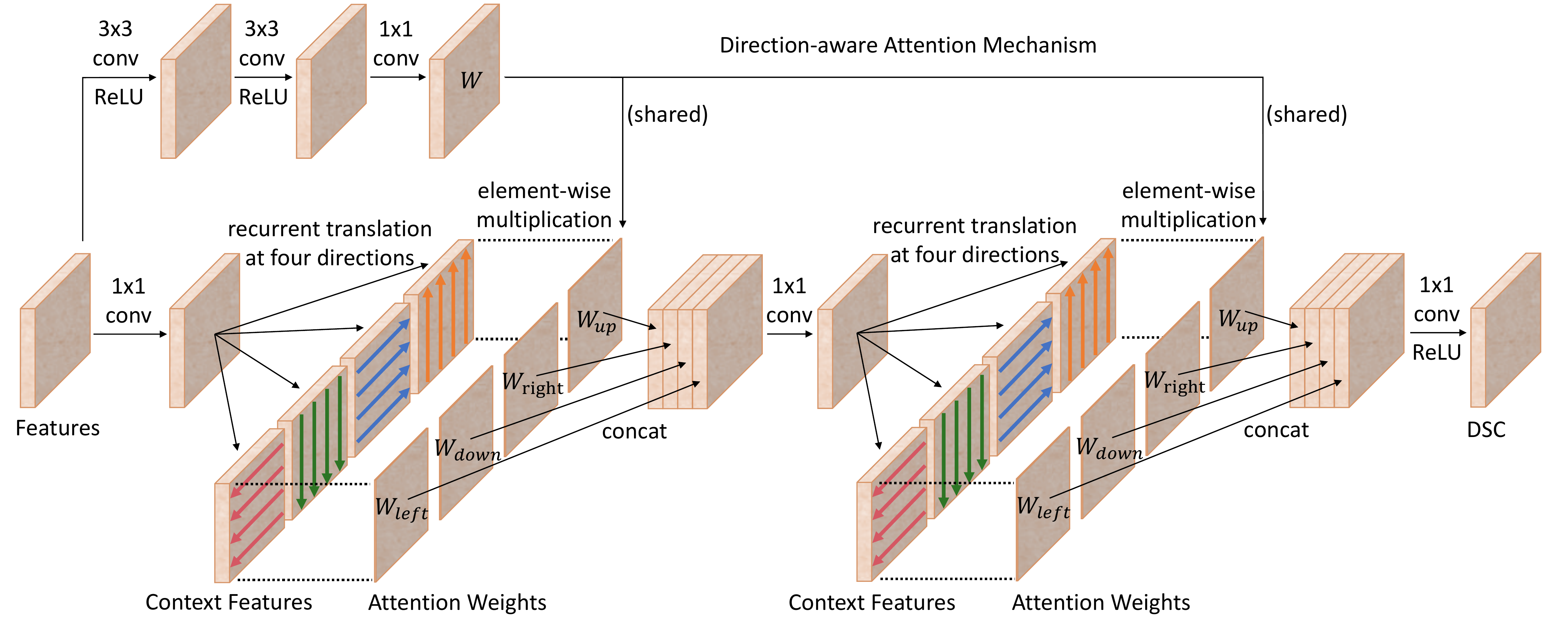}
	\caption{The schematic illustration of the {\em direction-aware spatial context module\/} ({\em DSC module\/}).
	We compute the direction-aware spatial context by adopting a spatial RNN to aggregate spatial context in four principal directions with two rounds of recurrent translations, and formulate the attention mechanism to generate maps of attention weights to combine the context features for different directions.
	We use the same set of weights in both rounds of recurrent translations.
	Best viewed in color.}
	\label{fig:context}
\end{figure*}

In this section, we first present the shadow detection network then the shadow removal network.
Figure~\ref{fig:arc} presents our overall shadow detection network, which employs multiple DSC modules (see Figure~\ref{fig:context}) to learn the direction-aware spatial context features in different scales.
%
%
Our network takes the whole image as input and outputs the shadow mask in an end-to-end manner.

First, it begins by using a convolutional neural network (CNN) to extract hierarchical feature maps in different scales over the CNN layers.
Feature maps at the shallower layers encode the fine details, which help preserve the shadow boundaries, while feature maps at the deep layers carry more global semantics, which help recognize the shadow and non-shadow regions.
%
Second, for each layer, we employ a DSC module to harvest spatial context in a direction-aware manner and produce the DSC features.
Third, we concatenate the DSC features with the corresponding convolutional features and upsample the concatenated feature maps to the size of the input.
Fourth, to leverage the complementary advantages of feature maps at different layers, we concatenate the upsampled feature maps and adopt a $1$$\times$$1$ convolution layer to produce the multi-level integrated features (MLIF).
Further, we apply the deep supervision mechanism~\cite{lee2015deeply,xie2015holistically} to impose a supervision signal to each layer as well as to the MLIF, and predict a shadow mask for each of them.
In the training process, we simultaneously minimize the prediction errors from multiple layers and obtain more discriminative features 
by directly providing the supervisions to the intermediate layers~\cite{lee2015deeply}.
Lastly, we concatenate all the predicted shadow masks and adopt a $1$$\times$$1$ convolution layer to generate the output shadow mask; see Figure~\ref{fig:arc}. 
In testing, we compute the mean shadow mask over the MLIF layer and fusion layer to produce the final prediction result, and adopt the fully connected conditional random field (CRF)~\cite{krahenbuhl2011efficient} to refine the result.
To adopt the network for shadow removal, we replace the shadow masks with shadow-free images as the ground truth, formulate a color compensation mechanism to adjust the shadow-free images for color and luminosity consistencies, and use a Euclidean loss to optimize the network; see Section~\ref{subsec:3.3} for details.


\begin{figure}[tp]
	\centering
	\includegraphics[width=0.98\linewidth]{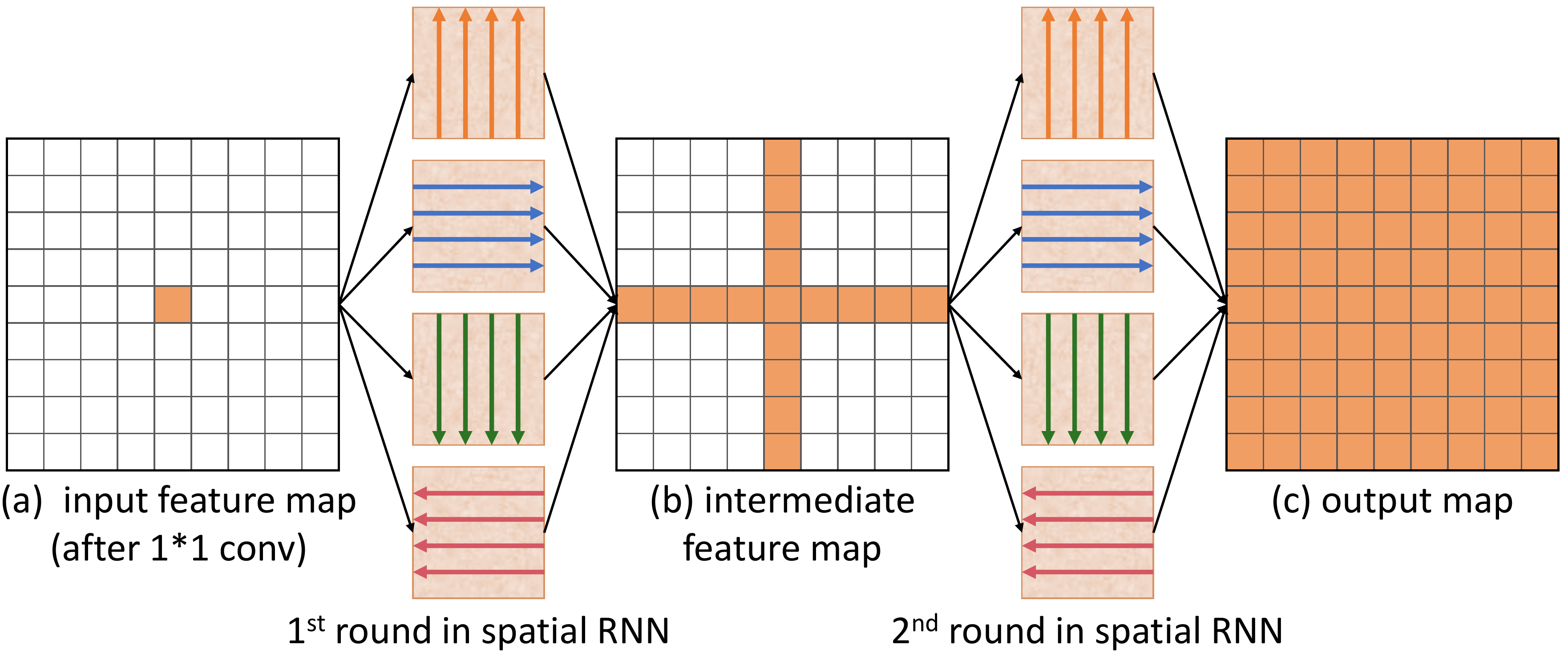}
	\caption{The schematic illustration of how spatial context information propagates in a two-round spatial RNN.}
	\label{fig:IRNN}
\end{figure}

In the following subsections, we first elaborate the DSC module that generates the DSC features (Section~\ref{subsec:3.1}).
After that, we present how we design the shadow detection network in Figure~\ref{fig:arc} using the DSC modules (Section~\ref{subsec:3.2}) then present how we adopt the network further for shadow removal (Section~\ref{subsec:3.3}).



\subsection{Direction-aware Spatial Context}
\label{subsec:3.1}

Figure~\ref{fig:context} shows our DSC module architecture, which takes feature maps as input and outputs the DSC features.
In this subsection, we first describe the concept of spatial context features and the spatial RNN model (Section~\ref{subsec:3.1.1}) then elaborate on how we formulate the direction-aware attention mechanism in a spatial RNN to learn the attention weights and generate DSC features (Section~\ref{subsec:3.1.2}).


\subsubsection{Spatial Context Features}
\label{subsec:3.1.1}

A recurrent neural network (RNN)~\cite{lecun2015deep} is an effective model to process 1D sequential data via three arrays of nodes: (i) an array of input nodes to receive data, (ii) an array of hidden nodes to update the internal states based on past and present data, and (iii) an array of output nodes to output data.
There are three kinds of data translations in an RNN: (i) from input nodes to hidden nodes, (ii) between adjacent hidden nodes, and (iii) from hidden nodes to output nodes.
By iteratively performing the data translations, the data received at the input nodes can propagate across the hidden nodes, and eventually produce target results at the output nodes.

For processing image data with 2D spatial context, RNNs have been extended to build the spatial RNN model~\cite{bell2016inside}; see the schematic illustration in Figure~\ref{fig:IRNN}.
Taking a 2D feature map from a CNN as input, 
we first perform a $1$$\times$$1$ convolution to simulate the input-to-hidden data translation in the RNN.
Then, we apply four independent data translations to aggregate the local spatial context along each principal direction (left, right, up, and down), and fuse the results into an intermediate feature map; see Figure~\ref{fig:IRNN}(b).
Lastly, we repeat the whole process to further propagate the aggregated spatial context in each principal direction and to generate the overall spatial context; see Figure~\ref{fig:IRNN}(c).

Comparing with Figure~\ref{fig:IRNN}(c), each pixel in Figure~\ref{fig:IRNN}(a) knows only its local spatial context, while each pixel in Figure~\ref{fig:IRNN}(b) further knows the spatial context in the four principal directions after the first round of data translations.
Therefore, after two rounds of data translations, each pixel can obtain relevant direction-aware global spatial context for learning the features to enhance the detection and removal of shadows.

To perform the data translations in a spatial RNN, we follow the IRNN model~\cite{le2015simple}, since it is fast, easy to train, and has a good performance for long-range data dependencies~\cite{bell2016inside}.
Denoting $h_{i,j}$ as the feature at pixel $(i,j)$, we perform one round of data translations to the right (similarly, for each of the other three directions) by repeating the following operation $n$ times.
\begin{equation}  \label{recurrent}
h_{i,j} = \max( \ \alpha_\text{right} \ h_{i,j-1}+h_{i,j} \ , \ 0 \ )\ ,
\end{equation}
where $n$ is the width of the feature map and $\alpha_\text{right}$ is the weight parameter in the recurrent translation layer for the right direction.
Note that $\alpha_\text{right}$, as well as the weights for the other directions, are initialized to be an identity matrix and are learned automatically through the training process.


\subsubsection{Direction-aware Spatial Context Features}
\label{subsec:3.1.2}

To efficiently learn the spatial context in a direction-aware manner, we further formulate the direction-aware attention mechanism in a spatial RNN to learn the attention weights and generate the direction-aware spatial context (DSC) features.
This design forms the DSC module we presented in Figure~\ref{fig:context}.


\vspace*{2mm}
\noindent
{\bf Direction-aware attention mechanism.} \
The purpose of the mechanism is to enable the spatial RNN to selectively leverage the spatial context aggregated in different directions through learning.
See the top-left blocks in the DSC module shown in Figure~\ref{fig:context}.
First, we employ two successive convolutional layers (with $3$$\times$$3$ kernels) followed by the ReLU~\cite{krizhevsky2012imagenet} non-linear operation then the third convolutional layer (with $1$$\times$$1$ kernels) to generate $\mathbf{W}$ of four channels.
We then split $\mathbf{W}$ into four maps of attention weights denoted as $\mathbf{W}_\text{left}$, $\mathbf{W}_\text{down}$, $\mathbf{W}_\text{right}$, and $\mathbf{W}_\text{up}$, each of one channel.
Mathematically, if we denote the above operators as $f_{att}$ and the input feature maps as $\mathbf{X}$, we have
\begin{equation}
\label{attention}
\mathbf{W} \ = \ f_{att}( \ \mathbf{X} \ ; \ \theta \ )\ ,
\end{equation}
where $\theta$ denotes the parameters in the convolution operations to be learned by $f_{att}$ (also known as the attention estimator network).

See again the DSC module shown in Figure~\ref{fig:context}.
The four maps of weights are multiplied with the spatial context features (from the recurrent data translations) in corresponding directions in an element-wise manner.
Hence, after we train the network, the network should learn $\theta$ for producing suitable attention weights to selectively leverage the spatial context in the spatial RNN.


\vspace*{2mm}
\noindent
{\bf Completing the DSC module.} \
Next, we provide details on the DSC module.
As shown in Figure~\ref{fig:context}, after we multiply the spatial context features with the attention weights, we concatenate the results and use a $1$$\times$$1$ convolution to simulate the hidden-to-hidden data translation in the RNN and reduce the feature dimensions by a quarter.
Then, we perform the second round of recurrent translations and use the same set of attention weights to select the spatial context, where we empirically found that sharing the attention weights instead of using two separate sets of weights leads to higher performance; see Section~\ref{sec:experiments_removal} for an experiment.
Note further that these attention weights are automatically learnt based on the deep features extracted from the input images, so they may vary from image to image.
Lastly, we use a $1$$\times$$1$ convolution followed by the ReLU~\cite{krizhevsky2012imagenet} non-linear operation on the concatenated feature maps to simulate the hidden-to-output translation and produce the output DSC features.


\subsection{Our Shadow Detection Network}
\label{subsec:3.2}

Our network is built upon the VGG network~\cite{simonyan2014very} with one DSC module applied to each layer, except for the first layer due to a large memory footprint.
Since it has a fully convolutional architecture with only convolution and pooling operations, the directional relationship among the pixels in the image space captured by the DSC module is thus preserved in the feature space.


\subsubsection{Training}


\vspace*{2mm}
\noindent
{\bf Loss function.} \
In natural images, shadows usually occupy smaller areas in the image space than the non-shadow regions.
Hence, if the loss function simply aims for the overall accuracy, it will incline to match the non-shadow regions, which have far more pixels.
Therefore, we use a weighted cross-entropy loss to optimize the shadow detection network in the training process.

Consider $y$ as the ground truth value of a pixel (where $y$$=$$1$, if it is in shadow, and $y$$=$$0$, otherwise) and $p$ as the pixel's prediction label (where $p\in[0,1]$).
The weighted cross entropy loss $L_i$ for the $i$-th CNN layer is a summation of the cross entropy loss weighted by the class distribution $L^d_i$ and the cross entropy loss weighted by per-class accuracy $L^a_i$ (i.e., $L_i = L^d_i + L^a_i$):
\begin{equation}  \label{loss1}
L^d_i = -(\frac{N_n}{N_p+N_n}) y \log(p) - (\frac{N_p}{N_p+N_n})(1 - y) \log(1 - p)\ ,
\end{equation}
and
\begin{equation}  \label{loss2}
L^a_i = -(1-\frac{TP}{N_p}) y \log(p) - (1-\frac{TN}{N_n})(1 - y) \log(1 - p)\ ,
\end{equation}
%
where $TP$ and $TN$ are the numbers of true positives and true negatives per image, and $N_p$ and $N_n$ are the numbers of shadow and non-shadow pixels per image, respectively, so $N_p$$+$$N_n$ is the total number of pixels in the $i$-th layer.
In practice, $L^d_i$ helps balance the detection of shadows and non-shadows; if the area of shadows is less than that of the non-shadow region, we penalize misclassified shadow pixels more than the misclassified non-shadow pixels.
On the other hand, inspired by~\cite{shrivastava2016training}, which has a higher preference to select misclassified examples to train the deep network, we enlarge the weights on the class (shadow or non-shadow) that is difficult to be classified.
To do so, we employ $L^a_i$, where the weight for shadow (or non-shadow) class is large when the number of correctly-classified shadow (or non-shadow) pixels is small, and vice versa.
We use the above loss function per layer in the shadow detection network presented in Figure~\ref{fig:arc}.
Hence, the overall loss function $L_\text{overall}$ is a summation of the individual loss on all the predicted shadow masks over the different scales:
\begin{equation}
\label{loss_t}
L_\text{overall} \ = \ \sum_{i} w_i L_i \ + \ w_m L_m \ + \ w_f L_f \ ,
\end{equation}
where $w_i$ and $L_i$ denote the weight and loss of the $i$-th layer (level) in the overall network, respectively;
$w_m$ and $L_m$ are the weight and loss of the MLIF layer; and
$w_f$ and $L_f$ are the weight and loss of the fusion layer, which is the last layer in the overall network;
see Figure~\ref{fig:arc}.
Note that $w_i$, $w_m$ and $w_f$ are empirically set to be one; see supplementary material for the related experiment.


\vspace*{2mm}
\noindent
{\bf Training parameters.} \
To accelerate the training process while reducing the overfitting, 
we take the weights of the VGG network~\cite{simonyan2014very} trained on ImageNet~\cite{deng2009imagenet} for a classification task to initialize the parameters in the feature extraction layers (see the frontal part of the network in Figure~\ref{fig:arc}) and initialize the parameters in the other layers by random noise, which follows a zero-mean Gaussian distribution with standard deviation of 0.1.
Stochastic gradient descent is used to optimize the whole network with a momentum value of 0.9 and a weight decay of $5$$\times$$10^{-4}$.
We empirically set the learning rate as $10^{-8}$ and terminate the learning process after 12k iterations; see supplementary material for the related experiment on the training iterations.
Moreover, we horizontally flip images for data argumentation.
Due to the limitation of GPU memory, we build the model in Caffe~\cite{jia2014caffe} with a mini-batch size of one, and update the model parameters in every ten training iterations.


\subsubsection{Testing}
In the testing process, our network produces one shadow mask per layer, including the MLIF layer and fusion layer, each with a supervision signal.
After that, we compute the mean shadow mask over the MLIF layer and fusion layer to produce the final prediction.
Lastly, we apply the fully connected conditional random field (CRF)~\cite{krahenbuhl2011efficient} to improve the detection result by considering the spatial coherence among the neighborhood pixels; see Figure~\ref{fig:arc}.


\subsection{Our Shadow Removal Network}
\label{subsec:3.3}

\begin{figure} [tp]
	\centering
	\includegraphics[width=0.98\linewidth]{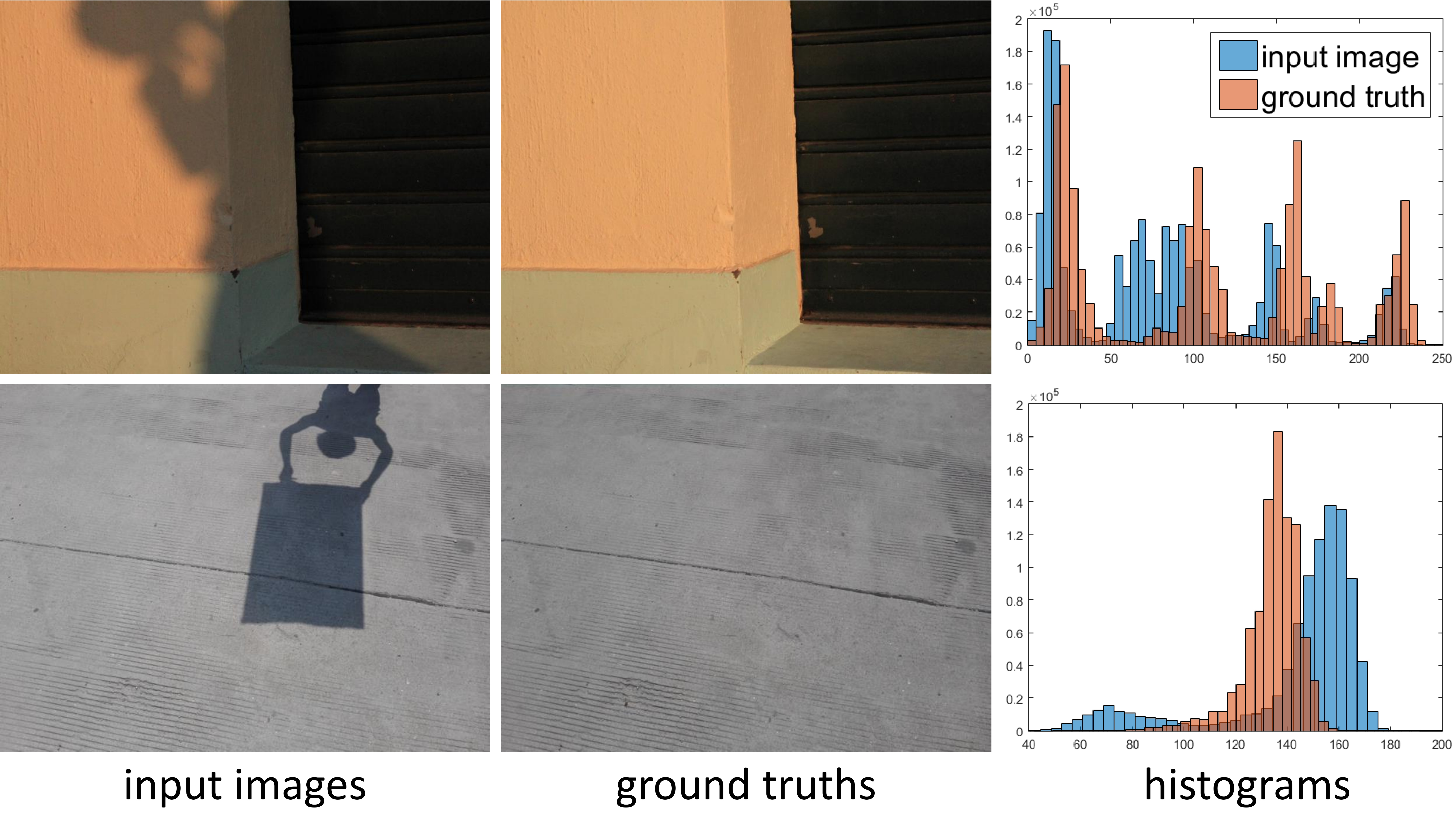}
	\caption{Inconsistencies between input (shadow images) and ground truth (shadow-free images).
Top row is ``IMG\_6456.jpg'' from SRD~\cite{qu2017deshadownet} and bottom row is ``109-5.png'' from ISTD~\cite{wang2018stacked}.}
	\label{fig:sample_compensation}
\end{figure}

To adopt our shadow detection network shown in Figure~\ref{fig:arc} for shadow removal, we have the following three modifications:
\begin{itemize}[leftmargin=*]
\item
First, we formulate a color compensation mechanism to address the color inconsistencies between the training pairs, i.e., shadow images (input) and shadow-free images (ground truth), and then to adjust the shadow-free images (Section~\ref{sssec:color_comp}).
%
\item
Second, we replace the shadow masks by the adjusted shadow-free images as the supervision (i.e., ground truth images) in the network for shadow removal; see Figure~\ref{fig:arc}.
\item
Third, we replace the weighted cross-entropy loss by a Euclidean loss to train and optimize the network using the adjusted shadow-free images (Section~\ref{sssec:training}).
%
\end{itemize}


\subsubsection{Color Compensation Mechanism}
\label{sssec:color_comp}

Training data for shadow removal is typically prepared by first taking a picture of the scene with shadows, and then taking another picture without the shadows by removing the associated objects.
Since the environmental luminosity and camera exposure may vary, a training pair may have inconsistent colors and luminosity; see Figure~\ref{fig:sample_compensation} for examples from two different benchmark datasets (SRD~\cite{qu2017deshadownet} \& ISTD~\cite{wang2018stacked}), where the inconsistencies are clearly revealed by the color histograms.
Existing network-based methods learn to remove shadows by optimizing the network to produce an output that matches the target ground truth.
Hence, given such inconsistent training pairs, the network could produce biased results that are brighter or darker.

To address the problem, we design a color compensation mechanism by finding a color transfer function for each pair of training images (input shadow images and ground truth shadow-free images).
Let $I_s$ and $I_n$ be a shadow image (input) and a shadow-free image (ground truth) of a training pair, respectively, and $\Omega_s$ and $\Omega_n$ be the shadow region and non-shadow region, respectively, in the image space.
In our formulation, we aim to find color transfer function $T_f$ that minimizes the color compensation error $E_c$ between the shadow image and shadow-free image over the non-shadow region (indicated by the shadow mask):
\begin{equation} \label{color_compensation}
\ E_c \ = \ | \ I_s \ - \ T_f(I_n) \ |^2_{\Omega_n} \ .
\end{equation}
We formulate $T_f$ using the following linear transformation (with which we empirically find sufficient for adjusting the colors in $I_n$ to match the colors in $I_s$):
\begin{equation} \label{color_transfer_func}
T_f(x) \ = \ \mathbf{M}_{\alpha} \ \cdot
\left(\begin{array}{c}
x_r \\
x_g \\
x_b \\
1
\end{array}\right)
\ ,
\end{equation}
%
where $x$ is a pixel in $I_n$ with color values $(x_r,x_g,x_b)$ and $\mathbf{M}_{\alpha}$ is a $3$$\times$$4$ matrix, which stores the parameters in the color transfer function.
Note that we solve Eq.~\eqref{color_compensation} for $T_f$ using the least-squares method by considering pixel pairs in the non-shadow regions $\Omega_n$ of $I_s$ and $I_n$.
Then, we apply $T_f$ to adjust the whole image of $I_n$ for each training pair, replace the shadow masks in Figure~\ref{fig:arc} by the adjusted shadow-free image (i.e., $T_f(I_n)$) as the new supervision, and train the shadow removal network in an end-to-end manner.

\subsubsection{Training}
\label{sssec:training}

\vspace*{2mm}
\noindent
{\bf Loss function.} \
We adopt a Euclidean loss to optimize the shadow removal network. 
In detail, we denote the network prediction as $\tilde{I}_n$, use the ``LAB'' color space for both $T_f(I_n)$ and $\tilde{I}_n$ in the training, and calculate the loss $L^r$ on the whole image domain:
%
\begin{equation}  \label{loss_removal}
L^r \ = \ | \ T_f(I_n) \ - \ \tilde{I}_n \ |^2_{\Omega_n \cup \Omega_s} \ .
\end{equation} 

We use the above loss function for each layer in the shadow removal network. 
The overall loss function $L_\text{overall}^r$ is the summation of the loss $L^r$ on all the $i$-th layers $L_i^r$, the MLIF layer $L_m^r$, and the fusion layer $L_f^r$:
\begin{equation}
\label{loss_t_removal}
L_\text{overall}^r \ = \ \sum_{i} w_i^r L_i^r \ + \ w_m^r L_m^r \ + \ w_f^r L_f^r \ .
\end{equation}
Similar to Eq.~\eqref{loss_t}, we empirically set $w_i^r$, $w_m^r$, and $w_f^r$ to one.

\vspace*{2mm}
\noindent
{\bf Training parameters.} \
Again, we initialize the parameters in the feature extraction layers (see the frontal part of the network shown in Figure~\ref{fig:arc}) by the well-trained VGG network~\cite{simonyan2014very} on ImageNet~\cite{deng2009imagenet} to accelerate the training process and reduce the overfitting, and initialize the parameters in the other layers by random noise as in shadow detection.
We use Adam~\cite{kingma2014adam} to optimize the shadow removal network with the first momentum value 0.9, second momentum value 0.99, and weight decay $5$$\times$$10^{-4}$.
This optimization approach adaptively adjusts the learning rates for individual parameters in the network. 
It decreases the learning rate for the frequently-updated parameters and increases the learning rate for the rarely-updated parameters.
We empirically set the basic learning rate as $10^{-5}$, reduce it by multiplying 0.316 at 90k and 130k iterations, following~\cite{santhanam2017generalized}, and stop the learning at 160k iterations.
Moreover, the images are horizontally and vertically flipped, randomly cropped, and rotated for data argumentation.
The model is built in Caffe~\cite{jia2014caffe} with a mini-batch size of one.


\subsubsection{Testing}
In the testing process, our network directly produces a shadow-free image for each layer, including the MLIF layer and fusion layer, each with a supervision signal.
After that, we compute the mean shadow-free image over the MLIF layer and fusion layer to produce the final result.

\section{Experiments on Shadow Detection}
\label{sec:experiments}

In this section, we present experiments to evaluate our shadow detection network:
comparing it with the state-of-the-art methods,
evaluating its network design and time performance, and 
showing the shadow detection results.
In the next section, we will present results and evaluations on the shadow removal network.

\begin{figure*}[t]
	\centering
	\begin{subfigure}{0.104\textwidth} 
		\includegraphics[width=\textwidth]{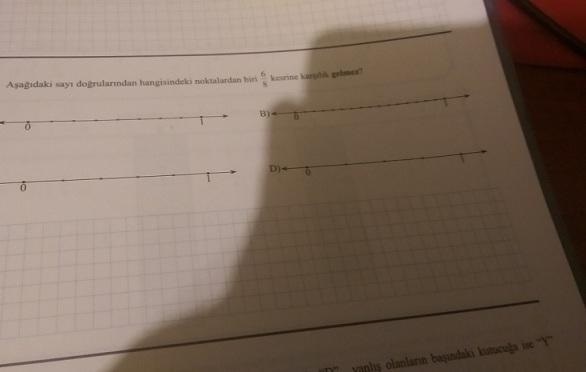}
	\end{subfigure}
	\begin{subfigure}{0.104\textwidth}
		\includegraphics[width=\textwidth]{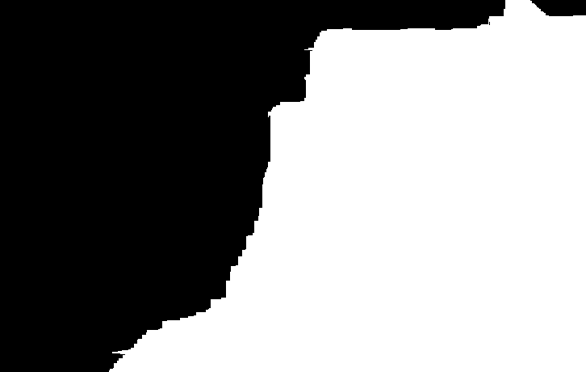}
	\end{subfigure}
	\begin{subfigure}{0.104\textwidth}
		\includegraphics[width=\textwidth]{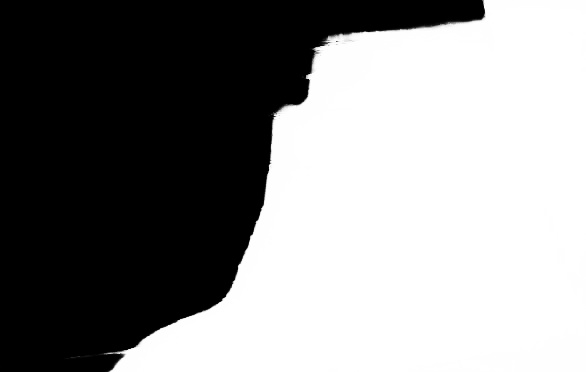}
	\end{subfigure}
	\begin{subfigure}{0.104\textwidth}
		\includegraphics[width=\textwidth]{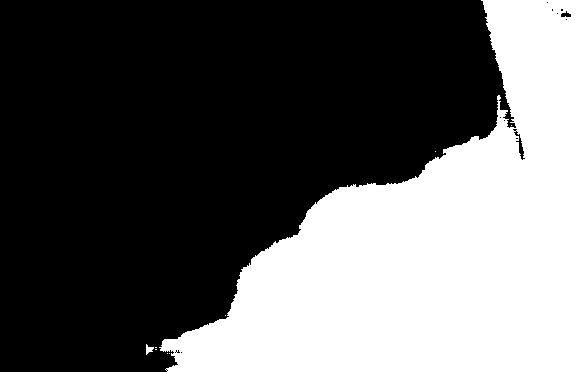}
	\end{subfigure}
	\begin{subfigure}{0.104\textwidth}
		\includegraphics[width=\textwidth]{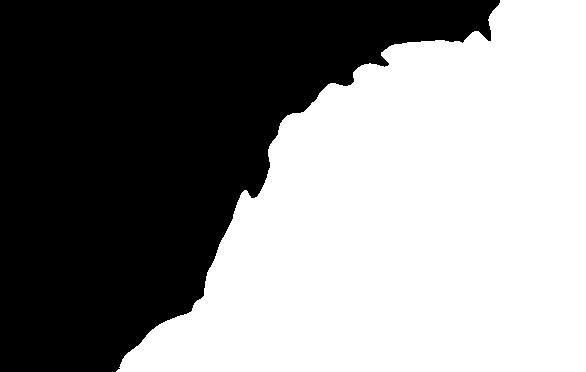}
	\end{subfigure}
	\begin{subfigure}{0.104\textwidth}
		\includegraphics[width=\textwidth]{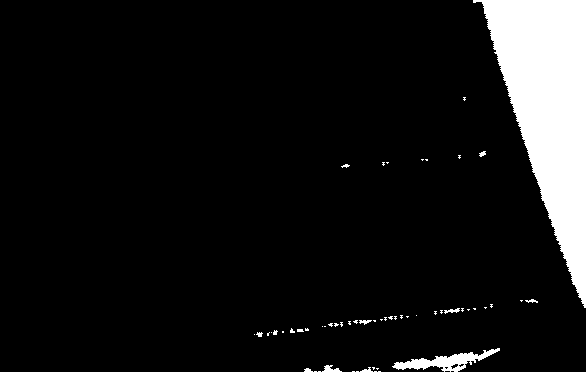}
	\end{subfigure}
	\begin{subfigure}{0.104\textwidth}
		\includegraphics[width=\textwidth]{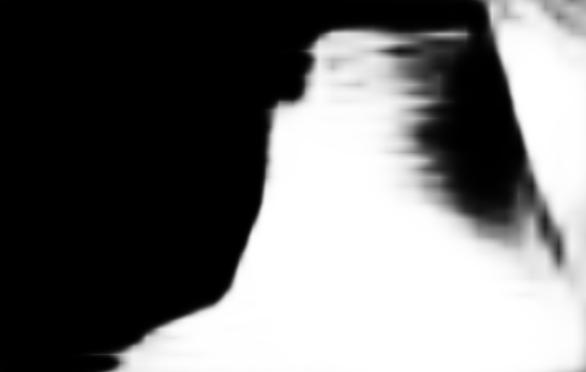}
	\end{subfigure}
	\begin{subfigure}{0.104\textwidth}
		\includegraphics[width=\textwidth]{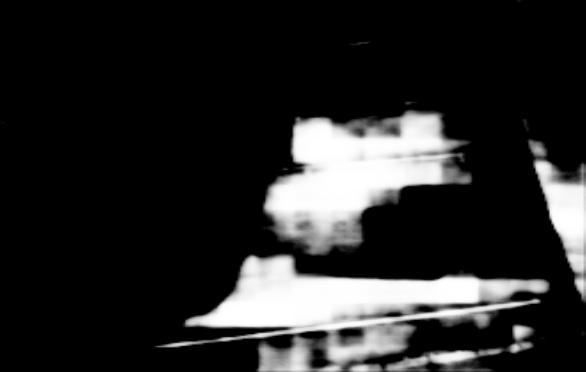}
	\end{subfigure}
	\begin{subfigure}{0.104\textwidth}
		\includegraphics[width=\textwidth]{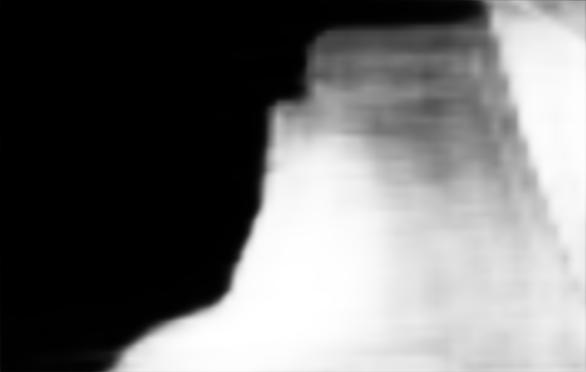}
	\end{subfigure}
	\ \\
	\vspace*{0.5mm}
	\begin{subfigure}{0.104\textwidth}
		\includegraphics[width=\textwidth]{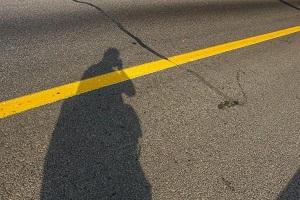}
	\end{subfigure}
	\begin{subfigure}{0.104\textwidth}
		\includegraphics[width=\textwidth]{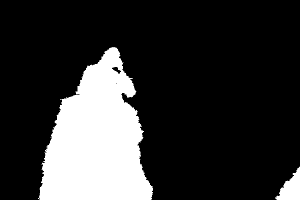}
	\end{subfigure}
	\begin{subfigure}{0.104\textwidth}
		\includegraphics[width=\textwidth]{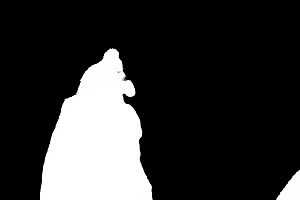}
	\end{subfigure}
	\begin{subfigure}{0.104\textwidth}
		\includegraphics[width=\textwidth]{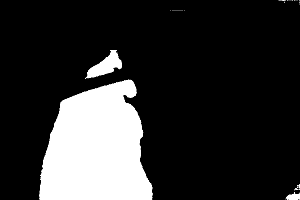}
	\end{subfigure}
	\begin{subfigure}{0.104\textwidth}
		\includegraphics[width=\textwidth]{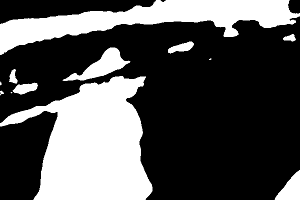}
	\end{subfigure}
	\begin{subfigure}{0.104\textwidth}
		\includegraphics[width=\textwidth]{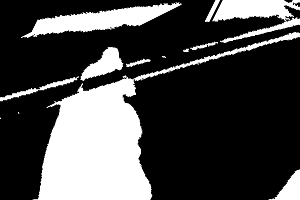}
	\end{subfigure}
	\begin{subfigure}{0.104\textwidth}
		\includegraphics[width=\textwidth]{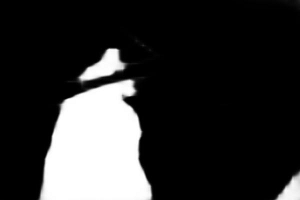}
	\end{subfigure}
	\begin{subfigure}{0.104\textwidth}
		\includegraphics[width=\textwidth]{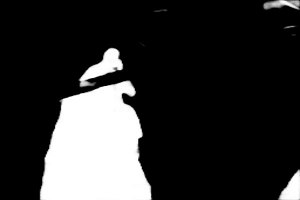}
	\end{subfigure}
	\begin{subfigure}{0.104\textwidth}
		\includegraphics[width=\textwidth]{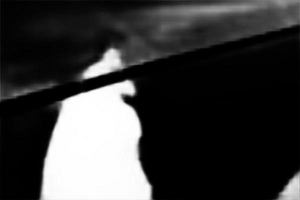}
	\end{subfigure}
	\ \\
	\vspace*{0.5mm}
	\begin{subfigure}{0.104\textwidth}
		\includegraphics[width=\textwidth]{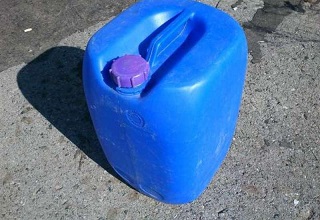}
	\end{subfigure}
	\begin{subfigure}{0.104\textwidth}
		\includegraphics[width=\textwidth]{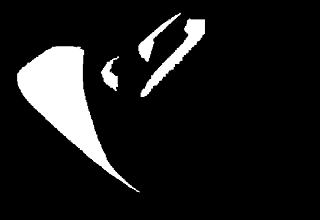}
	\end{subfigure}
	\begin{subfigure}{0.104\textwidth}
		\includegraphics[width=\textwidth]{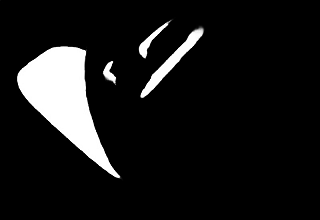}
	\end{subfigure}
	\begin{subfigure}{0.104\textwidth}
		\includegraphics[width=\textwidth]{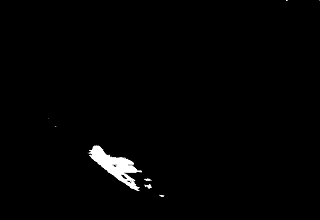}
	\end{subfigure}
	\begin{subfigure}{0.104\textwidth}
		\includegraphics[width=\textwidth]{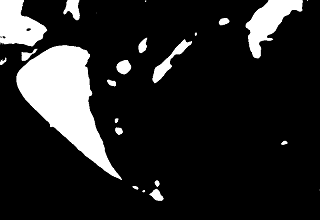}
	\end{subfigure}
	\begin{subfigure}{0.104\textwidth}
		\includegraphics[width=\textwidth]{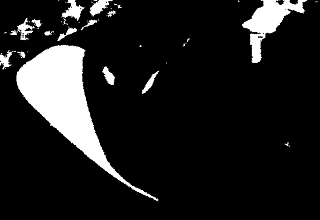}
	\end{subfigure}
	\begin{subfigure}{0.104\textwidth}
		\includegraphics[width=\textwidth]{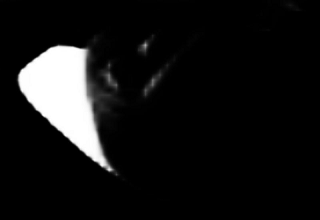}
	\end{subfigure}
	\begin{subfigure}{0.104\textwidth}
		\includegraphics[width=\textwidth]{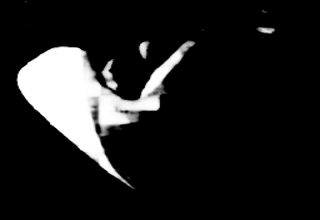}
	\end{subfigure}
	\begin{subfigure}{0.104\textwidth}
		\includegraphics[width=\textwidth]{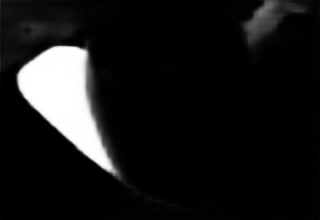}
	\end{subfigure}
	\ \\
	\vspace*{0.5mm}
	\begin{subfigure}{0.104\textwidth}
		\includegraphics[width=\textwidth]{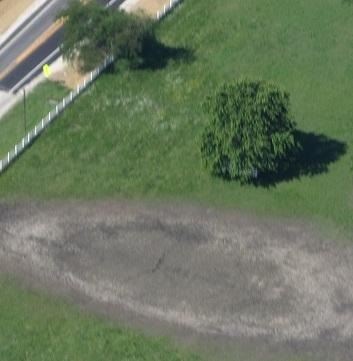}
	\end{subfigure}
	\begin{subfigure}{0.104\textwidth}
		\includegraphics[width=\textwidth]{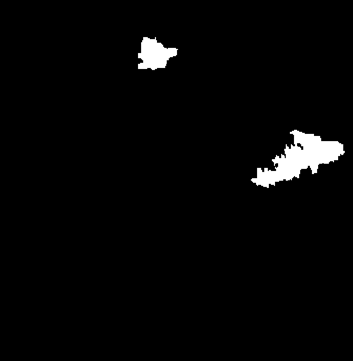}
	\end{subfigure}
	\begin{subfigure}{0.104\textwidth}
		\includegraphics[width=\textwidth]{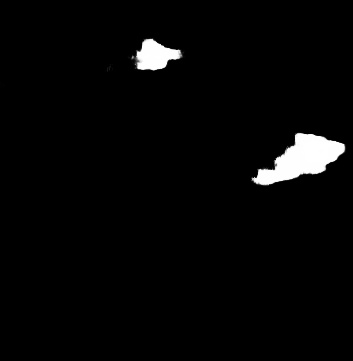}
	\end{subfigure}
	\begin{subfigure}{0.104\textwidth}
		\includegraphics[width=\textwidth]{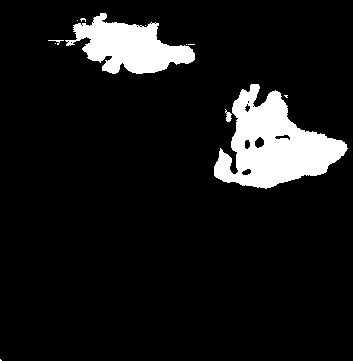}
	\end{subfigure}
	\begin{subfigure}{0.104\textwidth}
		\includegraphics[width=\textwidth]{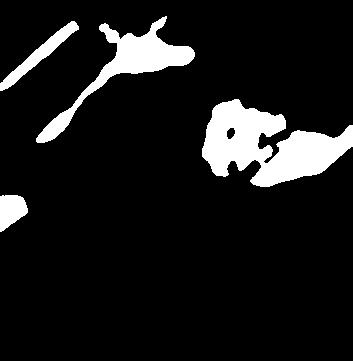}
	\end{subfigure}
	\begin{subfigure}{0.104\textwidth}
		\includegraphics[width=\textwidth]{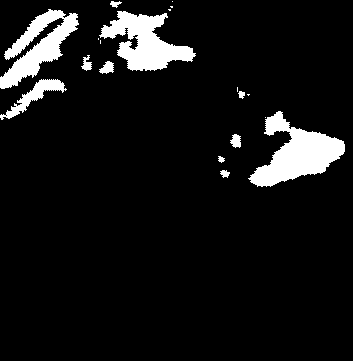}
	\end{subfigure}
	\begin{subfigure}{0.104\textwidth}
		\includegraphics[width=\textwidth]{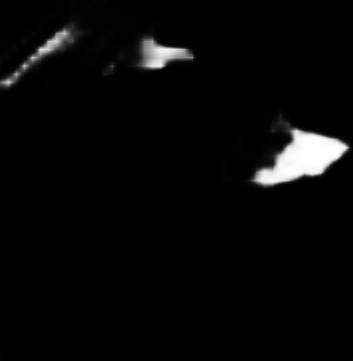}
	\end{subfigure}
	\begin{subfigure}{0.104\textwidth}
		\includegraphics[width=\textwidth]{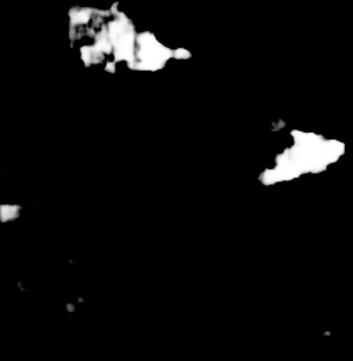}
	\end{subfigure}
	\begin{subfigure}{0.104\textwidth}
		\includegraphics[width=\textwidth]{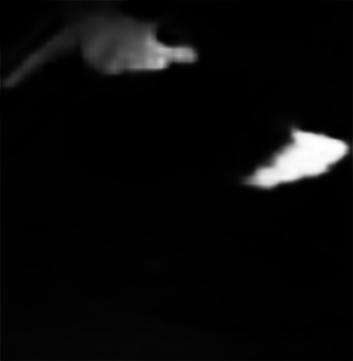}
	\end{subfigure}
	\ \\
	\vspace*{0.5mm}
	\begin{subfigure}{0.104\textwidth} 
		\includegraphics[width=\textwidth]{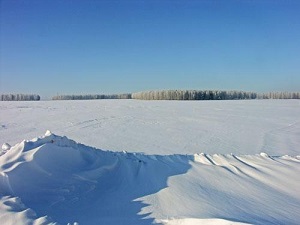}
	\end{subfigure}
	\begin{subfigure}{0.104\textwidth}
		\includegraphics[width=\textwidth]{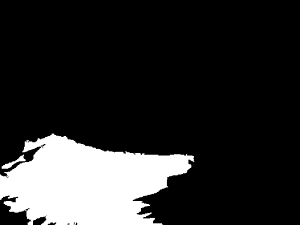}
	\end{subfigure}
	\begin{subfigure}{0.104\textwidth}
		\includegraphics[width=\textwidth]{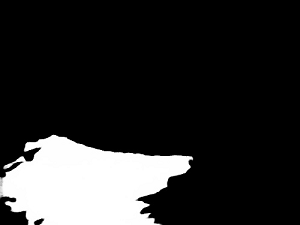}
	\end{subfigure}
	\begin{subfigure}{0.104\textwidth}
		\includegraphics[width=\textwidth]{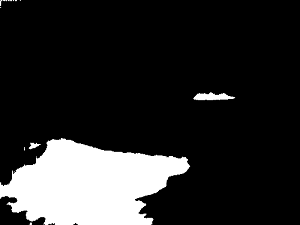}
	\end{subfigure}
	\begin{subfigure}{0.104\textwidth}
		\includegraphics[width=\textwidth]{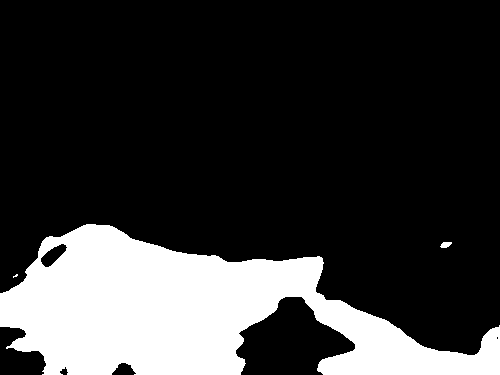}
	\end{subfigure}
	\begin{subfigure}{0.104\textwidth}
		\includegraphics[width=\textwidth]{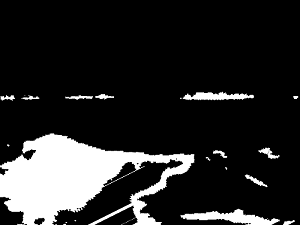}
	\end{subfigure}
	\begin{subfigure}{0.104\textwidth}
		\includegraphics[width=\textwidth]{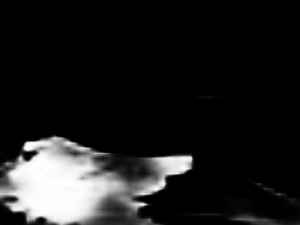}
	\end{subfigure}
	\begin{subfigure}{0.104\textwidth}
		\includegraphics[width=\textwidth]{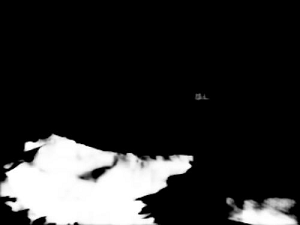}
	\end{subfigure}
	\begin{subfigure}{0.104\textwidth}
		\includegraphics[width=\textwidth]{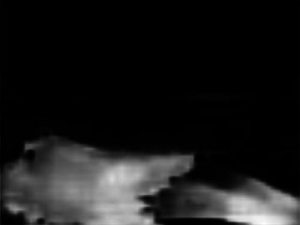}
	\end{subfigure}
	\ \\
	\vspace*{0.5mm}
	\begin{subfigure}{0.104\textwidth}
		\includegraphics[width=\textwidth]{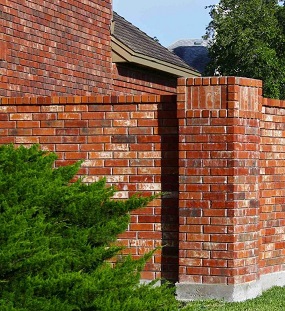}
		\vspace{-5.5mm} \caption*{{\footnotesize input}}
	\end{subfigure}
	\begin{subfigure}{0.104\textwidth}
		\includegraphics[width=\textwidth]{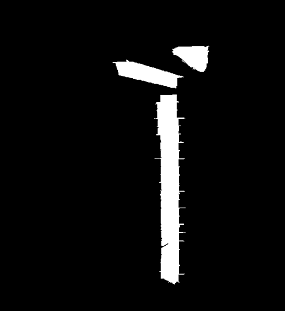}
		\vspace{-5.5mm} \caption*{{\footnotesize ground truth}} 
	\end{subfigure}
	\begin{subfigure}{0.104\textwidth}
		\includegraphics[width=\textwidth]{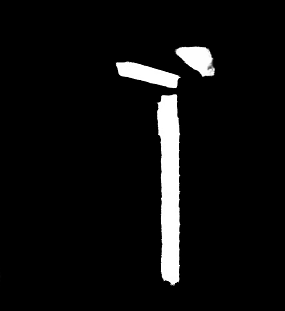}
		\vspace{-5.5mm} \caption*{{\footnotesize DSC (ours)}} 
	\end{subfigure}
	\begin{subfigure}{0.104\textwidth}
		\includegraphics[width=\textwidth]{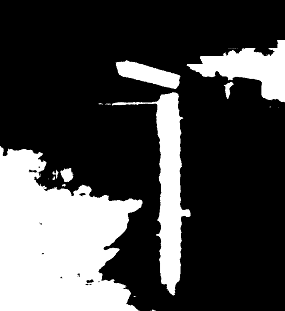}
		\vspace{-5.5mm} \caption*{{\footnotesize scGAN~\cite{nguyen2017shadow}}}
	\end{subfigure}
	\begin{subfigure}{0.104\textwidth}
		\includegraphics[width=\textwidth]{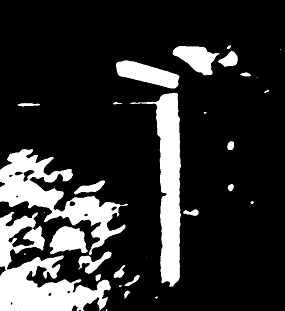}
		\vspace{-5.5mm} \caption*{{\footnotesize stkd'-CNN~\cite{vicente2016large}}}
	\end{subfigure}
	\begin{subfigure}{0.104\textwidth}
		\includegraphics[width=\textwidth]{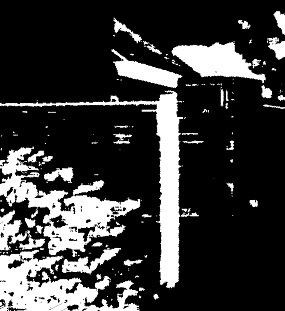}
		\vspace{-5.5mm} \caption*{{\footnotesize patd'-CNN~\cite{hosseinzadeh2017fast}}}
	\end{subfigure}
	\begin{subfigure}{0.104\textwidth}
		\includegraphics[width=\textwidth]{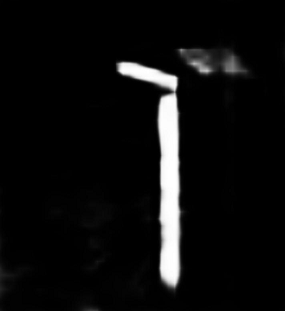}
		\vspace{-5.5mm} \caption*{{\footnotesize SRM~\cite{wang2017stagewise}}}
	\end{subfigure}
	\begin{subfigure}{0.104\textwidth}
		\includegraphics[width=\textwidth]{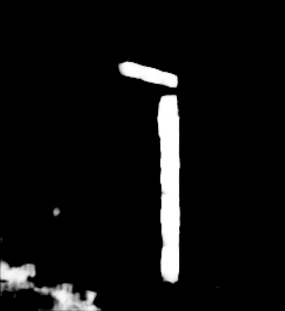}
		\vspace{-5.5mm} \caption*{{\footnotesize Amulet~\cite{zhang2017amulet}}}
	\end{subfigure}
	\begin{subfigure}{0.104\textwidth}
		\includegraphics[width=\textwidth]{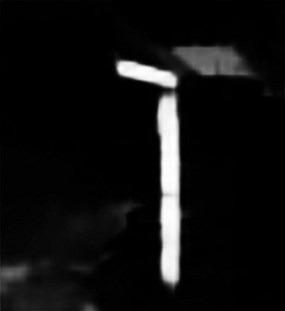}
		\vspace{-5.5mm} \caption*{\hspace*{-0.7mm}{\footnotesize PSPNet~\cite{Zhao_2017_CVPR}}}
	\end{subfigure}
	\vspace*{-1mm}
	\caption{Visual comparison of shadow masks produced by our method and other methods ($4^{th}$-$9^{th}$ columns) against ground truth images shown in $2^{nd}$ column.
		Note that stkd'-CNN and patd'-CNN stand for stacked-CNN and patched-CNN, respectively.}
	\label{fig:comparison_real_photos}
	\vspace*{-2mm}
\end{figure*}


\subsection{Shadow Detection Datasets \& Evaluation Metrics}


\vspace*{2mm}
\noindent
{\bf Benchmark datasets.} \
We employ two benchmark datasets.
The first one is the SBU Shadow Dataset~\cite{vicente2016noisy,vicente2016large}, which is the largest publicly available annotated shadow dataset with 4089 training images and 638 testing images, which cover a wide variety of scenes.
%
The second dataset we employed is the UCF Shadow Dataset~\cite{zhu2010learning}.
It includes 221 images that are divided into 111 training images and 110 testing images, following~\cite{shen2015shadow}.
%
We train our shadow detection network using the SBU training set.


\vspace*{2mm}
\noindent
{\bf Evaluation metrics.} \
We employ two commonly-used metrics to quantitatively evaluate the shadow detection performance.
The first one is the accuracy metric:
\begin{equation}
accuracy \ = \ \frac{TP+TN}{N_p+N_n} \ ,
\end{equation}
where $TP$, $TN$, $N_p$ and $N_n$ are true positives, true negatives, number of shadow pixels, and number of non-shadow pixels, respectively, as defined in Section~\ref{subsec:3.2}.
Since $N_p$ is usually much smaller than $N_n$ in natural images, we employ the second metric called the balance error rate (BER) to obtain a more balanced evaluation by equally treating the shadow and non-shadow regions:
\begin{equation}
BER \ = \ (1-\frac{1}{2}(\frac{TP}{N_p}+\frac{TN}{N_n}))\times 100 \ .
\end{equation}
Note that unlike the accuracy metric, for BER, the lower its value, the better the detection result is.

\begin{figure*}[tp]
	\centering
	
	\begin{subfigure}{0.104\textwidth} 
		\includegraphics[width=\textwidth]{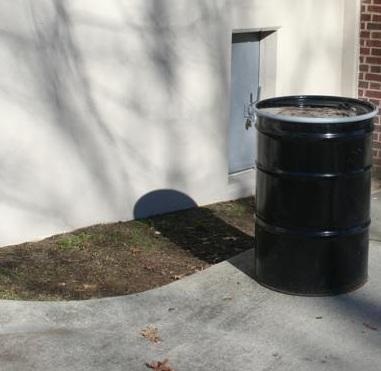}
	\end{subfigure}
	\begin{subfigure}{0.104\textwidth}
		\includegraphics[width=\textwidth]{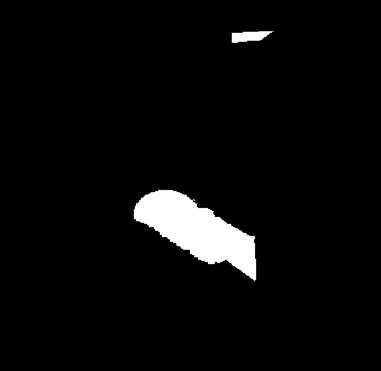}
	\end{subfigure}
	\begin{subfigure}{0.104\textwidth}
		\includegraphics[width=\textwidth]{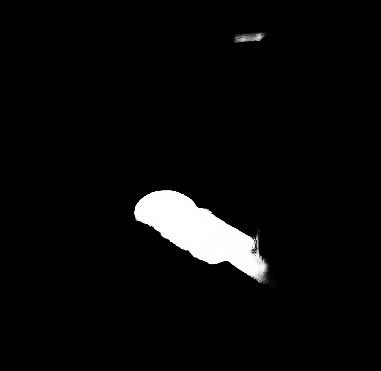}
	\end{subfigure}
	\begin{subfigure}{0.104\textwidth}
		\includegraphics[width=\textwidth]{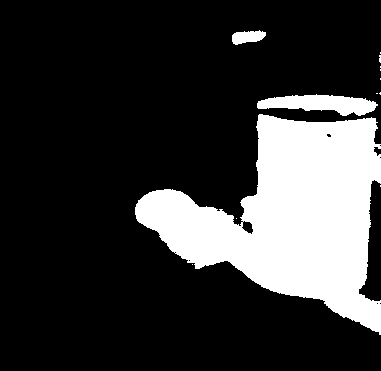}
	\end{subfigure}
	\begin{subfigure}{0.104\textwidth}
		\includegraphics[width=\textwidth]{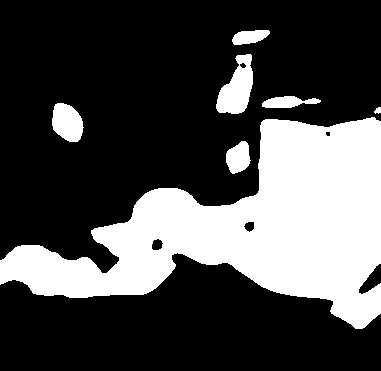}
	\end{subfigure}
	\begin{subfigure}{0.104\textwidth}
		\includegraphics[width=\textwidth]{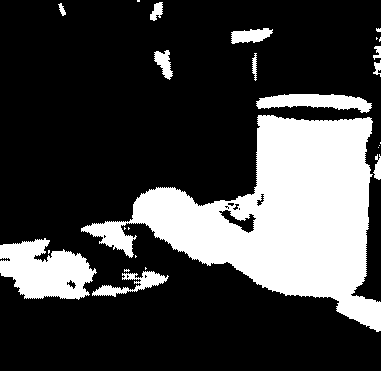}
	\end{subfigure}
	\begin{subfigure}{0.104\textwidth}
		\includegraphics[width=\textwidth]{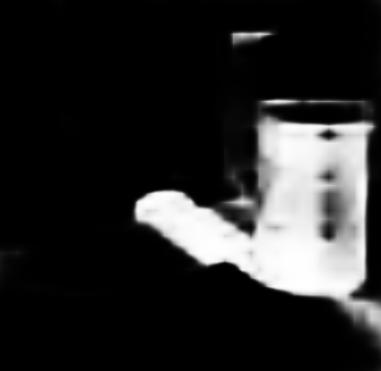}
	\end{subfigure}
	\begin{subfigure}{0.104\textwidth}
		\includegraphics[width=\textwidth]{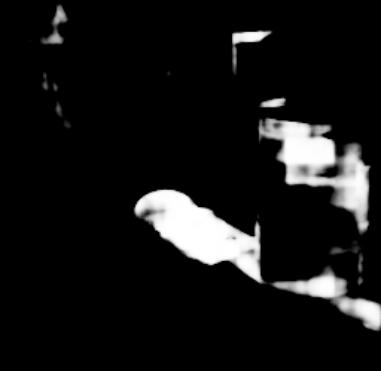}
	\end{subfigure}
	\begin{subfigure}{0.104\textwidth}
		\includegraphics[width=\textwidth]{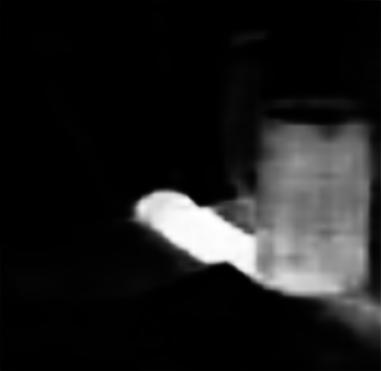}
	\end{subfigure}
	
	\ \\
	
	\vspace*{0.5mm}
	\begin{subfigure}{0.104\textwidth}
		\includegraphics[width=\textwidth]{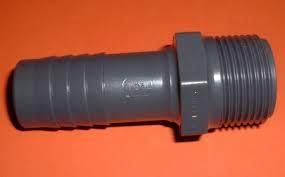}
	\end{subfigure}
	\begin{subfigure}{0.104\textwidth}
		\includegraphics[width=\textwidth]{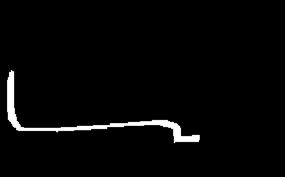}
	\end{subfigure}
	\begin{subfigure}{0.104\textwidth}
		\includegraphics[width=\textwidth]{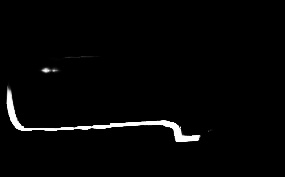}
	\end{subfigure}
	\begin{subfigure}{0.104\textwidth}
		\includegraphics[width=\textwidth]{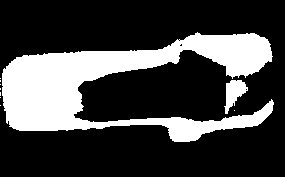}
	\end{subfigure}
	\begin{subfigure}{0.104\textwidth}
		\includegraphics[width=\textwidth]{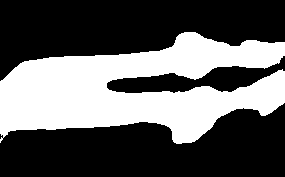}
	\end{subfigure}
	\begin{subfigure}{0.104\textwidth}
		\includegraphics[width=\textwidth]{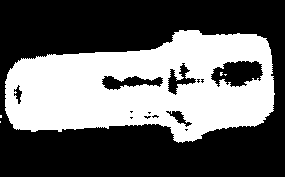}
	\end{subfigure}
	\begin{subfigure}{0.104\textwidth}
		\includegraphics[width=\textwidth]{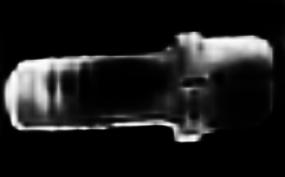}
	\end{subfigure}
	\begin{subfigure}{0.104\textwidth}
		\includegraphics[width=\textwidth]{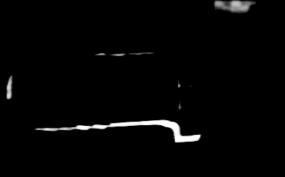}
	\end{subfigure}
	\begin{subfigure}{0.104\textwidth}
		\includegraphics[width=\textwidth]{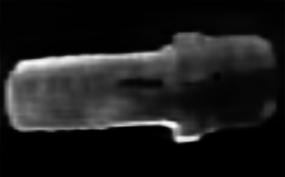}
	\end{subfigure}
	
	\ \\
	
	\vspace*{0.5mm}
	\begin{subfigure}{0.104\textwidth}
		\includegraphics[width=\textwidth]{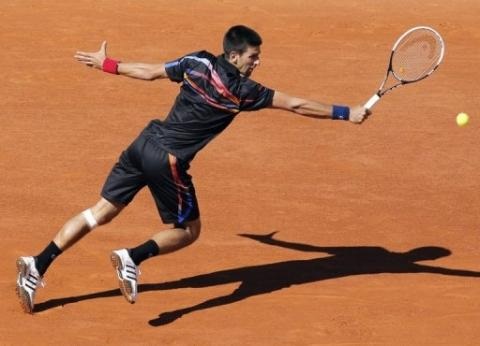}
	\end{subfigure}
	\begin{subfigure}{0.104\textwidth}
		\includegraphics[width=\textwidth]{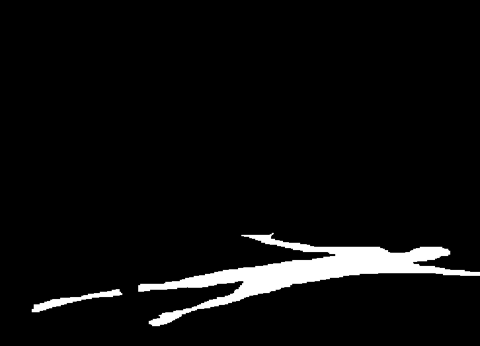}
	\end{subfigure}
	\begin{subfigure}{0.104\textwidth}
		\includegraphics[width=\textwidth]{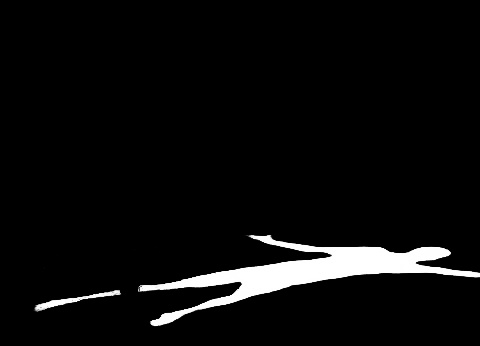}
	\end{subfigure}
	\begin{subfigure}{0.104\textwidth}
		\includegraphics[width=\textwidth]{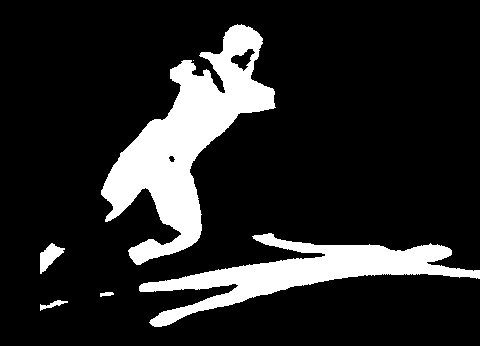}
	\end{subfigure}
	\begin{subfigure}{0.104\textwidth}
		\includegraphics[width=\textwidth]{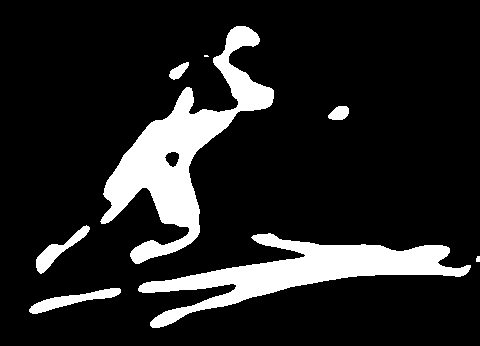}
	\end{subfigure}
	\begin{subfigure}{0.104\textwidth}
		\includegraphics[width=\textwidth]{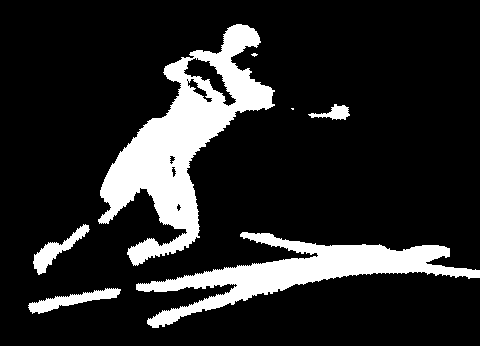}
	\end{subfigure}
	\begin{subfigure}{0.104\textwidth}
		\includegraphics[width=\textwidth]{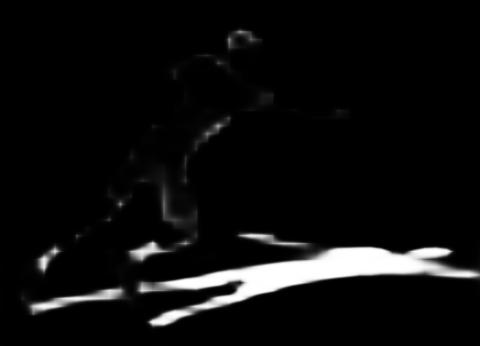}
	\end{subfigure}
	\begin{subfigure}{0.104\textwidth}
		\includegraphics[width=\textwidth]{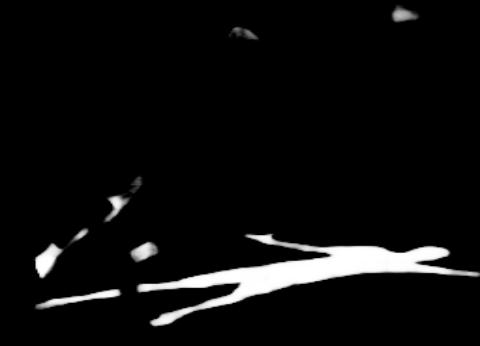}
	\end{subfigure}
	\begin{subfigure}{0.104\textwidth}
		\includegraphics[width=\textwidth]{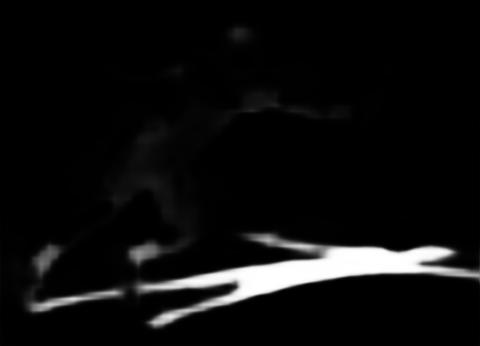}
	\end{subfigure}
	
	\ \\
	\vspace*{0.5mm}
	\begin{subfigure}{0.104\textwidth}
		\includegraphics[width=\textwidth]{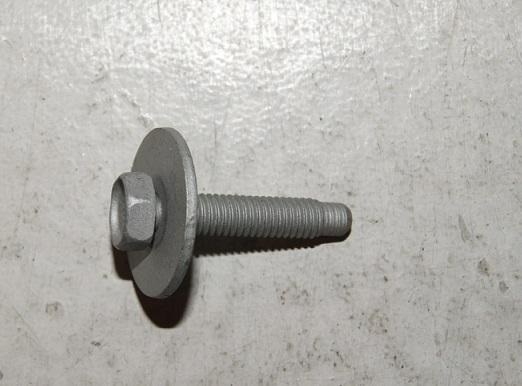}
	\end{subfigure}
	\begin{subfigure}{0.104\textwidth}
		\includegraphics[width=\textwidth]{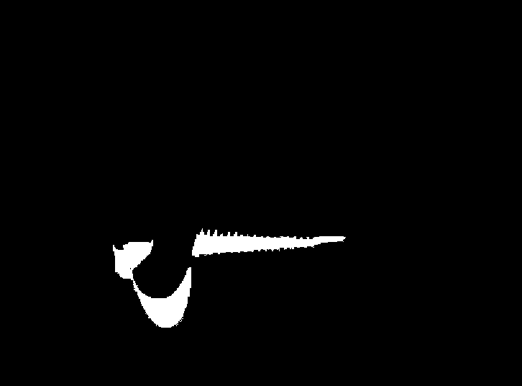}
	\end{subfigure}
	\begin{subfigure}{0.104\textwidth}
		\includegraphics[width=\textwidth]{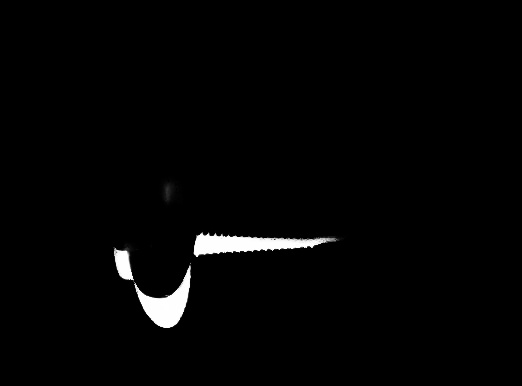}
	\end{subfigure}
	\begin{subfigure}{0.104\textwidth}
		\includegraphics[width=\textwidth]{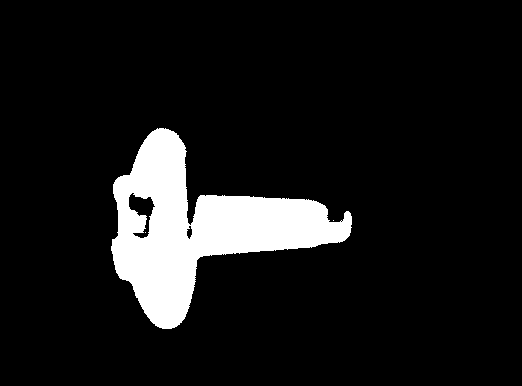}
	\end{subfigure}
	\begin{subfigure}{0.104\textwidth}
		\includegraphics[width=\textwidth]{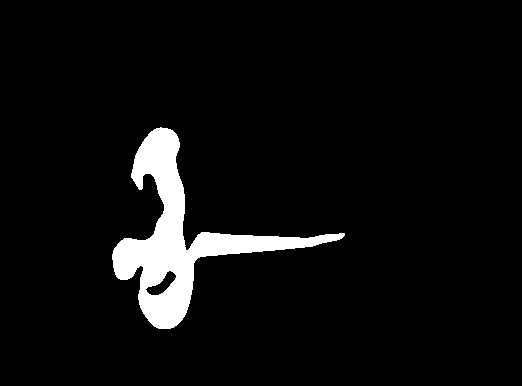}
	\end{subfigure}
	\begin{subfigure}{0.104\textwidth}
		\includegraphics[width=\textwidth]{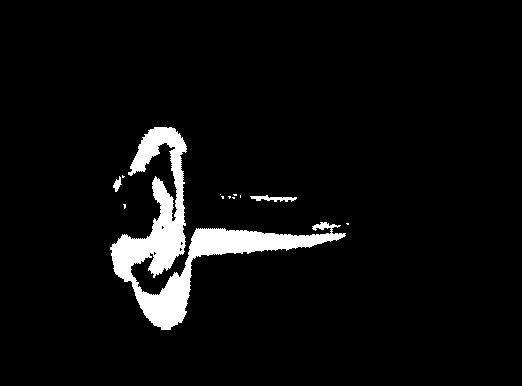}
	\end{subfigure}
	\begin{subfigure}{0.104\textwidth}
		\includegraphics[width=\textwidth]{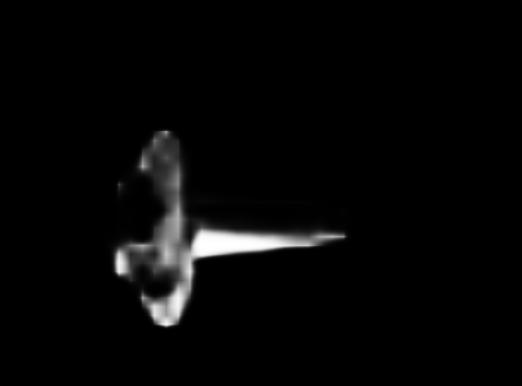}
	\end{subfigure}
	\begin{subfigure}{0.104\textwidth}
		\includegraphics[width=\textwidth]{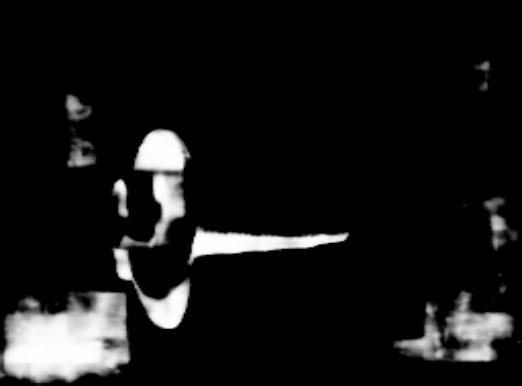}
	\end{subfigure}
	\begin{subfigure}{0.104\textwidth}
		\includegraphics[width=\textwidth]{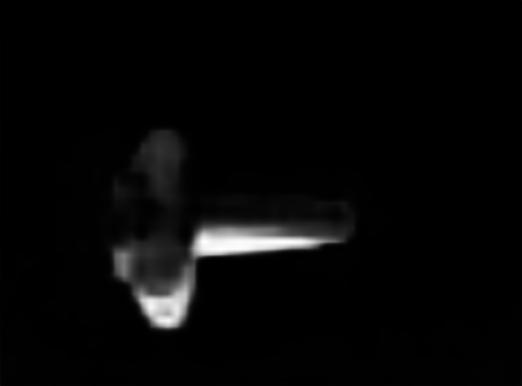}
	\end{subfigure}

	\vspace*{0.5mm}
	\begin{subfigure}{0.104\textwidth}
		\includegraphics[width=\textwidth]{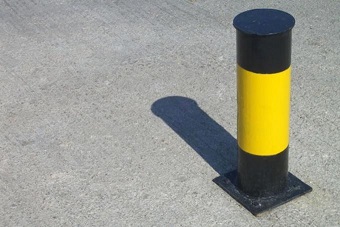}
		\vspace{-5.5mm} \caption*{{\footnotesize input}}
	\end{subfigure}
	\begin{subfigure}{0.104\textwidth}
		\includegraphics[width=\textwidth]{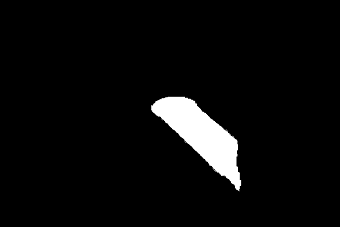}
		\vspace{-5.5mm} \caption*{{\footnotesize ground truth}} 
	\end{subfigure}
	\begin{subfigure}{0.104\textwidth}
		\includegraphics[width=\textwidth]{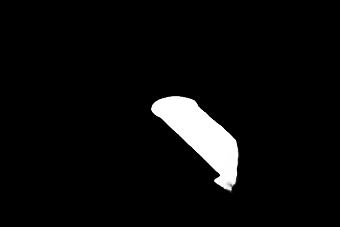}
		\vspace{-5.5mm} \caption*{{\footnotesize DSC (ours)}} 
	\end{subfigure}
	\begin{subfigure}{0.104\textwidth}
		\includegraphics[width=\textwidth]{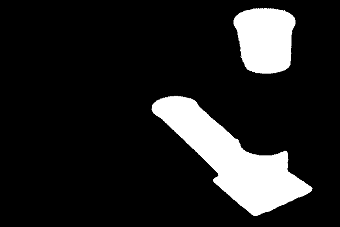}
		\vspace{-5.5mm} \caption*{{\footnotesize scGAN~\cite{nguyen2017shadow}}}
	\end{subfigure}
	\begin{subfigure}{0.104\textwidth}
		\includegraphics[width=\textwidth]{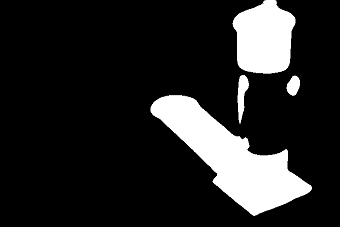}
		\vspace{-5.5mm} \caption*{{\footnotesize stkd'-CNN~\cite{vicente2016large}}}
	\end{subfigure}
	\begin{subfigure}{0.104\textwidth}
		\includegraphics[width=\textwidth]{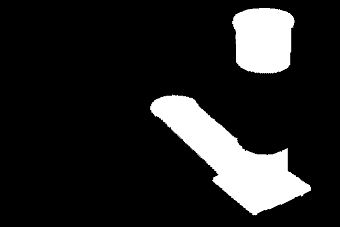}
		\vspace{-5.5mm} \caption*{{\footnotesize patd'-CNN~\cite{hosseinzadeh2017fast}}}
	\end{subfigure}
	\begin{subfigure}{0.104\textwidth}
		\includegraphics[width=\textwidth]{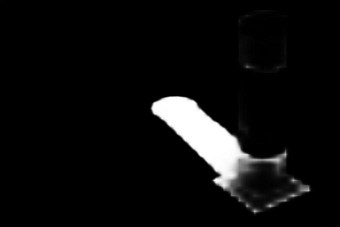}
		\vspace{-5.5mm} \caption*{{\footnotesize SRM~\cite{wang2017stagewise}}}
	\end{subfigure}
	\begin{subfigure}{0.104\textwidth}
		\includegraphics[width=\textwidth]{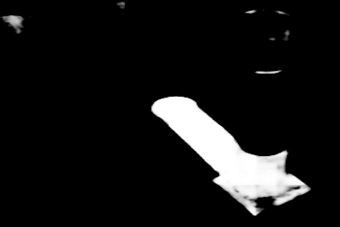}
		\vspace{-5.5mm} \caption*{{\footnotesize Amulet~\cite{zhang2017amulet}}}
	\end{subfigure}
	\begin{subfigure}{0.104\textwidth}
		\includegraphics[width=\textwidth]{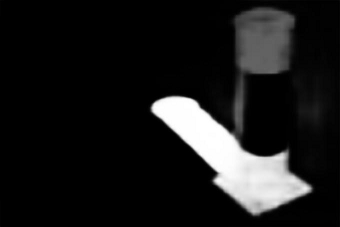}
		\vspace{-5.5mm} \caption*{\hspace*{-0.7mm}{\footnotesize PSPNet~\cite{Zhao_2017_CVPR}}}
	\end{subfigure}

	\vspace*{-1mm}
	\caption{More visual comparison results on shadow detection (continue from Figure~\ref{fig:comparison_real_photos}).}
	\label{fig:comparison_real_photos2}
	\vspace*{-2mm}
\end{figure*}


\begin{table}[!t]
	\begin{center}
		\caption{Comparing our method (DSC) with recent methods
			for shadow detection (scGAN~\cite{nguyen2017shadow}, stacked-CNN~\cite{vicente2016large}, patched-CNN~\cite{hosseinzadeh2017fast}, and Unary-Pairwise~\cite{guo2011single}),
			for saliency detection (SRM~\cite{wang2017stagewise} and Amulet~\cite{zhang2017amulet}),
			and
			for semantic image segmentation (PSPNet~\cite{Zhao_2017_CVPR}).
			Note that the results on the UCF dataset are different from~\cite{hu2018direction} due to the different test splits.}
		\vspace*{-1mm}
		\label{table:state-of-the-art}
		\begin{tabular}{c|c|c|c|c}
			&
			\multicolumn{2}{c|}{SBU~\cite{vicente2016noisy, vicente2016large}} &
			\multicolumn{2}{c}{UCF~\cite{zhu2010learning}}
			\\
			\hline
			method & accuracy & BER & accuracy & BER
			\\
			\hline
			\hline
			\textbf{DSC (ours)} & \textbf{0.97} & \textbf{5.59} & \textbf{0.95} & \textbf{10.38}
			\\
			\hline
			\hline
			scGAN~\cite{nguyen2017shadow} & 0.90 & 9.10 & 0.86 & 11.50
			\\
			stacked-CNN~\cite{vicente2016large} & 0.88 & 11.00 & 0.84 & 13.00
			\\
			patched-CNN~\cite{hosseinzadeh2017fast} & 0.88 & 11.56  &-&-
			\\
			Unary-Pairwise~\cite{guo2011single} & 0.86 & 25.03 &-&-
			\\
			\hline
			\hline
			SRM~\cite{wang2017stagewise} & 0.96 & 7.25 & 0.93 & 11.27
			\\
			Amulet~\cite{zhang2017amulet} & 0.93 & 15.13 & 0.92 & 19.62
			\\
			\hline
			\hline
			PSPNet~\cite{Zhao_2017_CVPR} & 0.95 & 8.57 & 0.94 & 10.94
			\\
			\hline
		\end{tabular}
	\end{center}
	\vspace*{-1mm}
\end{table}


\subsection{Comparison with the State-of-the-art}

\vspace*{2mm}
\noindent
{\bf Comparison with recent shadow detection methods.} \
We compare our method with four recent shadow detection methods: scGAN~\cite{nguyen2017shadow}, stacked-CNN~\cite{vicente2016large}, patched-CNN~\cite{hosseinzadeh2017fast}, and Unary-Pairwise~\cite{guo2011single}.
The first three are network-based methods, while the last one is based on hand-crafted features.
For a fair comparison, we obtain their shadow detection results either by directly taking the results from the authors or by generating the results using the implementations provided by the authors using the recommended parameter setting.

Table~\ref{table:state-of-the-art} reports the comparison results, showing that our method performs favorably against all the other methods for both benchmark datasets on both accuracy and BER.
Our shadow detection network is trained using the SBU training set~\cite{vicente2016noisy, vicente2016large}, but still outperforms others on the UCF dataset, thus showing its generalization capability.
Further, we show visual comparison results in Figures~\ref{fig:comparison_real_photos} and~\ref{fig:comparison_real_photos2}, which show various challenging cases, e.g., a light shadow next to a dark shadow, shadows around complex backgrounds, and black objects around shadows.
Without understanding the global image semantics, it is hard to locate these shadows, and the non-shadow regions could be easily misrecognized as shadows.
From the results, we can see that our method can effectively locate shadows and avoid false positives compared to others, e.g., for black objects misrecognized by others as shadows, our method could still recognize them as non-shadows.

\begin{table}  [tp]
	\begin{center}
		\caption{Component analysis.
			We train three networks using the SBU training set and test them using the SBU testing set~\cite{vicente2016noisy,vicente2016large}:
			``basic'' denotes the architecture shown in Figure~\ref{fig:context} but without all the DSC modules;
			``basic+context'' denotes the ``basic'' network with spatial context but not direction-aware spatial context; and
			``DSC" is the overall network shown in Figure~\ref{fig:context}.}
		\label{table:ablation}
		\begin{tabular}{c|c|c}
			\hline
			network & BER & improvement \\ 
			\hline
			basic &  6.55 & -\\ 
			
			basic+context & 6.23 & 4.89\% \\ 
			
			DSC & \textbf{5.59} & 10.27\% \\ 
			
			\hline
			
		\end{tabular} 
	\end{center}
\end{table}

\begin{figure} [tp]
	\centering
	\includegraphics[width=0.98\linewidth]{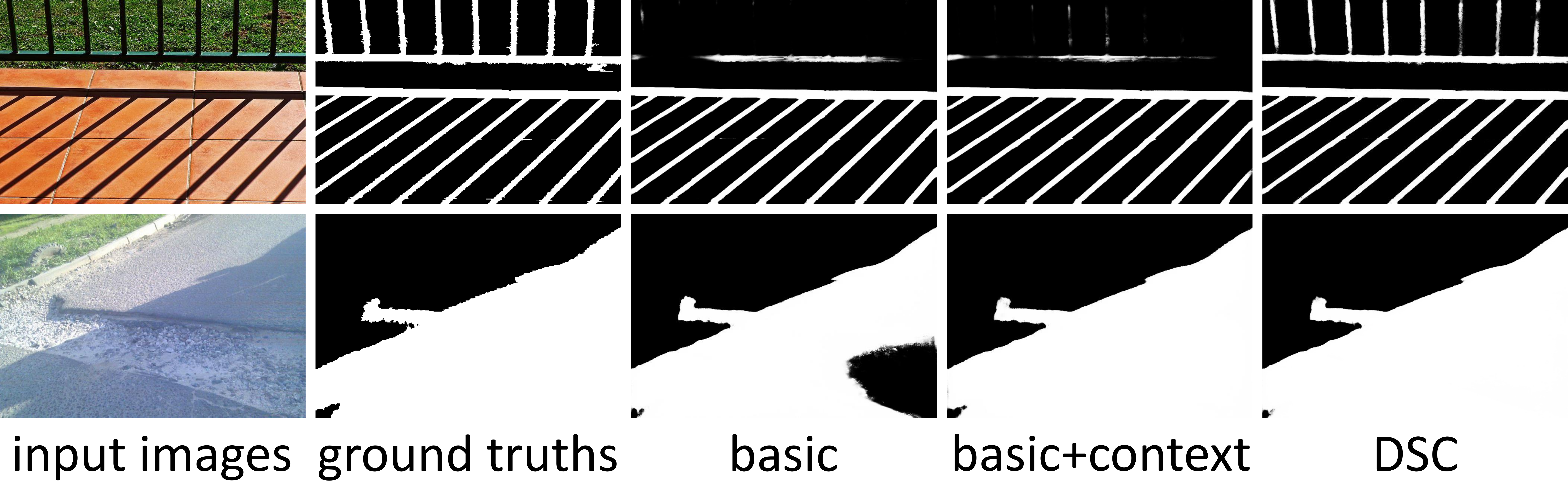}
	\caption{Visual comparison results of component analysis.}
	\label{fig:component_analysis}
	\vspace{-1mm}
\end{figure}


\noindent
{\bf Comparison with recent saliency detection \& semantic segmentation methods.} \
Deep networks for saliency detection and semantic image segmentation may also be used for shadow detection by training the networks using datasets of annotated shadows.
Thus, we perform another experiment using two recent deep models for saliency detection (SRM~\cite{wang2017stagewise} and Amulet~\cite{zhang2017amulet}) and a recent deep model for semantic image segmentation (PSPNet~\cite{Zhao_2017_CVPR}).

For a fair comparison, we use the implementations provided by the authors, adopt the parameters trained on ImageNet~\cite{deng2009imagenet} for a classification task to initialize their models, re-train the models on the SBU training set for shadow detection, and adjust the training parameters to obtain the best shadow detection results.
The last three rows in Table~\ref{table:state-of-the-art} report the comparison results on the accuracy and BER metrics.
Although these methods achieve good results for both metrics, our method still performs favorably against them for both benchmark datasets.
Please also refer to the last three columns in Figures~\ref{fig:comparison_real_photos} and~\ref{fig:comparison_real_photos2} for visual comparison results.


\begin{table}[tp]
	\begin{center}
		\caption{DSC architecture analysis.
			By varying the parameters in the DSC architecture (see the $2^{nd}$ and $3^{rd}$ columns below), we can produce slightly different overall networks and explore their performance (see the last column).}
		\label{table:rnn}
		\begin{tabular}{c|c|c}
			\hline
			number of rounds & shared $\mathbf{W}? $ & BER  \\
			\hline
			1 & - & 5.85
			\\
			2 & Yes & \textbf{5.59}
			\\
			3 & Yes & 5.85
			\\
			\hline
			2 & No & 6.02
			\\
			3 & No & 5.93
			\\
			\hline			
		\end{tabular} 
	\end{center}
	\vspace{-4mm}
\end{table}



\subsection{Evaluation on the Network Design}


\noindent
{\bf Component analysis.} \
We perform an experiment to evaluate the effectiveness of the DSC module design.
Here, we use the SBU dataset and consider two baseline networks.
The first baseline (denoted as ``basic'') is a network constructed by removing all the DSC modules from the overall network shown in Figure~\ref{fig:arc}.
The second baseline (denoted as ``basic+context") considers the spatial context but ignores the direction-aware attention weights.
Compared with the first baseline, the second baseline includes all the DSC modules, but removes the direction-aware attention mechanism in the DSC modules, i.e., without computing $\mathbf{W}$ and directly concatenating the context features;
see Figure~\ref{fig:context}.
This is equivalent to setting all attention weights $\mathbf{W}$ to one; 
see supplementary material for the architecture of the two baselines.

Table~\ref{table:ablation} reports the comparison results, showing that our basic network with multi-scale features and the weighed cross entropy loss can produce better results.
Moreover, considering the spatial context and DSC features can lead to further
improvement; see also Figure~\ref{fig:component_analysis} for the visual comparison results.

\vspace*{2mm}
\noindent
{\bf DSC architecture analysis.} \
We encountered two questions when designing the network structure with the DSC modules:
(i) how many rounds of recurrent translations in the spatial RNN; and 
(ii) whether to share the attention weights or to use separate attention weights in different rounds of recurrent translations.

We modify our network for these two parameters and produce the comparison results shown in Table~\ref{table:rnn}.
From the results, we can see that having two rounds of recurrent translations and sharing the attention weights in both rounds produce the best result.
When there is only one round of recurrent translations, the global image context cannot well propagate over the spatial domain, so the amount of information exchange is insufficient for learning the shadows, while having three rounds of recurrent translations with separate copies of attention weights will introduce excessive parameters that make the network hard to be trained.

\begin{figure} [tp]
	\centering
	\includegraphics[width=0.98\linewidth]{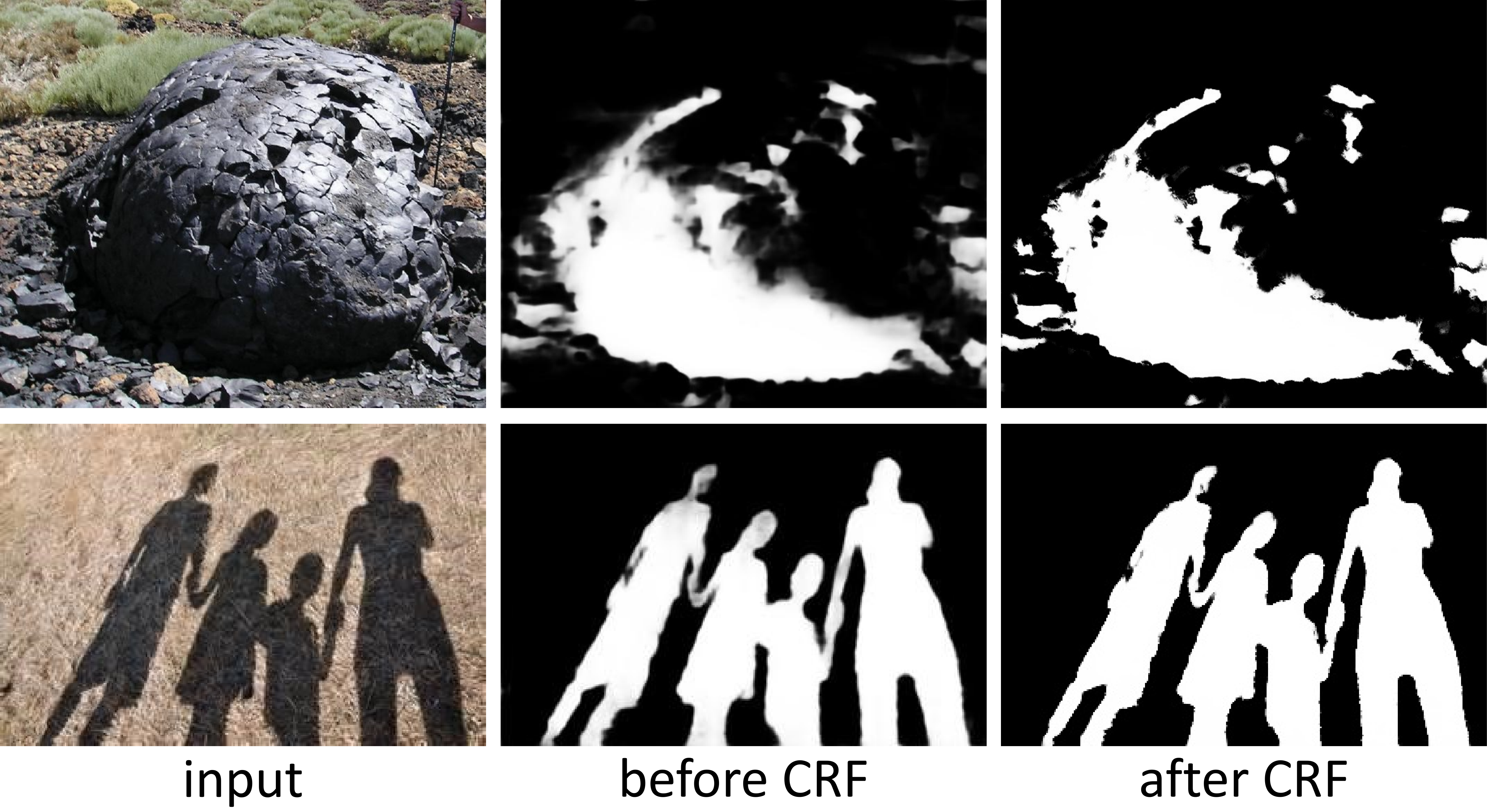}
	\vspace{-2mm}
	\caption{Effectiveness of CRF~\cite{krahenbuhl2011efficient}.}
	\label{fig:CRF}
\end{figure}


\vspace*{2mm}
\noindent
{\bf Feature extraction network analysis.} \
We also evaluate the feature extraction network shown in Figure~\ref{fig:arc}.
Here, we use the deeper network, ResNet-101~\cite{he2016deep} with 101 layers, to replace the VGG network, which has only 16 layers.
In the ResNet-101, it has multiple layers that produce the output feature maps of the same scale, and we choose the output of the last layer for each scale, i.e., res2c, res3b3, res4b22, and res5c, to produce the DSC features,
since the last layer should have the strongest features. The other network parts and parameter settings are kept unchanged.
	

The resulting BER values are 5.59 and 5.73 for the VGG network and ResNet-101, respectively, showing that they have similar performance, with the VGG performing slightly better.
The deeper network (ResNet-101) allows the generation of stronger semantic features, but loses the details due to the small-sized feature maps when accounting for the limited GPU memory.

\vspace*{2mm}
\noindent
{\bf Effectiveness of CRF.} \
Next, we evaluate the effectiveness of CRF~\cite{krahenbuhl2011efficient} as a post-processing step. The BER values for ``before CRF'' and ``after CRF'' are 5.68 and 5.59, respectively, showing that CRF helps improve the shadow detection result; see also Figure~\ref{fig:CRF} for the visual comparison results.

\begin{figure} [tp]
	\centering
	\includegraphics[width=0.98\linewidth]{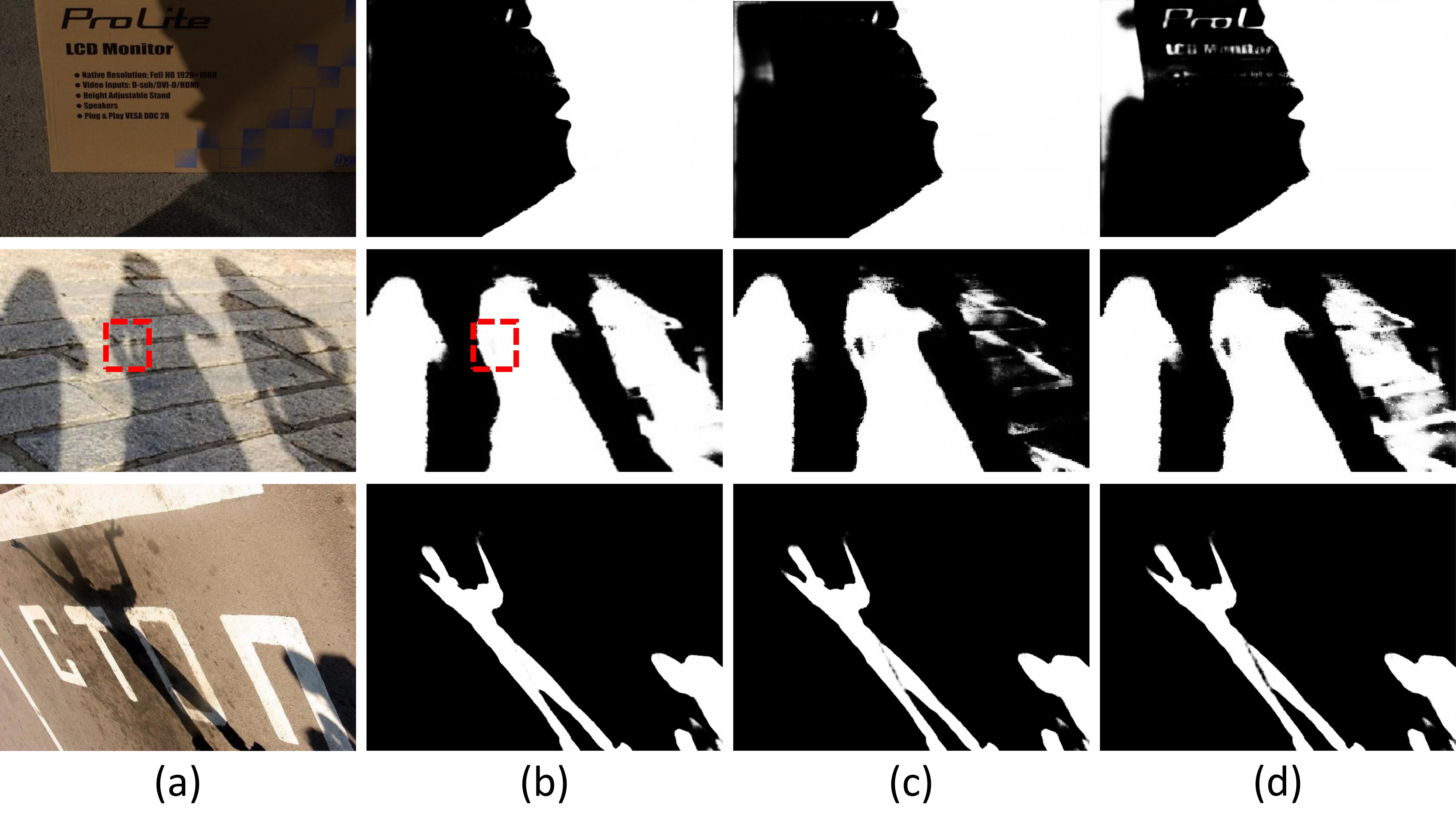}
	\vspace{-2mm}
	\caption{Effectiveness of the spatial context on shadow detection.
(a) input images; (b) DSC results; (c) using only the spatial context in the vertical direction; and (d) using only the spatial context in the horizontal direction.}
	\label{fig:results_direction}
\end{figure}

\vspace*{2mm}
\noindent
{\bf DSC feature analysis.} \
Lastly, we show how the spatial context features in different directions affect the shadow detection performance.
Here, we ignore the spatial context in horizontal directions when detecting shadows by setting the associated attention weights in the left and right directions as zero; see Figure~\ref{fig:context}.
Similarly, we ignore the spatial context in vertical directions by setting the associated attention weights in the up and down directions as zero.
Figure~\ref{fig:results_direction} presents some results, showing that when using only the spatial context in vertical/horizontal directions, we could misrecognize black regions as shadows and miss some unobvious shadows.
However, our network may fail to recognize tiny shadow/non-shadow regions, since the surrounding information aggregated by our DSC module may cover the original features, thus omitting those tiny regions; see the second row in Figure~\ref{fig:results_direction} and the supplementary material for more results.

\if
\phil{save space by removing table 4 and just talk about the numbers here?}
\textcolor{blue}{Table~\ref{table:basic_network} reports the results, showing that the performances of different feature extraction network are similar. The deeper network provides stronger semantic features, but it loses detail information due to smaller sizes of feature maps considering the GPU memory limitation. 
On the other hand, our DSC module is effective on various architectures of feature extract networks. 
}
\fi

\begin{figure} [tbp]
	\centering
	\includegraphics[width=0.95\linewidth]{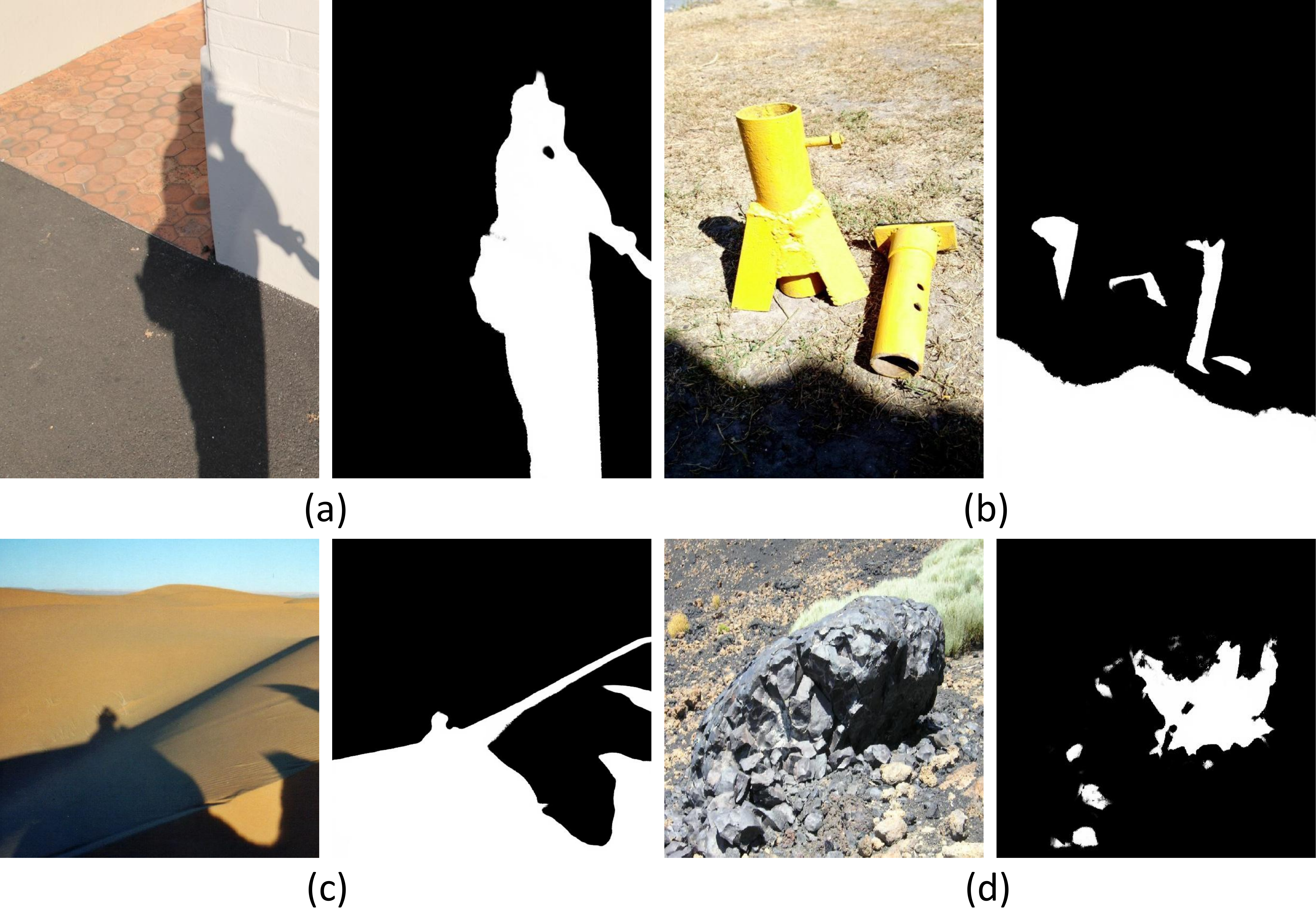}
	\caption{More shadow detection results produced by our method.}
	\label{fig:more_results}
	\vspace{-4mm}
\end{figure}

\begin{figure*}[tp]
	\centering

	\vspace*{0.5mm}
	\begin{subfigure}{0.138\textwidth} 
		\includegraphics[width=\textwidth]{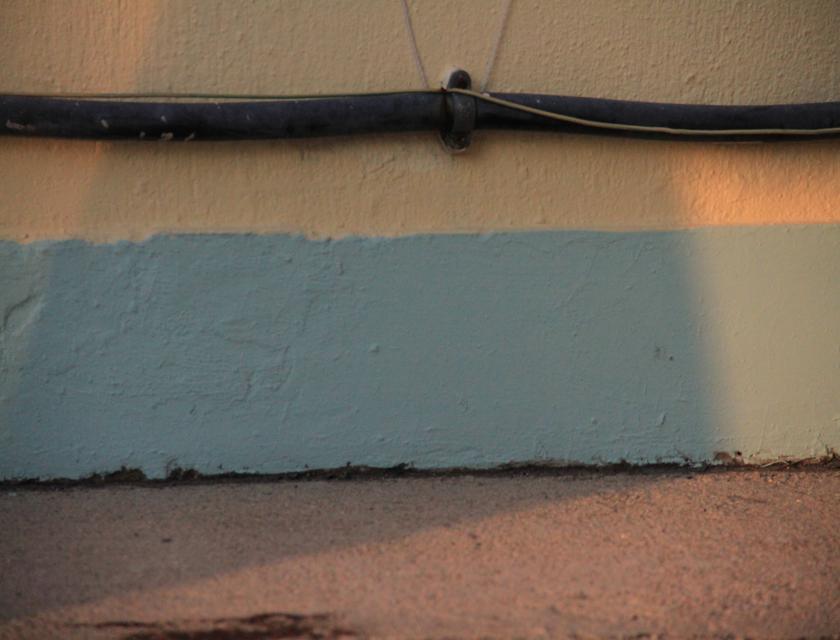}
	\end{subfigure}
	\begin{subfigure}{0.138\textwidth}
		\includegraphics[width=\textwidth]{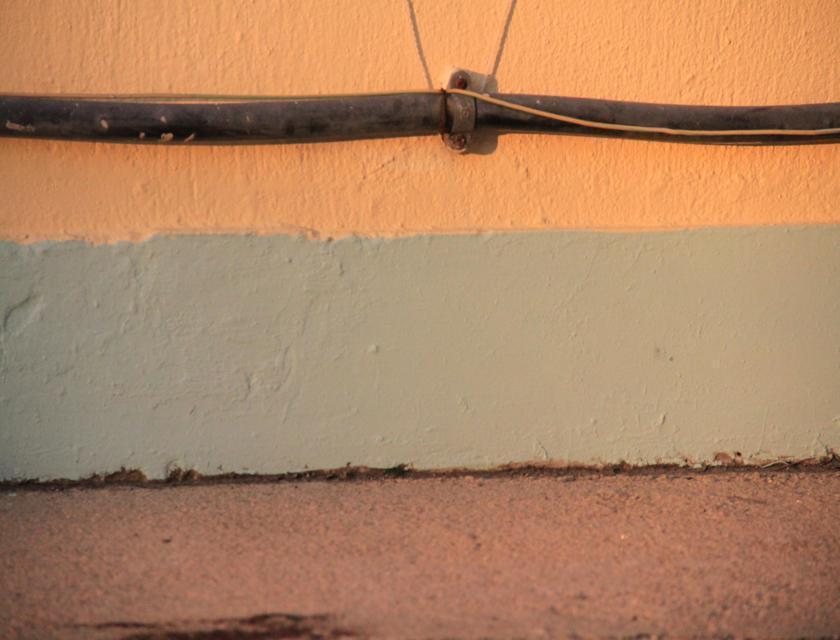}
	\end{subfigure}
    \begin{subfigure}{0.138\textwidth}
    	\includegraphics[width=\textwidth]{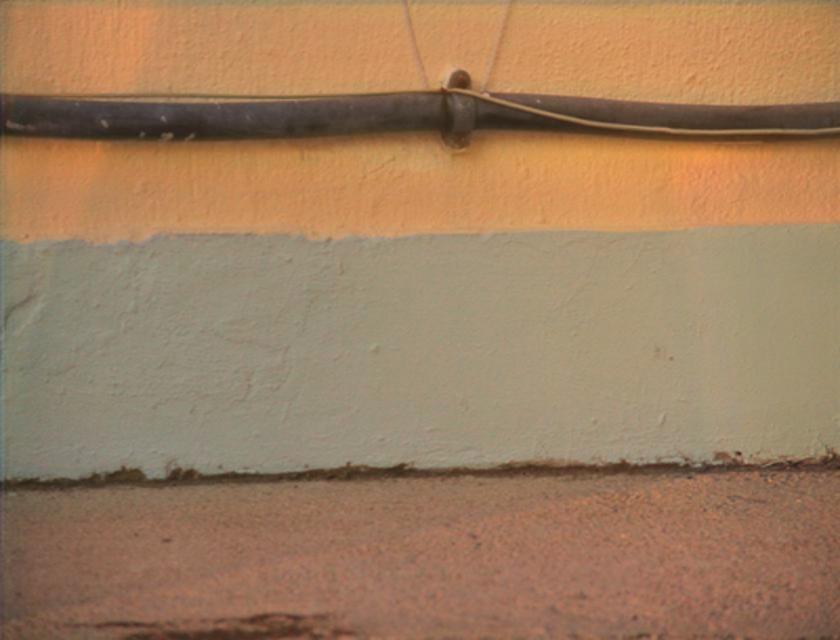}
    \end{subfigure}
	\begin{subfigure}{0.138\textwidth}
		\includegraphics[width=\textwidth]{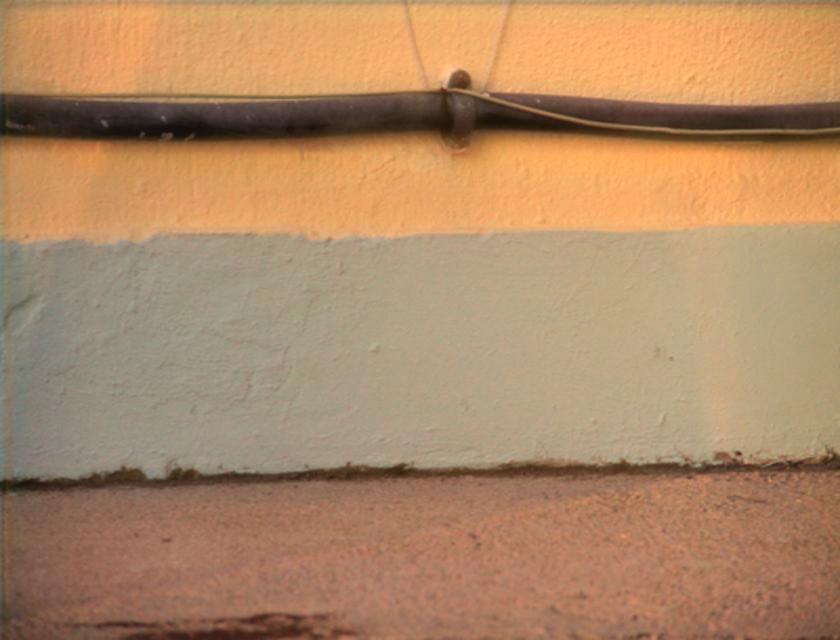}
	\end{subfigure}
	\begin{subfigure}{0.138\textwidth}
		\includegraphics[width=\textwidth]{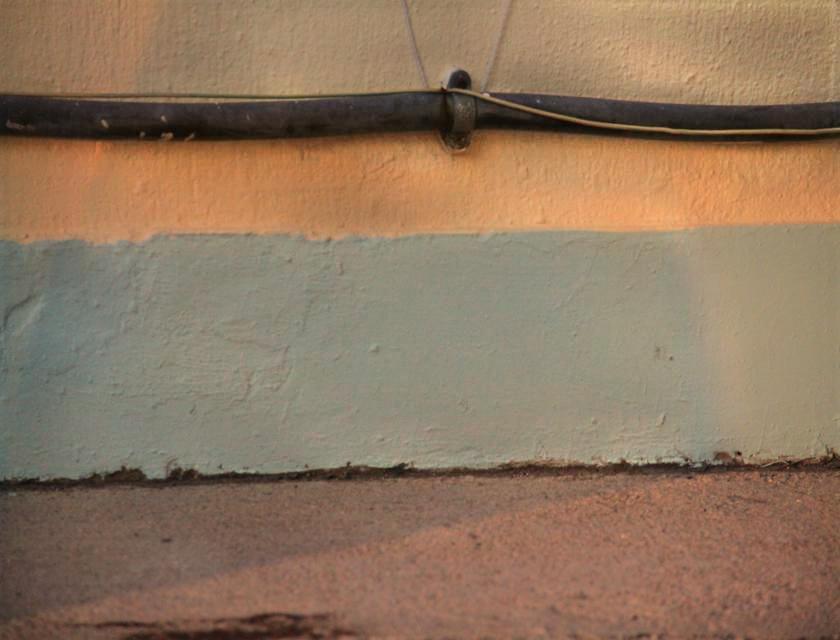}
	\end{subfigure}
	\begin{subfigure}{0.138\textwidth}
		\includegraphics[width=\textwidth]{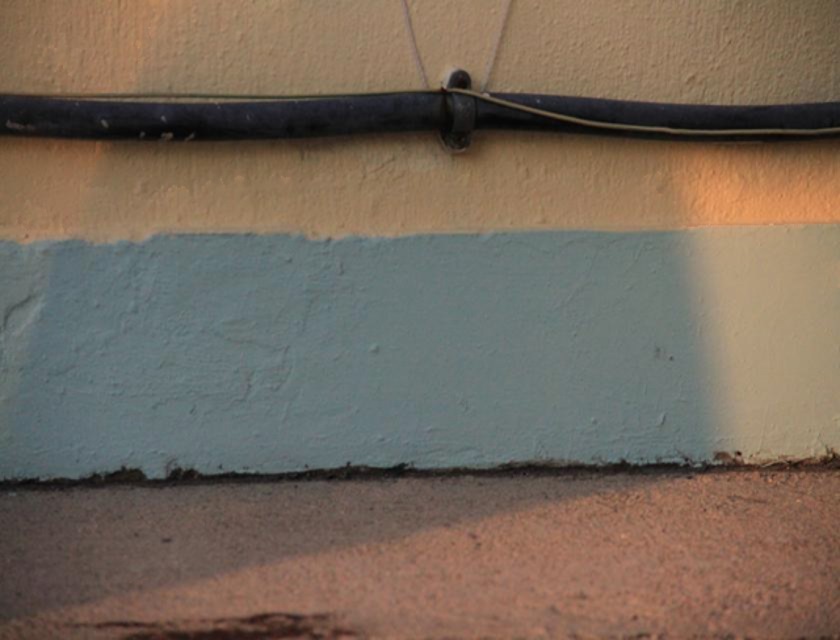}
	\end{subfigure}
	\begin{subfigure}{0.138\textwidth}
		\includegraphics[width=\textwidth]{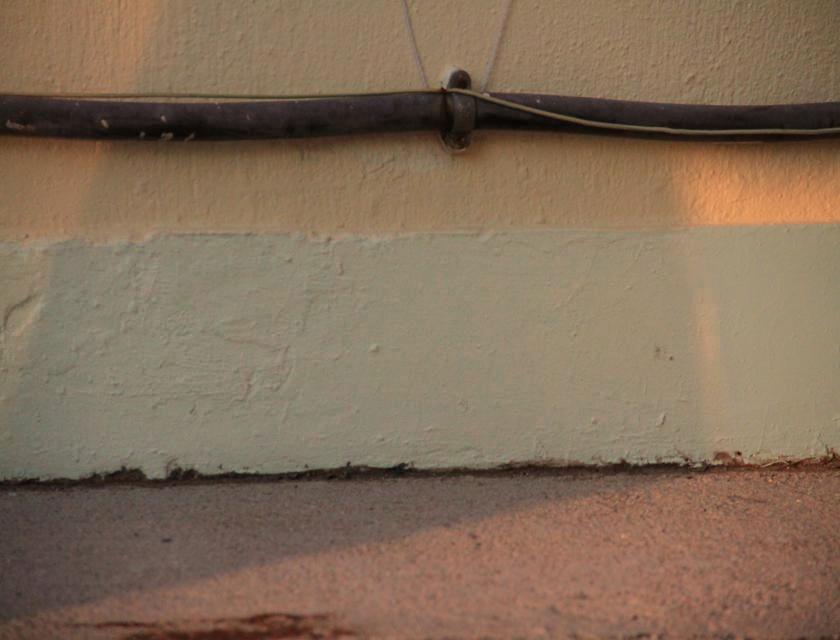}
	\end{subfigure}
	
	\vspace*{0.5mm}
	\begin{subfigure}{0.138\textwidth} 
		\includegraphics[width=\textwidth]{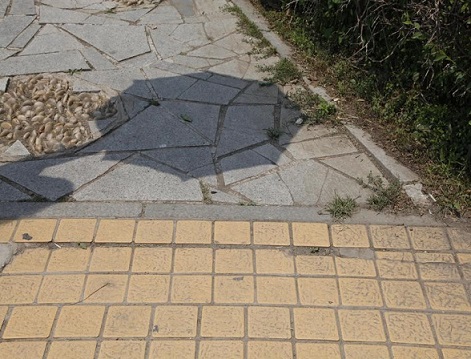}
	\end{subfigure}
	\begin{subfigure}{0.138\textwidth}
		\includegraphics[width=\textwidth]{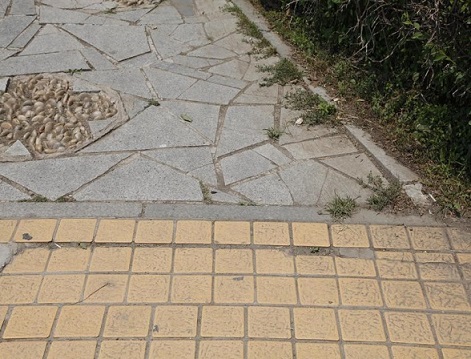}
	\end{subfigure}
    \begin{subfigure}{0.138\textwidth}
    	\includegraphics[width=\textwidth]{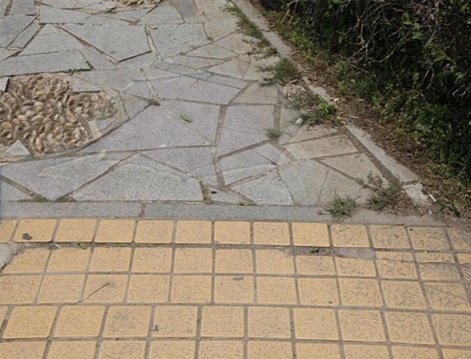}
    \end{subfigure}
	\begin{subfigure}{0.138\textwidth}
		\includegraphics[width=\textwidth]{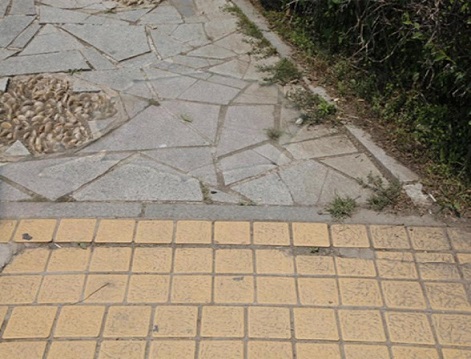}
	\end{subfigure}
	\begin{subfigure}{0.138\textwidth}
		\includegraphics[width=\textwidth]{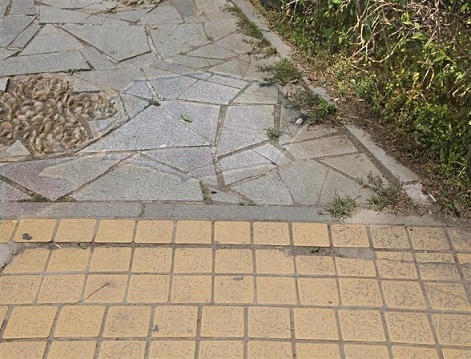}
	\end{subfigure}
	\begin{subfigure}{0.138\textwidth}
		\includegraphics[width=\textwidth]{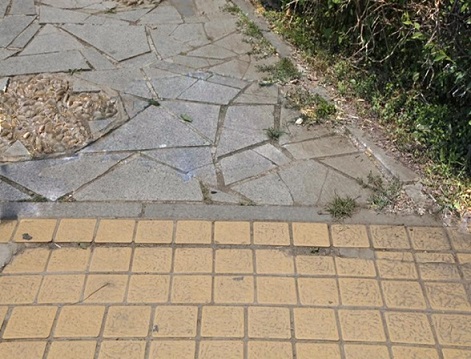}
	\end{subfigure}
	\begin{subfigure}{0.138\textwidth}
		\includegraphics[width=\textwidth]{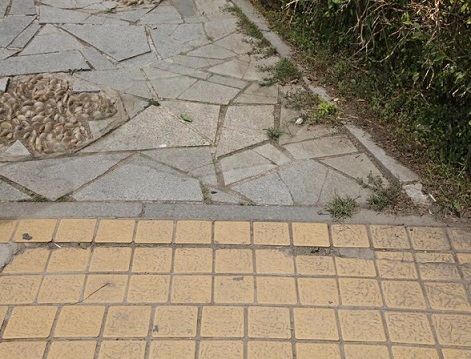}
	\end{subfigure}
	\ \\
	
	\vspace*{0.5mm}
	\begin{subfigure}{0.138\textwidth} 
		\includegraphics[width=\textwidth]{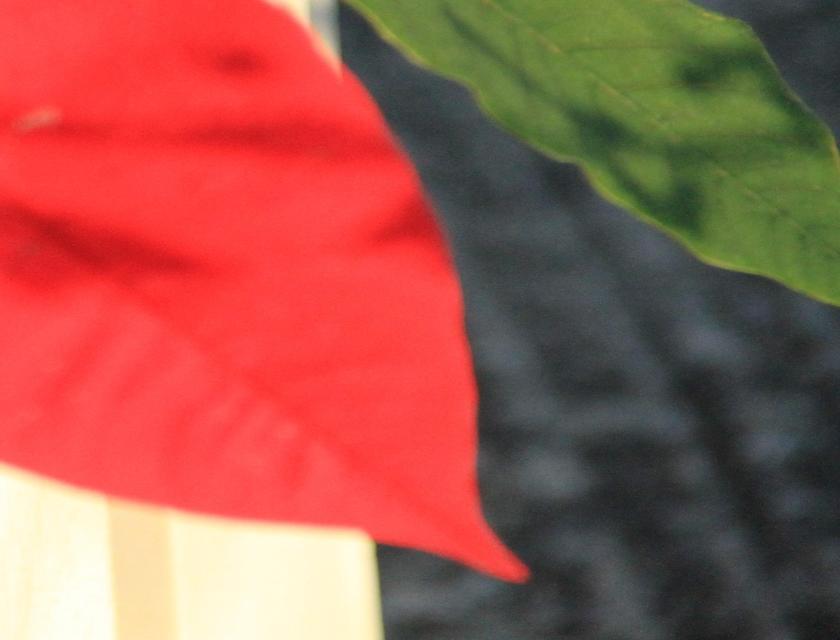}
	\end{subfigure}
	\begin{subfigure}{0.138\textwidth}
		\includegraphics[width=\textwidth]{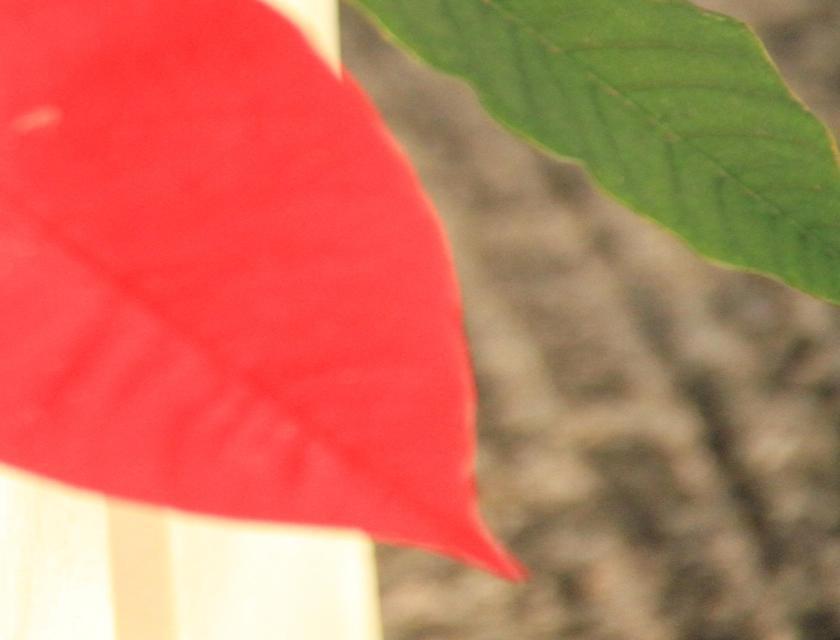}
	\end{subfigure}
    \begin{subfigure}{0.138\textwidth}
    	\includegraphics[width=\textwidth]{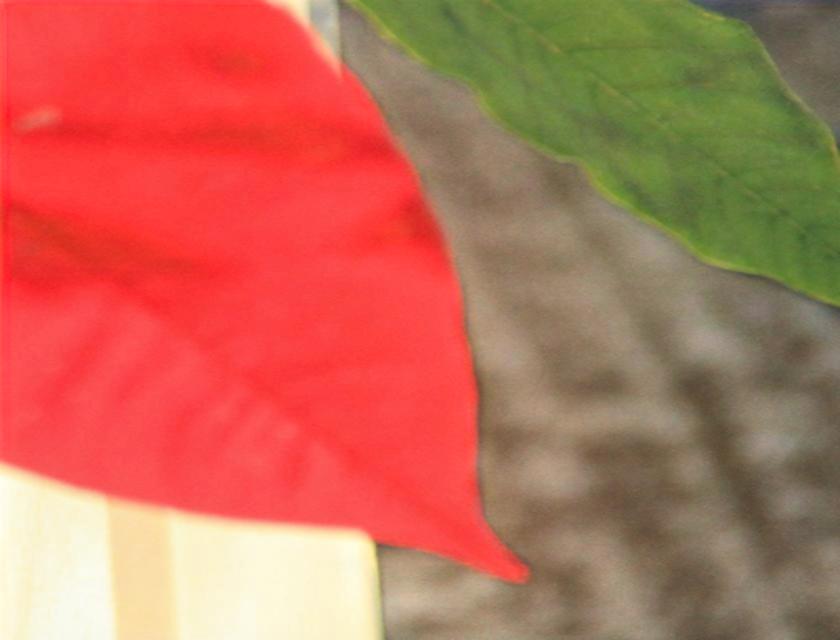}
    \end{subfigure}
	\begin{subfigure}{0.138\textwidth}
		\includegraphics[width=\textwidth]{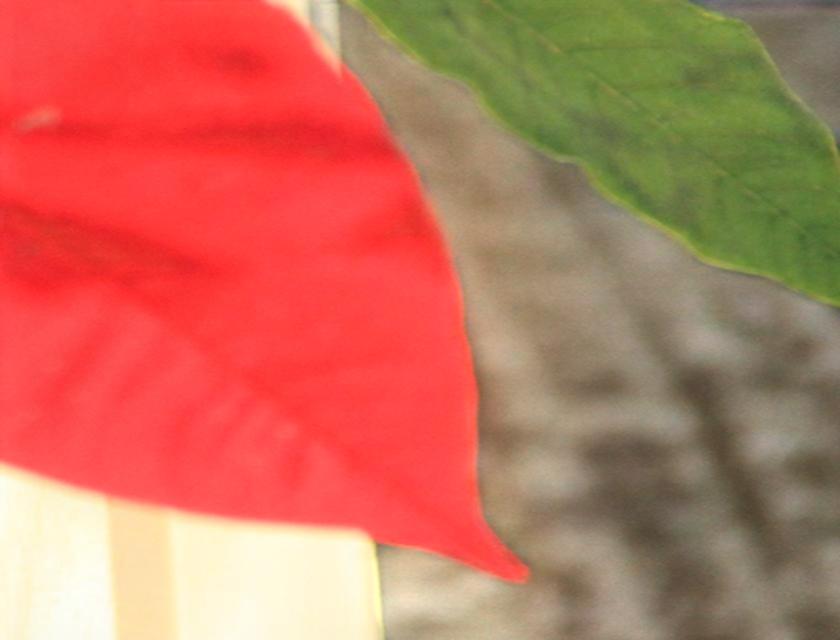}
	\end{subfigure}
	\begin{subfigure}{0.138\textwidth}
		\includegraphics[width=\textwidth]{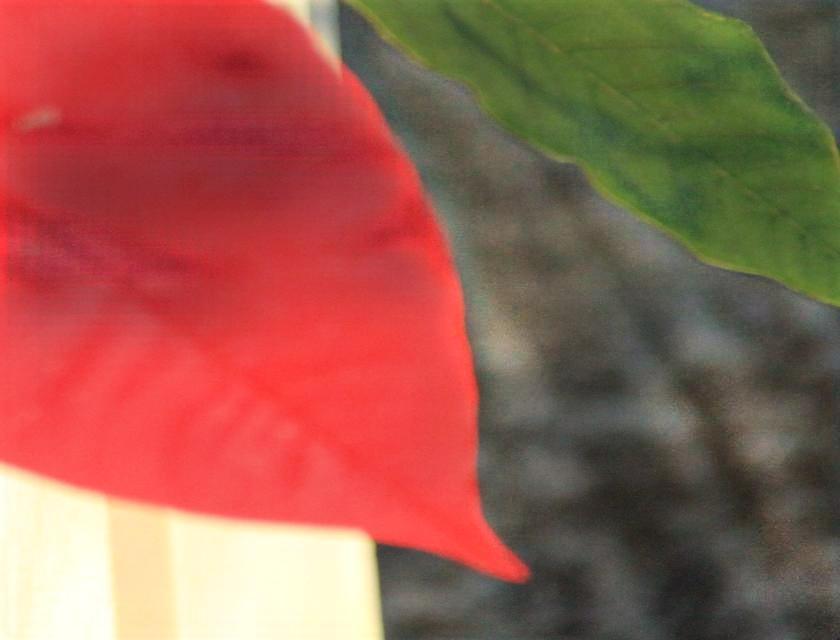}
	\end{subfigure}
	\begin{subfigure}{0.138\textwidth}
		\includegraphics[width=\textwidth]{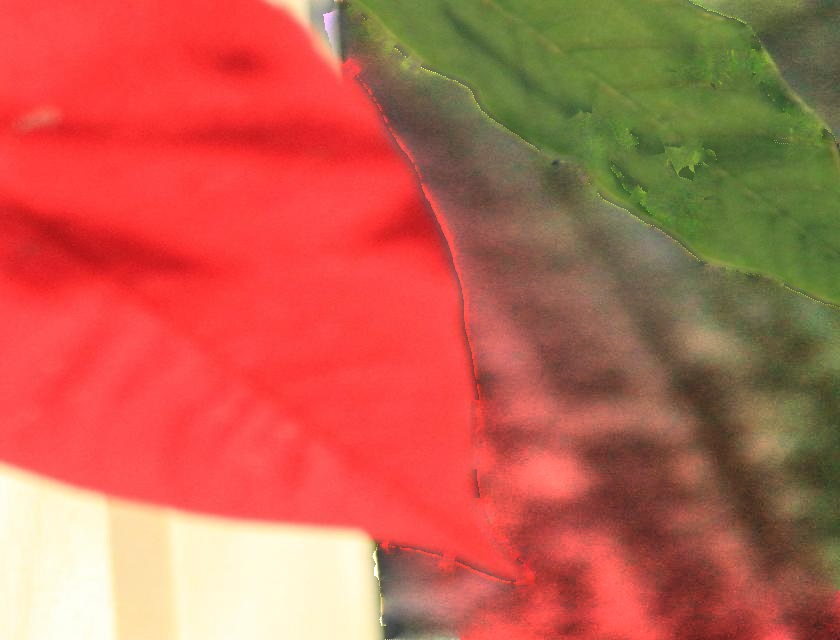}
	\end{subfigure}
	\begin{subfigure}{0.138\textwidth}
		\includegraphics[width=\textwidth]{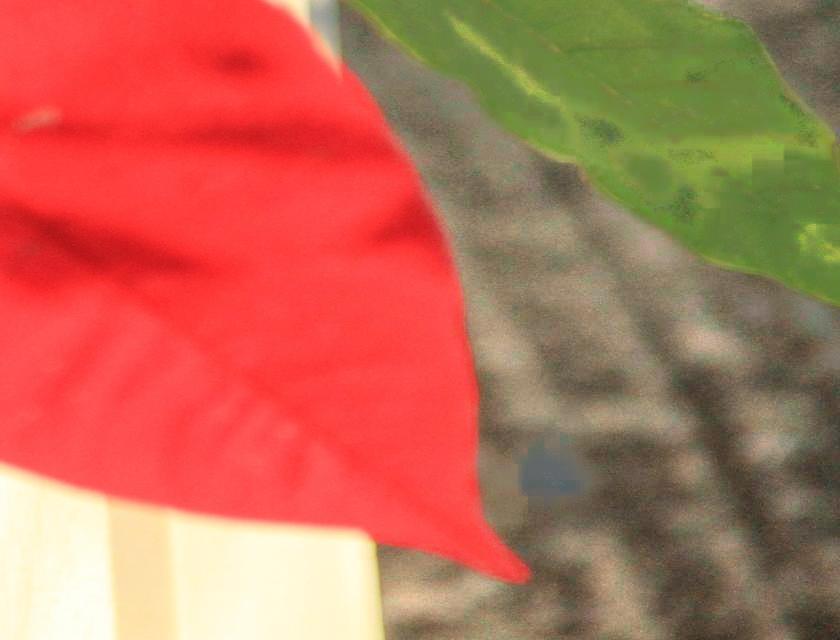}
	\end{subfigure}
	\ \\	
	
	\vspace*{0.5mm}
	\begin{subfigure}{0.138\textwidth}
		\includegraphics[width=\textwidth]{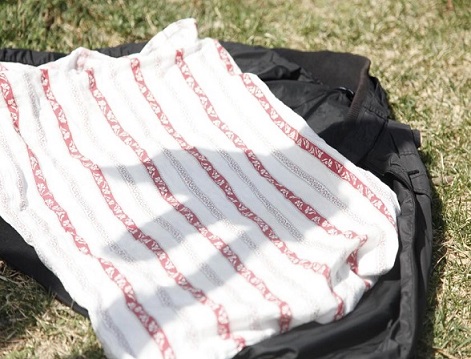}
		\vspace{-5.5mm} \caption*{{\footnotesize input}}
	\end{subfigure}
	\begin{subfigure}{0.138\textwidth}
		\includegraphics[width=\textwidth]{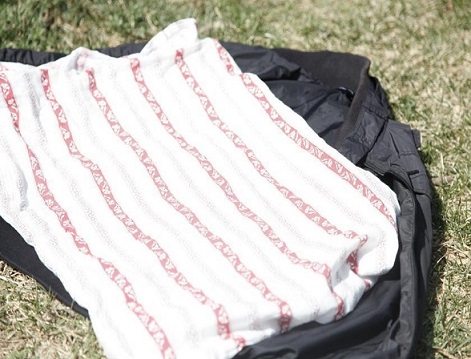}
		\vspace{-5.5mm} \caption*{{\footnotesize ground truth}} 
	\end{subfigure}
    \begin{subfigure}{0.138\textwidth}
    	\includegraphics[width=\textwidth]{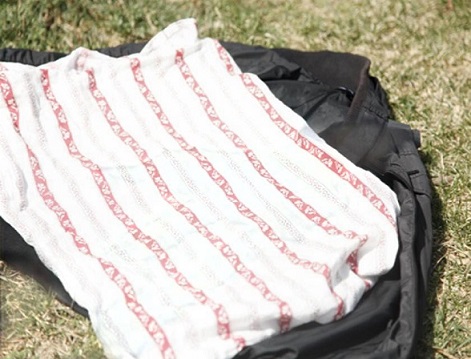}
    	\vspace{-5.5mm} \caption*{{\footnotesize DSC+ (ours)}} 
    \end{subfigure}
	\begin{subfigure}{0.138\textwidth}
		\includegraphics[width=\textwidth]{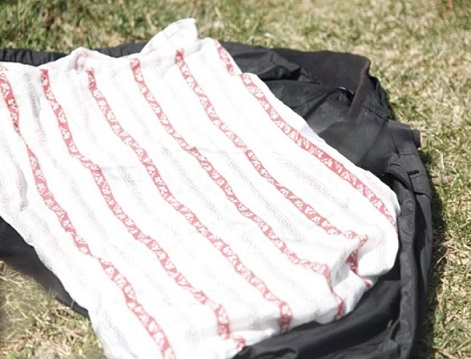}
		\vspace{-5.5mm} \caption*{{\footnotesize DSC (ours)}} 
	\end{subfigure}
	\begin{subfigure}{0.138\textwidth}
		\includegraphics[width=\textwidth]{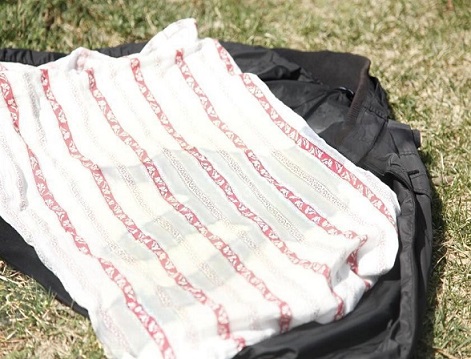}
		\vspace{-5.5mm} \caption*{{\footnotesize DeshadowNet~\cite{qu2017deshadownet}}}
	\end{subfigure}
	\begin{subfigure}{0.138\textwidth}
		\includegraphics[width=\textwidth]{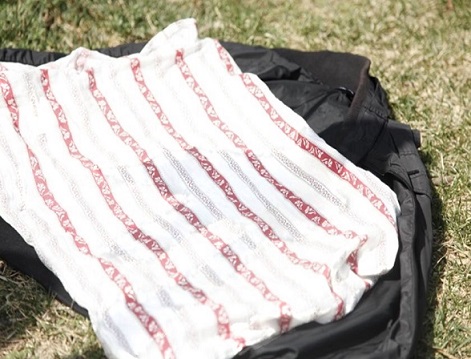}
		\vspace{-5.5mm} \caption*{{\footnotesize Gong \emph{et al.}~\cite{gong2014interactive}}}
	\end{subfigure}
	\begin{subfigure}{0.138\textwidth}
		\includegraphics[width=\textwidth]{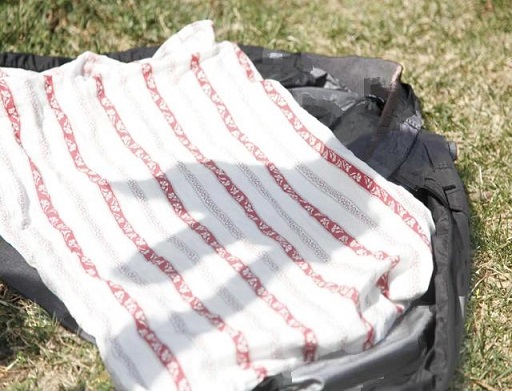}
		\vspace{-5.5mm} \caption*{{\footnotesize Guo \emph{et al.}~\cite{guo2013paired}}}
	\end{subfigure}

	\vspace*{-1mm}
	\caption{Visual comparison of shadow removal results on the SRD dataset~\cite{qu2017deshadownet}.}
	\label{fig:comparison_removal_real_photos1}
	\vspace*{-1mm}

\end{figure*}

\begin{figure} [tp]
	\centering
	\includegraphics[width=0.95\linewidth]{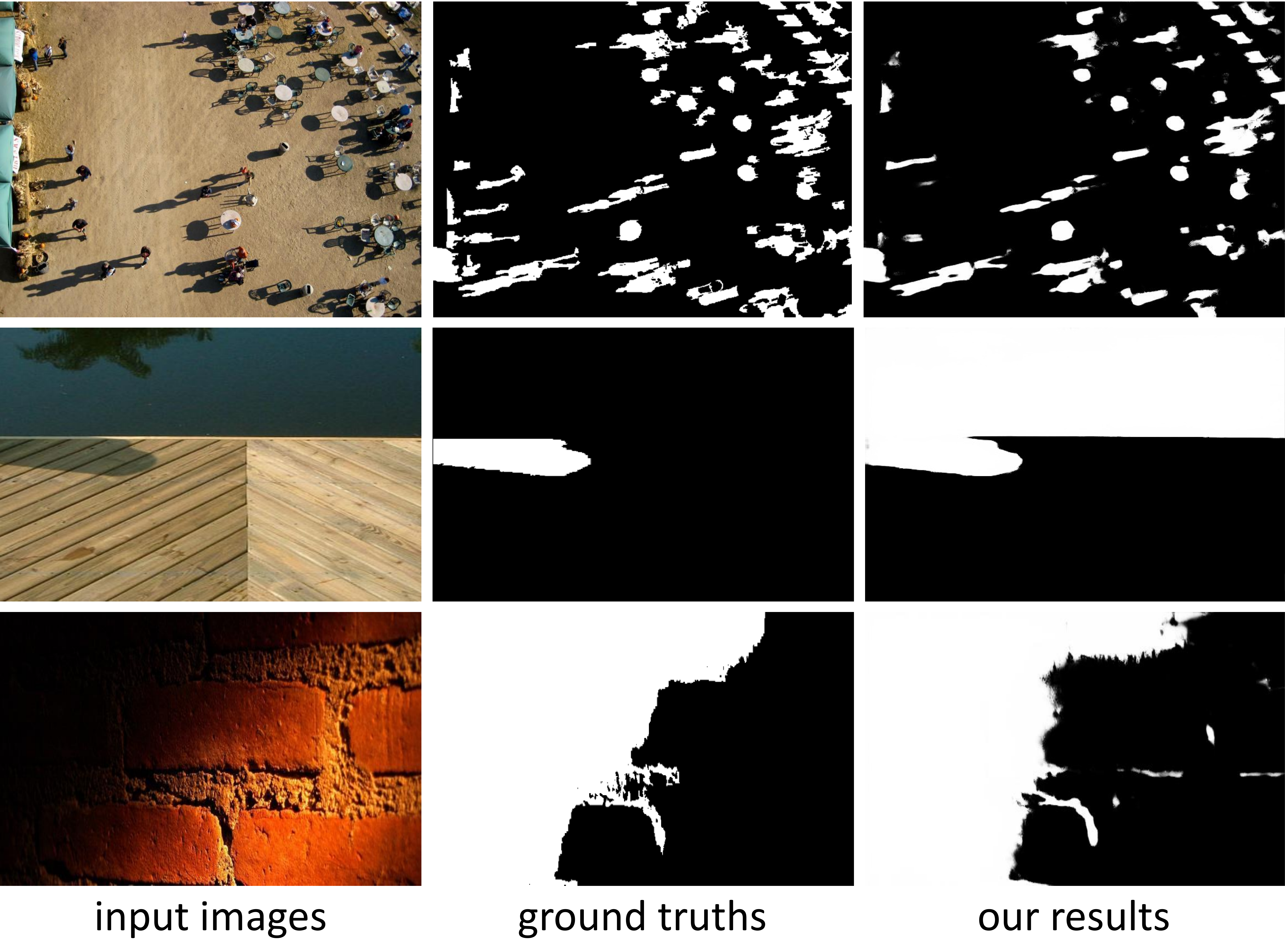}
	\vspace*{-1mm}
	\caption{Failure cases on shadow detection.}
	\label{fig:fail_case}
	\vspace*{-1mm}
\end{figure}


\subsection{Additional Results}

\vspace*{2mm}
\noindent
{\bf More shadow detection results.} \
Figure~\ref{fig:more_results} shows more results:
(a) light and dark shadows next to each other;
(b) small and unconnected shadows;
(c) no clear boundary between shadow and non-shadow regions; and
(d) shadows of irregular shapes.
Our method can still detect these shadows fairly well, but it fails in some extremely complex scenes:
(a) a scene with many small shadows (see the $1^{st}$ row in Figure~\ref{fig:fail_case}), where the features in the deep layers lose the detail information and features in the shallow layers lack the semantics for the shadow context;
(b) a scene with a large black region (see the $2^{nd}$ row in Figure~\ref{fig:fail_case}), where there are insufficient surrounding context to indicate whether it is a shadow or simply a black object; and
(c) a scene with soft shadows (see the $3^{rd}$ row in Figure~\ref{fig:fail_case}), where the difference between the soft shadow regions and the non-shadow regions is small.


\vspace*{2mm}
\noindent
{\bf Time performance.} \
Our network is fast, due to its fully convolutional architecture and the simple implementation of RNN model~\cite{le2015simple}.
We trained and tested our network for shadow detection on a single GPU (NVIDIA GeForce TITAN Xp), and used an input size of 400$\times$400 for each image.
It takes around 16.5 hours to train the whole network on the SBU training set and around 0.16 seconds on average to test one image.
For post-processing with the CRF~\cite{krahenbuhl2011efficient}, it takes another 0.5 seconds to test an image.


\section{Experiments on Shadow Removal}
\label{sec:experiments_removal}


\subsection{Shadow Removal Datasets \& Evaluation Metrics}


\vspace*{2mm}
\noindent
{\bf Benchmark datasets.} \
We employ two shadow removal benchmark datasets.
The first one is SRD~\cite{qu2017deshadownet}, which is the first large-scale dataset with shadow image and shadow-free image pairs, containing 2680 training pairs and 408 testing pairs.
It includes images under different illuminations and a variety of scenes, and the shadows are cast on different kinds of backgrounds with various shapes and silhouettes.
The second one is ISTD~\cite{wang2018stacked}, which contains the triplets of shadow image, shadow mask, and shadow-free image, including 1330 training triplets and 540 testing triplets. 
This dataset covers various shadow shapes under 135 different cases of ground materials.


\vspace*{2mm}
\noindent
{\bf Evaluation metrics.} \
We quantitatively evaluate the shadow removal performance by calculating the root-mean-square error (RMSE) in ``LAB'' color space between the ground truth image and predicted shadow-free image, following~\cite{guo2013paired,wang2018stacked,qu2017deshadownet}.
Hence, a low RMSE value indicates good performance.


\begin{figure*}[tp]
	\centering
	
			\vspace*{0.5mm}
	\begin{subfigure}{0.138\textwidth} 
		\includegraphics[width=\textwidth]{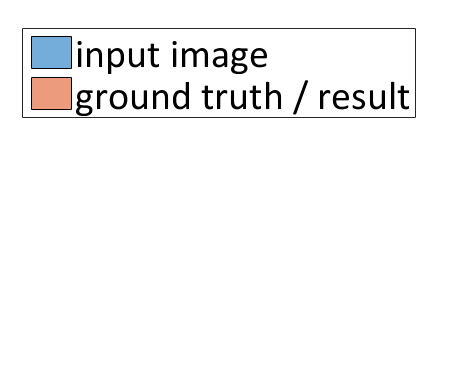}
	\end{subfigure}
	\begin{subfigure}{0.138\textwidth} 
		\includegraphics[width=\textwidth]{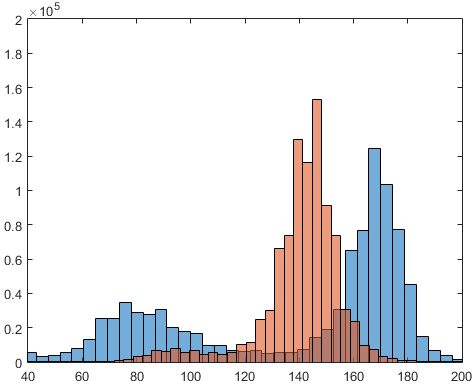}
	\end{subfigure}
	\begin{subfigure}{0.138\textwidth} 
		\includegraphics[width=\textwidth]{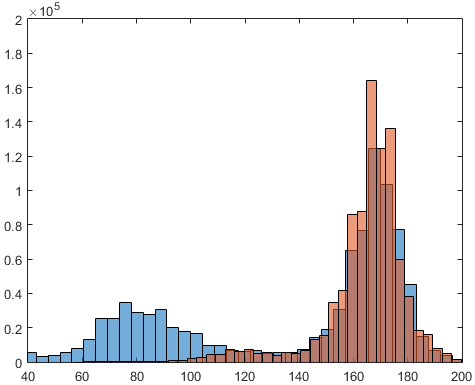}
	\end{subfigure}
	\begin{subfigure}{0.138\textwidth} 
		\includegraphics[width=\textwidth]{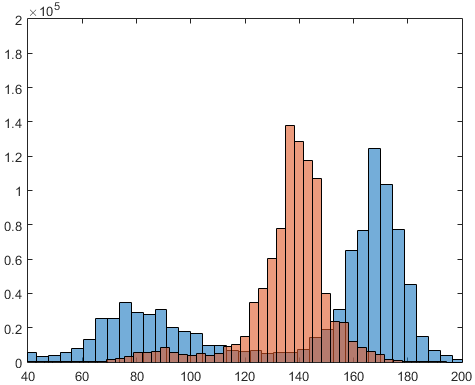}
	\end{subfigure}
	\begin{subfigure}{0.138\textwidth} 
		\includegraphics[width=\textwidth]{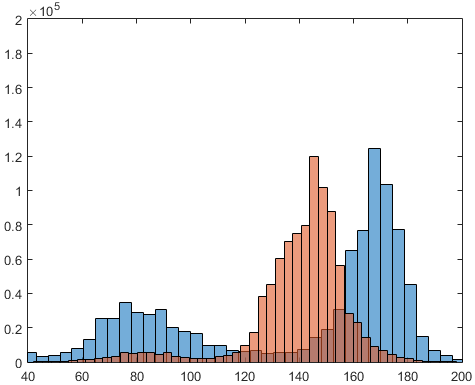}
	\end{subfigure}
	\begin{subfigure}{0.138\textwidth} 
		\includegraphics[width=\textwidth]{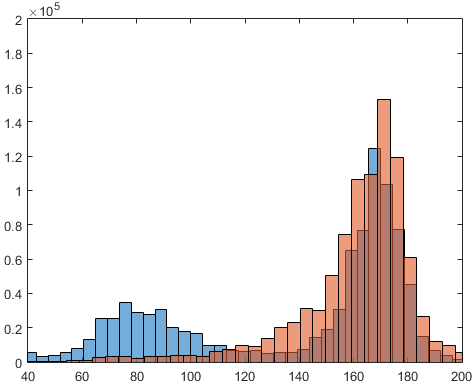}
	\end{subfigure}
	\begin{subfigure}{0.138\textwidth} 
		\includegraphics[width=\textwidth]{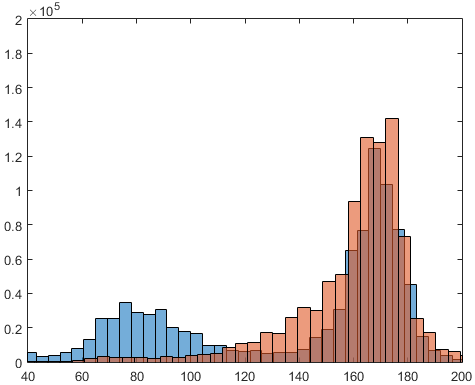}
	\end{subfigure}
	
	\vspace*{0.5mm}
	
	\begin{subfigure}{0.138\textwidth} 
		\includegraphics[width=\textwidth]{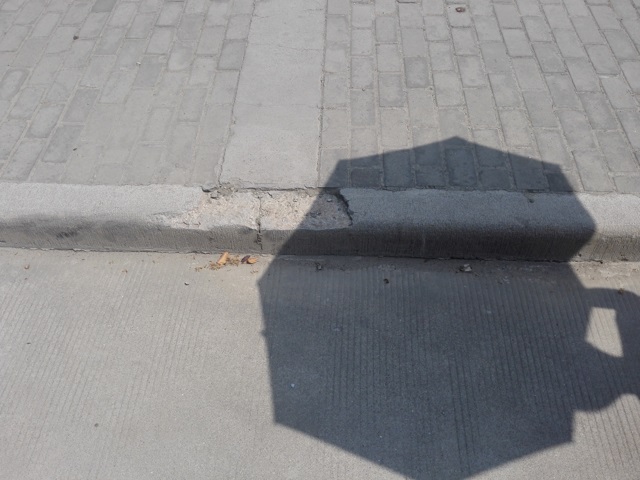}
	\end{subfigure}
	\begin{subfigure}{0.138\textwidth}
		\includegraphics[width=\textwidth]{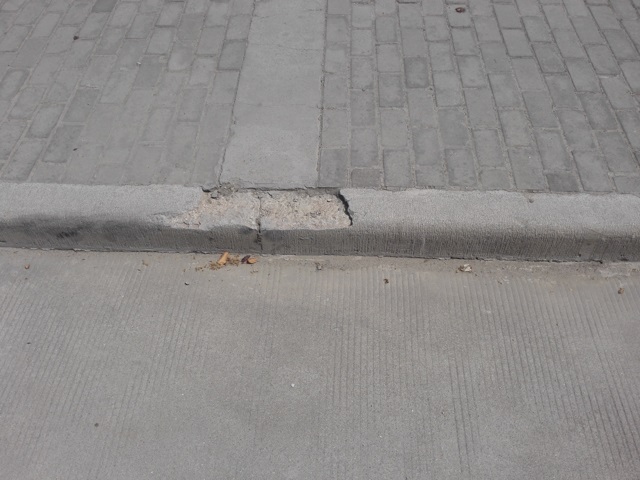}
	\end{subfigure}
     \begin{subfigure}{0.138\textwidth}
     	\includegraphics[width=\textwidth]{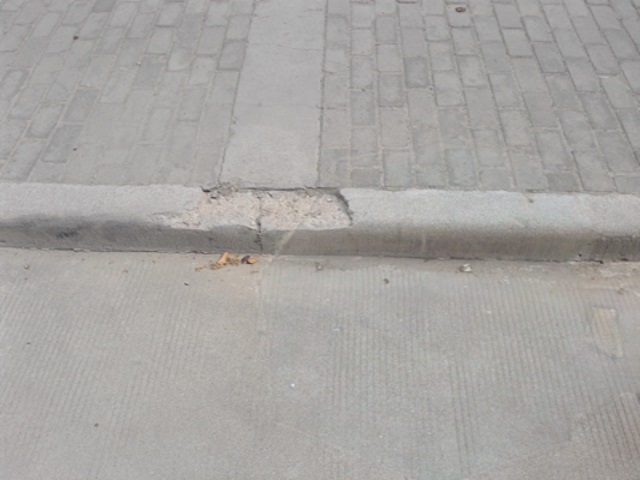}
     \end{subfigure}
	\begin{subfigure}{0.138\textwidth}
		\includegraphics[width=\textwidth]{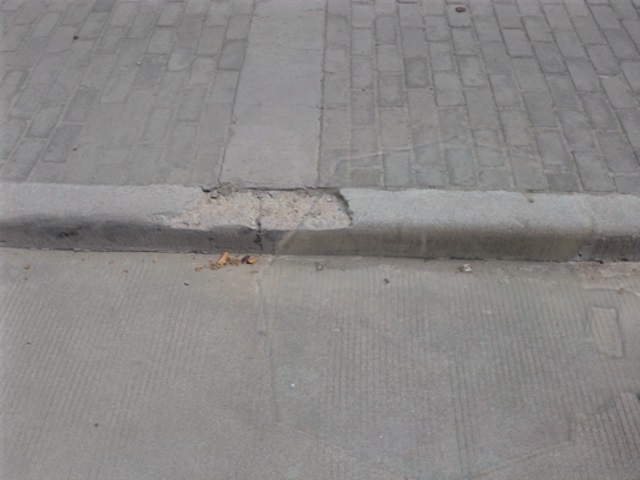}
	\end{subfigure}
	\begin{subfigure}{0.138\textwidth}
		\includegraphics[width=\textwidth]{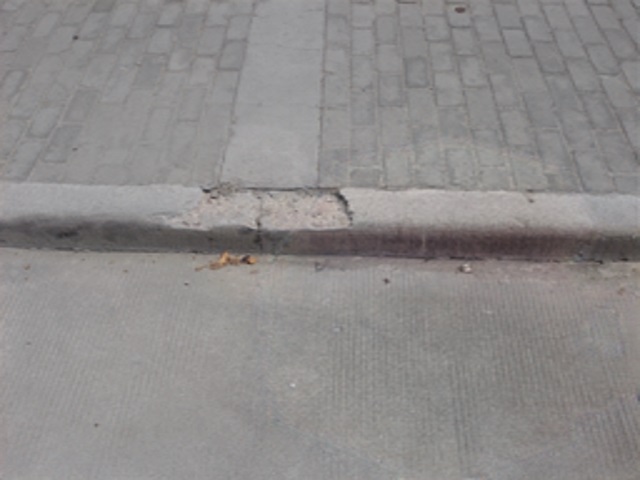}
	\end{subfigure}
	\begin{subfigure}{0.138\textwidth}
		\includegraphics[width=\textwidth]{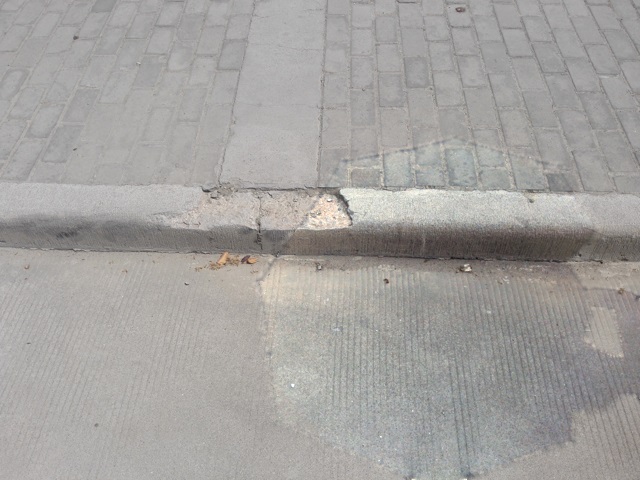}
	\end{subfigure}
	\begin{subfigure}{0.138\textwidth}
		\includegraphics[width=\textwidth]{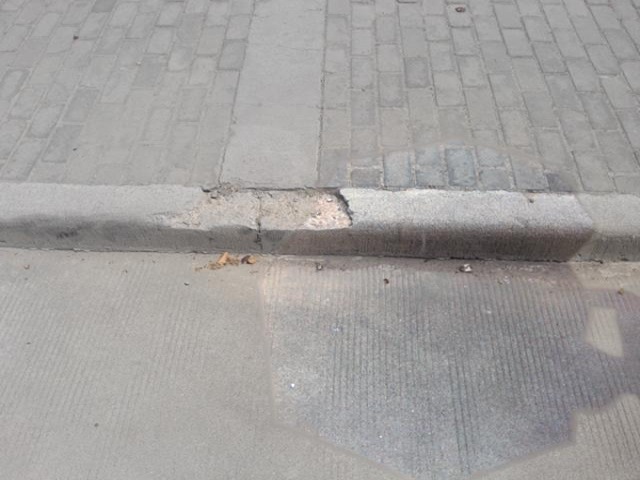}
	\end{subfigure}

	\vspace*{0.5mm}
	\begin{subfigure}{0.138\textwidth} 
		\includegraphics[width=\textwidth]{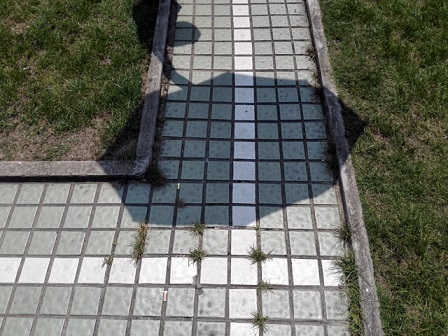}
	\end{subfigure}
	\begin{subfigure}{0.138\textwidth}
		\includegraphics[width=\textwidth]{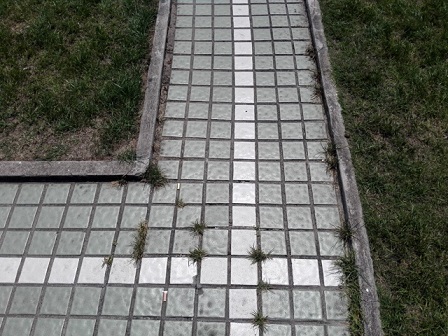}
	\end{subfigure}
     \begin{subfigure}{0.138\textwidth}
     	\includegraphics[width=\textwidth]{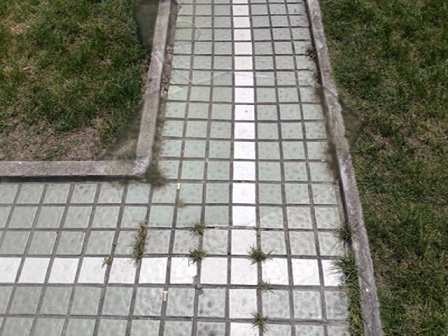}
     \end{subfigure}
	\begin{subfigure}{0.138\textwidth}
		\includegraphics[width=\textwidth]{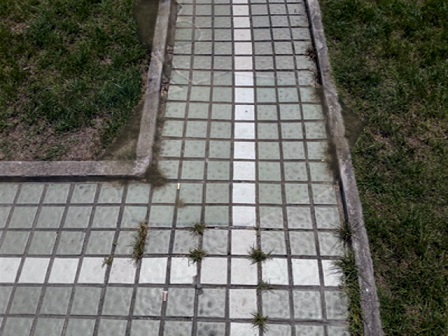}
	\end{subfigure}
	\begin{subfigure}{0.138\textwidth}
		\includegraphics[width=\textwidth]{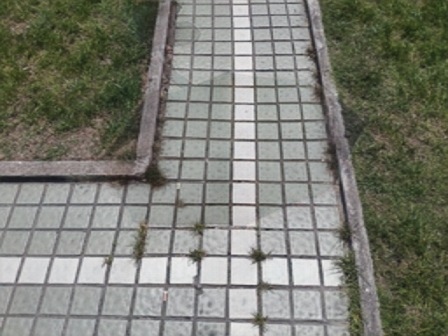}
	\end{subfigure}
	\begin{subfigure}{0.138\textwidth}
		\includegraphics[width=\textwidth]{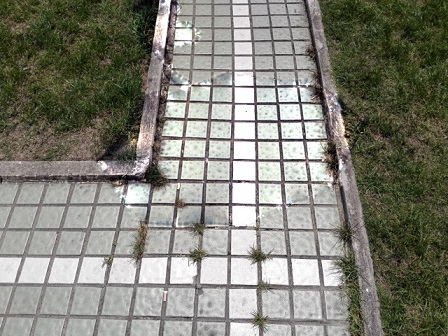}
	\end{subfigure}
	\begin{subfigure}{0.138\textwidth}
		\includegraphics[width=\textwidth]{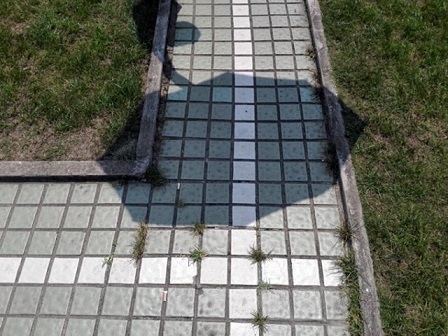}
	\end{subfigure}

	\vspace*{0.5mm}
	\begin{subfigure}{0.138\textwidth} 
		\includegraphics[width=\textwidth]{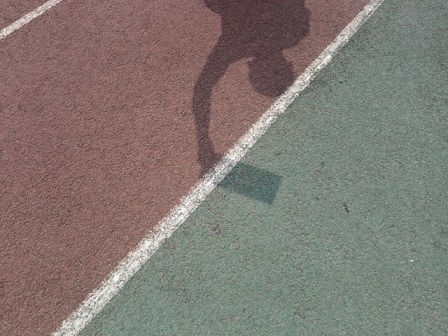}
	\end{subfigure}
	\begin{subfigure}{0.138\textwidth}
		\includegraphics[width=\textwidth]{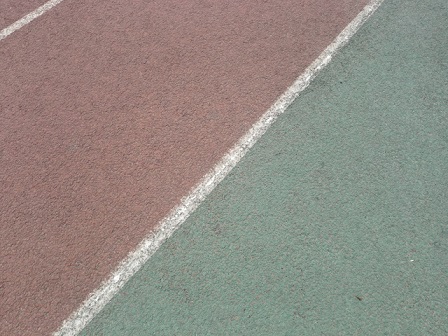}
	\end{subfigure}
    \begin{subfigure}{0.138\textwidth}
    	\includegraphics[width=\textwidth]{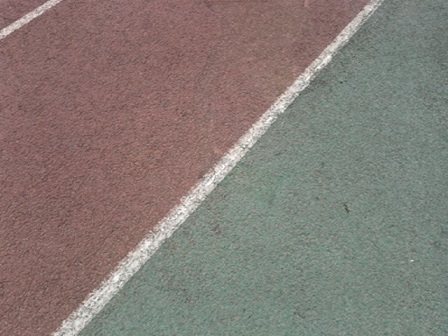}
    \end{subfigure}
	\begin{subfigure}{0.138\textwidth}
		\includegraphics[width=\textwidth]{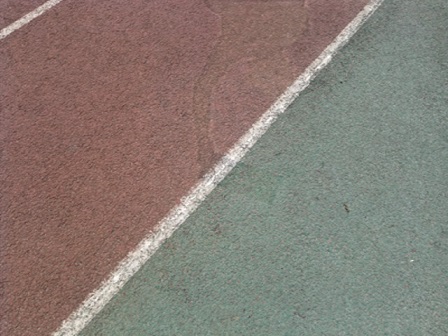}
	\end{subfigure}
	\begin{subfigure}{0.138\textwidth}
		\includegraphics[width=\textwidth]{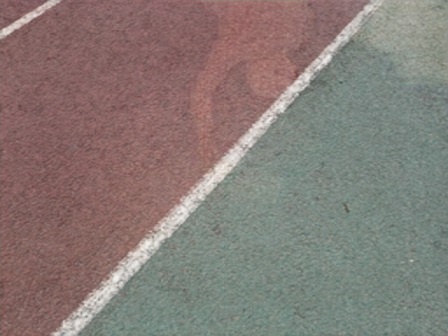}
	\end{subfigure}
	\begin{subfigure}{0.138\textwidth}
		\includegraphics[width=\textwidth]{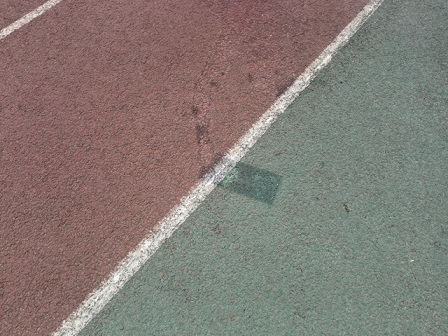}
	\end{subfigure}
	\begin{subfigure}{0.138\textwidth}
		\includegraphics[width=\textwidth]{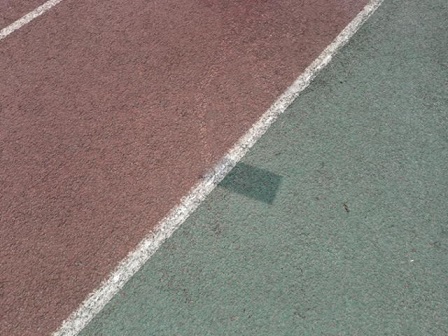}
	\end{subfigure}

    \vspace*{0.5mm}
    \begin{subfigure}{0.138\textwidth} 
    	\includegraphics[width=\textwidth]{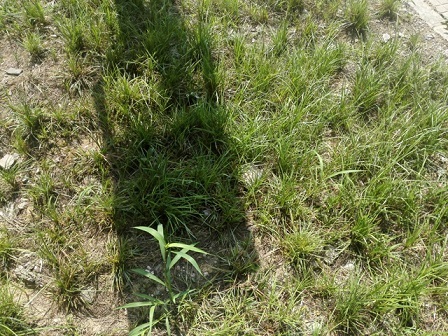}
    	\vspace{-5.5mm} \caption*{{\footnotesize input}}
    \end{subfigure}
    \begin{subfigure}{0.138\textwidth}
    	\includegraphics[width=\textwidth]{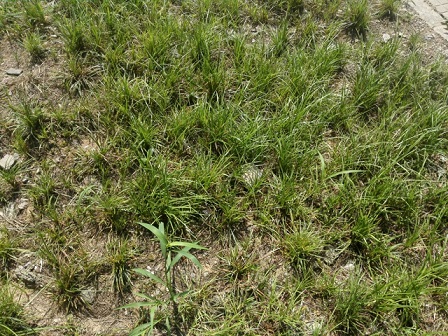}
    	\vspace{-5.5mm} \caption*{{\footnotesize ground truth}} 
    \end{subfigure}
    \begin{subfigure}{0.138\textwidth}
    	\includegraphics[width=\textwidth]{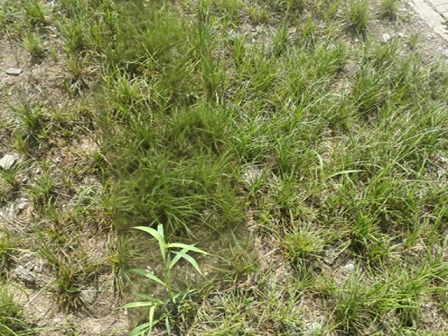}
    	\vspace{-5.5mm} \caption*{{\footnotesize DSC+ (ours)}} 
    \end{subfigure}
    \begin{subfigure}{0.138\textwidth}
    	\includegraphics[width=\textwidth]{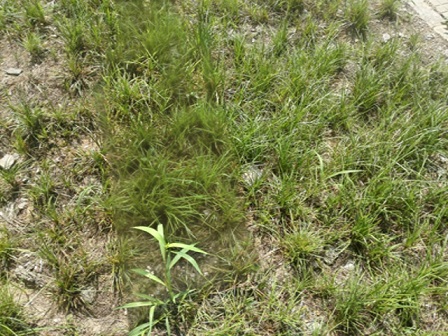}
    	\vspace{-5.5mm} \caption*{{\footnotesize DSC (ours)}} 
    \end{subfigure}
    \begin{subfigure}{0.138\textwidth}
    	\includegraphics[width=\textwidth]{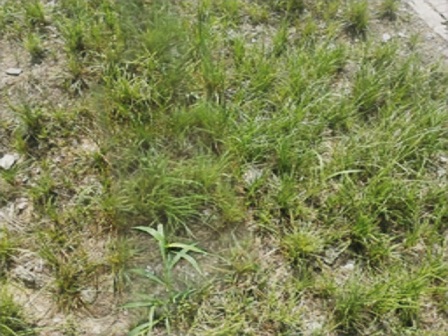}
    	\vspace{-5.5mm} \caption*{{\footnotesize ST-CGAN~\cite{wang2018stacked}}}
    \end{subfigure}
    \begin{subfigure}{0.138\textwidth}
    	\includegraphics[width=\textwidth]{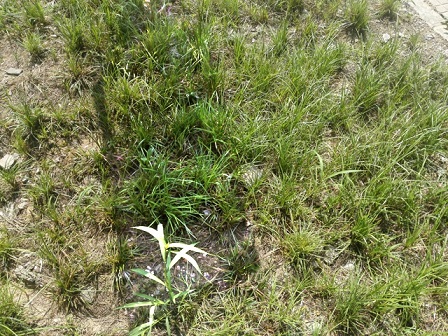}
    	\vspace{-5.5mm} \caption*{{\footnotesize Gong \emph{et al.}~\cite{gong2014interactive}}}
    \end{subfigure}
    \begin{subfigure}{0.138\textwidth}
    	\includegraphics[width=\textwidth]{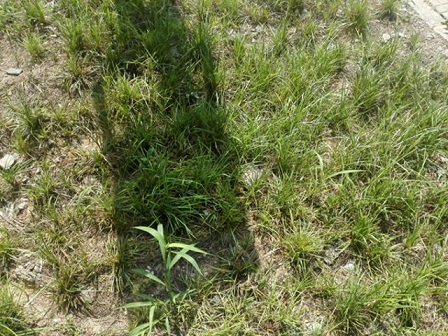}
    	\vspace{-5.5mm} \caption*{{\footnotesize Guo \emph{et al.}~\cite{guo2013paired}}}
    \end{subfigure}
	
	\caption{Visual comparison of shadow removal results on the ISTD dataset~\cite{wang2018stacked}.
The histograms in the top (first) row reveal the intensity distribution of the images in the second row, where the blue histograms show the intensity distribution of the leftmost input image and the red histograms show the intensity distribution of the ground truth or result images in the second row below the histograms.}
	\label{fig:comparison_removal_real_photos2}
	
\end{figure*}


\subsection{Comparison with the State-of-the-art}

The shadow removal methods compute the RMSE directly between the predicted shadow-free image and ground truth shadow-free image 
{\em without\/} any color adjustment.
Hence, for a fair quantitative comparison between our method and the state-of-the-art methods, we apply our network trained on the original shadow-free images that are without the color adjustment.
We denote this network as ``DSC'', and our network trained on the shadow-free images with the color adjustment as ``DSC+''.



\begin{table}[!t]
	\begin{center}
		\caption{Comparing our method (DSC) with recent methods for shadow removal in terms of RMSE.
			Note that the code of ST-CGAN~\cite{wang2018stacked} and DeshadowNet~\cite{qu2017deshadownet} are not publicly available, so we can only directly compare with their RMSE results (i.e., 6.64 and 7.47) on their respective datasets.}
		\label{table:state-of-the-art_removal}
		\begin{tabular}{c|c|c}
			&
			SRD~\cite{qu2017deshadownet} &
			ISTD~\cite{wang2018stacked}
			\\
			\hline
			\hline
			\textbf{DSC (ours)} & \textbf{6.21} &  \textbf{6.67}   \\
			\hline
			\hline
			ST-CGAN~\cite{wang2018stacked} & - & 7.47\\
			
			DeshadowNet~\cite{qu2017deshadownet} & 6.64 &  -    \\
			
			Gong \emph{et al.}~\cite{gong2014interactive} & 8.73 &  8.53  \\
			
			Guo \emph{et al.}~\cite{guo2013paired} & 12.60 & 9.30 \\
			
			Yang \emph{et al.}~\cite{yang2012shadow} & 22.57 & 15.63 \\
			\hline
			
		\end{tabular}
	\end{center}
	\vspace{-3mm}
\end{table}

We consider the following five recent shadow removal methods in our comparison:
ST-CGAN~\cite{wang2018stacked}, DeshadowNet~\cite{qu2017deshadownet}, Gong~\emph{et al.}~\cite{gong2014interactive}, Guo~\emph{et al.}~\cite{guo2013paired}, and Yang~\emph{et al.}~\cite{yang2012shadow}.
We obtain their shadow removal results directly from the authors or by generating them using the public code with the recommended parameter setting.
Table~\ref{table:state-of-the-art_removal} presents the comparison results; note that we do not have the result of ST-CGAN~\cite{wang2018stacked} on the SRD dataset~\cite{qu2017deshadownet} (and similarly, the result of DeshadowNet~\cite{qu2017deshadownet} on the ISTD dataset~\cite{wang2018stacked}), since we do not have the code for these two methods.
DeshadowNet~\cite{qu2017deshadownet} and ST-CGAN~\cite{wang2018stacked} are the two most recent shadow removal methods, which exploit the global image semantics in the convolutional neural network by a multi-context architecture and adversarial learning.
By further considering the global context information in a direction-aware manner, we can see from Table~\ref{table:state-of-the-art_removal} that our method outperforms them on the respective dataset, demonstrating the effectiveness of our network.


We provide visual comparison results on these two datasets in Figures~\ref{fig:comparison_removal_real_photos1} and~\ref{fig:comparison_removal_real_photos2}, which show several challenging cases, e.g., dark non-shadow regions (the $2^{nd}$ row in Figure~\ref{fig:comparison_removal_real_photos1}) and shadows across multiple types of backgrounds.
From the results, we can see that our methods (DSC \& DSC+) can effectively remove the shadows as well as maintain the input image contents in the non-shadow regions. 
By introducing the color compensation mechanism, our DSC+ model can further produce shadow-free images that are more consistent with the input images.
In the comparison results, other methods may change the colors on the non-shadow regions or fail to remove parts of the shadows.


\begin{table}[!t]
	\begin{center}
		\caption{Evaluate our methods (DSC \& DSC+) on the original ground truth ($I_n$) and the adjusted ground truth ($T_f(I_n)$). The performance is evaluated by using the RMSE metric.}
	\label{table:adjusted_GT}
	\begin{tabular}{c||c|c||c|c}
		&
		\multicolumn{2}{c||}{SRD~\cite{qu2017deshadownet}} &
		\multicolumn{2}{c}{ISTD~\cite{wang2018stacked}}
		\\
		\hline
		& $I_n$ & $T_f(I_n)$
		& $I_n$ & $T_f(I_n)$
		\\
		\hline
		DSC &6.21 & 6.66 & 6.67 &  8.54  \\
		DSC+ & 6.75 & \textbf{6.12} & 8.91 & \textbf{4.90}\\
		\hline
		
	\end{tabular}
	
\end{center}
\vspace{-3mm}
\end{table}


\subsection{Evaluation on the Network Design}

\vspace*{2mm}
\noindent
{\bf Color compensation mechanism analysis.} \
The first two pairs of images (input images and ground truth images) in the $2^{nd}$ and $3^{rd}$ rows of Figure~\ref{fig:comparison_removal_real_photos2} show the inconsistent color and luminosity (also revealed by the first histogram on the top row).
Methods based on neural networks (e.g., DSC and ST-CGAN~\cite{wang2018stacked}) could produce inconsistent results due to inconsistencies in the training pairs; see the $4^{th}$ and $5^{th}$ columns in Figure~\ref{fig:comparison_removal_real_photos2} from the left.

By first adjusting the ground truth shadow-free images, our DSC+ can learn to generate shadow-free images whose colors are more consistent and faithful to the input images; see the $3^{rd}$ column in Figure~\ref{fig:comparison_removal_real_photos2} and the histograms above.
%
%
Furthermore, we tried to use the adjusted shadow-free images ($T_f(I_n)$) instead of the original shadow-free images ($I_n$) as the ground truth images to compute the RMSE for our methods (DSC and DSC+).
Table~\ref{table:adjusted_GT} shows the comparison results: DSC has a large RMSE when compared with the adjusted ground truth images (6.21 vs. 6.66 and 6.67 vs. 8.54), while DSC+ shows a clear improvement (6.66 vs. 6.12 and 8.54 vs. 4.90), especially on the ISTD dataset~\cite{wang2018stacked}.
Moreover, since the shadow masks are available for the ISTD dataset, we calculate the RMSE between the shadow removal results and the input images on the non-shadow regions. The RMSE values for DSC and DSC+ are 7.07 and 3.36, respectively, further revealing the effectiveness of DSC+.

\begin{table}[!t]
	\begin{center}
		\caption{Train and test our method (DSC) on different color spaces. The performance is evaluated by using the RMSE metric.}
		\label{table:color_space}
		\begin{tabular}{c|c|c}
			color space &
			SRD~\cite{qu2017deshadownet} &
			ISTD~\cite{wang2018stacked}
			\\
			\hline
			LAB & 6.21 &  \textbf{6.67}   \\

			RGB & \textbf{6.05} & 6.92 \\
			\hline
			
		\end{tabular}
	\end{center}
\vspace{-3mm}
\end{table}

\vspace*{2mm}
\noindent
{\bf Color space analysis.} \
We performed another experiment to evaluate the choice of color space in data processing.
In this experiment, we consider the ``LAB'' and ``RGB'' color spaces, and train a shadow removal network for each of them.
As shown in the results presented in Table~\ref{table:color_space}, the performance of the two networks is similar.
Since the overall (summed) performance with the LAB color space is slightly better, we thus choose to use LAB in our method.
However, in any case, both results outperform the state-of-the-art methods on shadow removal shown in Table~\ref{table:state-of-the-art_removal}.

\subsection{Additional Results}

\begin{figure} [tp]
	\centering
	\includegraphics[width=0.95\linewidth]{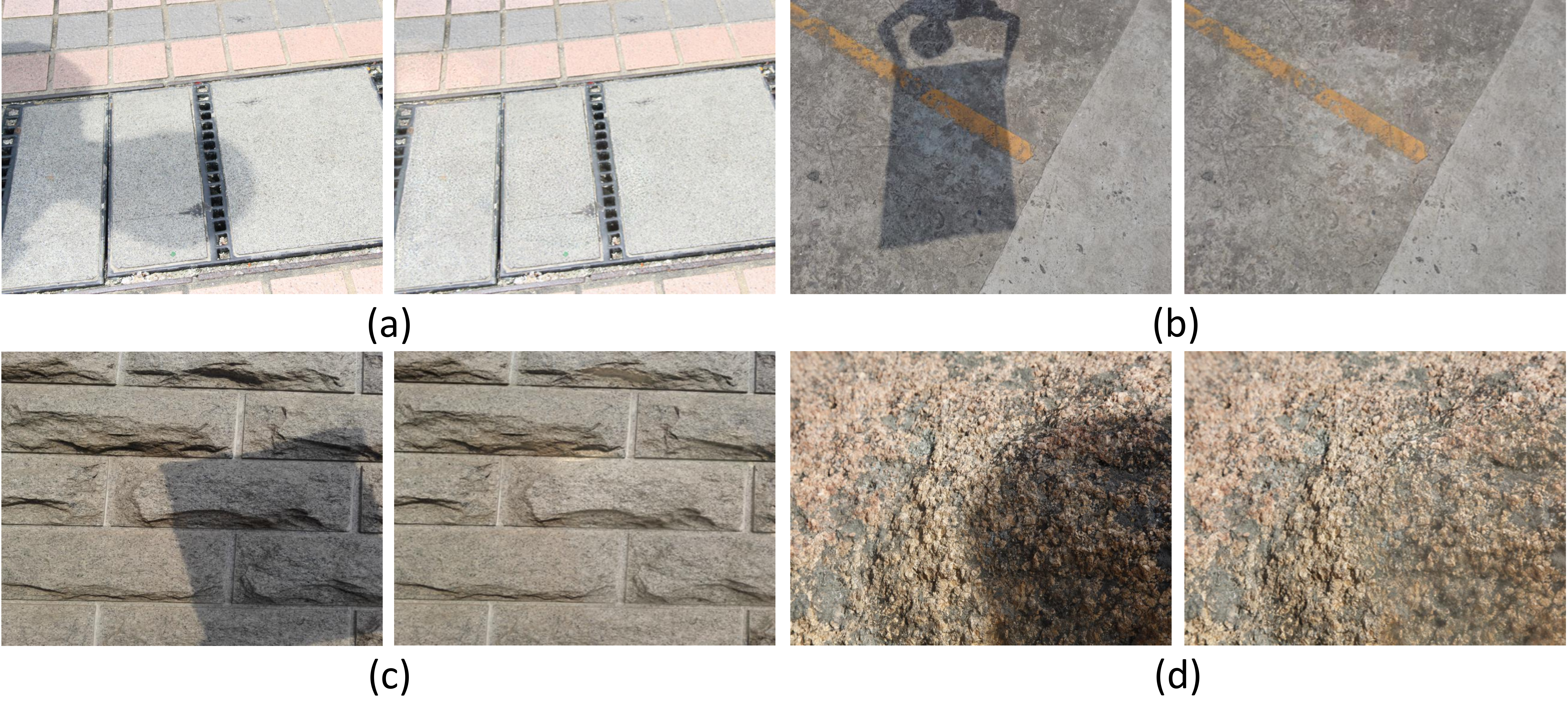}
	\vspace{-2mm}
	\caption{More shadow removal results produced from our DSC+.}
	\label{fig:more_results_removal}
\end{figure}

\begin{figure} [tp]
	\centering
	\includegraphics[width=0.95\linewidth]{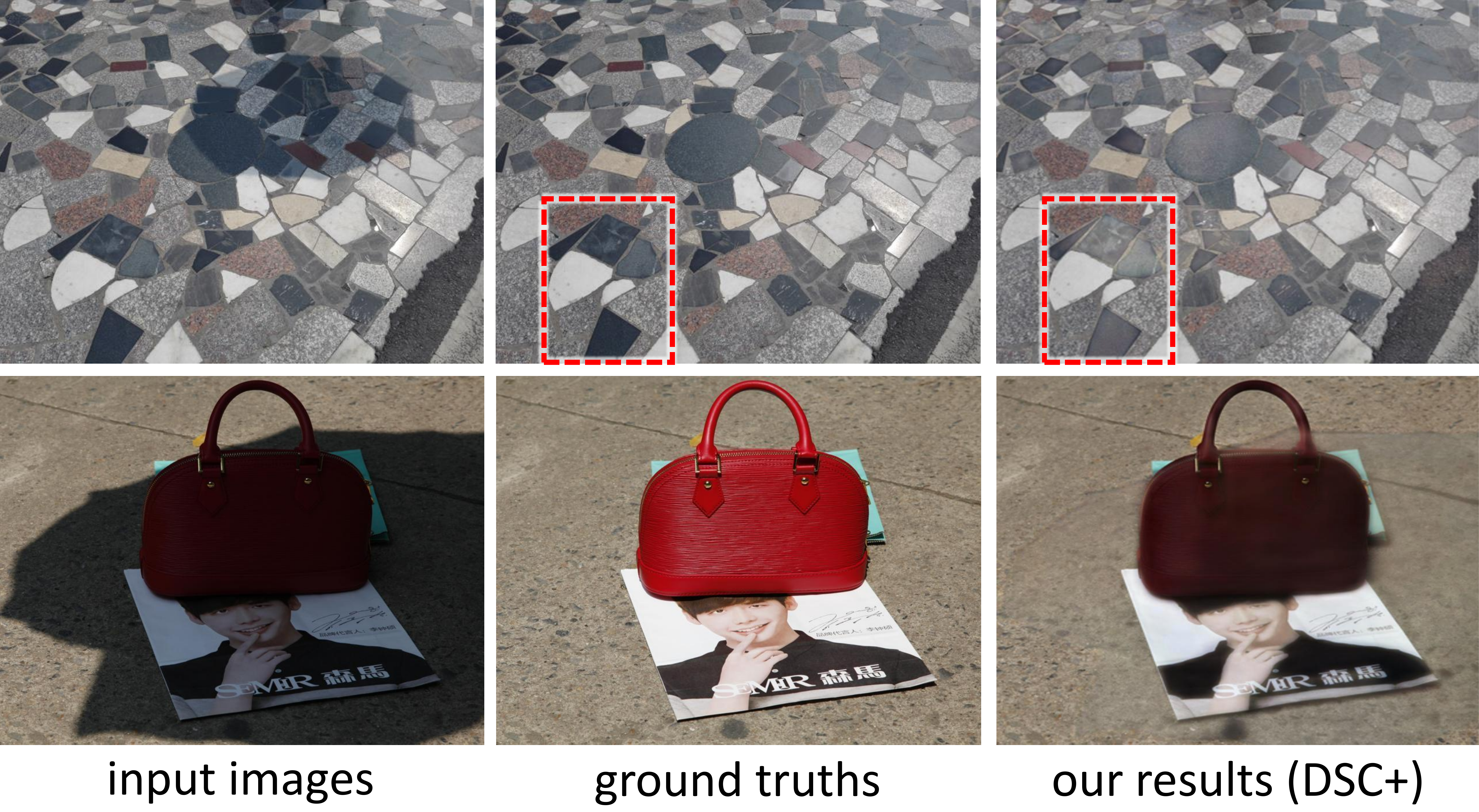}
	\vspace{-2mm}
	\caption{Failure cases on shadow removal.}
	\label{fig:fail_case_removal}
\end{figure}

\vspace*{2mm}
\noindent
{\bf More shadow removal results.} \
Figure~\ref{fig:more_results_removal} presents more results:
(a) and (b) show shadows across backgrounds of different colors, (c) shows small, unconnected shadows of irregular shapes on the stones, and (d) shows shadows on a complex background.
Our method can still reasonably remove these shadows. However, for the cases shown in Figure~\ref{fig:fail_case_removal}:
(a) it overly removes the fragmented black tiles on the floor (see the red dashed boxes in the figure), where the surrounding context provides incorrect information, and (b) it fails to recover the original (bright) color of the handbag, due to the lack of information.
We believe that more training data is needed for the network to learn and overcome these problems.

\vspace*{2mm}
\noindent
{\bf Time performance.} \
Same as our shadow detection network, we trained and tested our shadow removal network on the same GPU (NVIDIA GeForce TITAN Xp), and used the same input image size as in our shadow detection network.
It takes around 22 hours to train the whole network on the SRD training set and another 22 hours to train it on the ISTD training set. In testing, it only takes around 0.16 seconds on average to process a 400$\times$400 image.

\section{Conclusion}
\label{sec::conclusion}

We present a novel network for single-image shadow detection and removal by harvesting the direction-aware spatial context.
Our key idea is to analyze the multi-level spatial context in a direction-aware manner by formulating a direction-aware attention mechanism in a spatial RNN.
By training the network to automatically learn the attention weights for leveraging the spatial context in different directions in the spatial RNN, we can produce direction-aware spatial context (DSC) features and formulate the DSC module.
Then, we adopt multiple DSC modules in a multi-layer convolutional neural network to detect shadows by predicting the shadow masks in different scales, and design a weighted cross entropy loss function to make effective the training process.
%
%
Further, we adopt the network for shadow removal by replacing the shadow masks with shadow-free images, applying a Euclidean loss to optimize the network, and introducing a color compensation mechanism to address the color and luminosity inconsistency problem.
In the end, we test our network on two benchmark datasets for shadow detection and another two benchmark datasets for shadow removal, compare our network with various state-of-the-art methods, and show its superiority over the state-of-the-art methods for both shadow detection and shadow removal.

In the future, we plan to explore our network for other applications, e.g., saliency detection and semantic segmentation, further enhancing the shadow removal results by exploring strategies in image completion, and studying time-varying shadows in videos.

\ifCLASSOPTIONcompsoc
  \section*{Acknowledgments}
\else
  \section*{Acknowledgment}
\fi

This work was supported by the National Basic Program of China, 973 Program (Project no. 2015CB351706), the Shenzhen Science and Technology Program (Project no. JCYJ20170413162617606), and the Hong Kong Research Grants Council (Project no. CUHK 14225616, PolyU 152035/17E, \& CUHK 14203416).
Xiaowei Hu is funded by the Hong Kong Ph.D. Fellowship.
We thank reviewers for their valuable comments, Michael S. Brown for his discussion, and Tomas F. Yago Vicente, Minh Hoai Nguyen, Vn Nguyen, Moein Shakeri, Jiandong Tian and Jifeng Wang for sharing their results and evaluation code with us.


\ifCLASSOPTIONcaptionsoff
  \newpage
\fi



%

{\small
	\bibliographystyle{IEEEtran}
	\bibliography{egbib}

\begin{thebibliography}{10}
\providecommand{\url}[1]{#1}
\csname url@samestyle\endcsname
\providecommand{\newblock}{\relax}
\providecommand{\bibinfo}[2]{#2}
\providecommand{\BIBentrySTDinterwordspacing}{\spaceskip=0pt\relax}
\providecommand{\BIBentryALTinterwordstretchfactor}{4}
\providecommand{\BIBentryALTinterwordspacing}{\spaceskip=\fontdimen2\font plus
\BIBentryALTinterwordstretchfactor\fontdimen3\font minus
  \fontdimen4\font\relax}
\providecommand{\BIBforeignlanguage}[2]{{%
\expandafter\ifx\csname l@#1\endcsname\relax
\typeout{** WARNING: IEEEtran.bst: No hyphenation pattern has been}%
\typeout{** loaded for the language `#1'. Using the pattern for}%
\typeout{** the default language instead.}%
\else
\language=\csname l@#1\endcsname
\fi
#2}}
\providecommand{\BIBdecl}{\relax}
\BIBdecl

\bibitem{hu2018direction}
X.~Hu, L.~Zhu, C.-W. Fu, J.~Qin, and P.-A. Heng, ``Direction-aware spatial
  context features for shadow detection,'' in \emph{IEEE Conference on Computer
  Vision and Pattern Recognition}, 2018, pp. 7454--7462, oral presentation.

\bibitem{lalonde2009estimating}
J.-F. Lalonde, A.~A. Efros, and S.~G. Narasimhan, ``Estimating natural
  illumination from a single outdoor image,'' in \emph{IEEE International
  Conference on Computer Vision}, 2009, pp. 183--190.

\bibitem{junejo2008estimating}
I.~N. Junejo and H.~Foroosh, ``Estimating geo-temporal location of stationary
  cameras using shadow trajectories,'' in \emph{European Conference on Computer
  Vision}, 2008, pp. 318--331.

\bibitem{okabe2009attached}
T.~Okabe, I.~Sato, and Y.~Sato, ``Attached shadow coding: Estimating surface
  normals from shadows under unknown reflectance and lighting conditions,'' in
  \emph{IEEE International Conference on Computer Vision}, 2009, pp.
  1693--1700.

\bibitem{karsch2011rendering}
K.~Karsch, V.~Hedau, D.~Forsyth, and D.~Hoiem, ``Rendering synthetic objects
  into legacy photographs,'' \emph{ACM Transactions on Graphics (SIGGRAPH
  Asia)}, vol.~30, no.~6, pp. 157:1--157:12, 2011.

\bibitem{cucchiara2003detecting}
R.~Cucchiara, C.~Grana, M.~Piccardi, and A.~Prati, ``Detecting moving objects,
  ghosts, and shadows in video streams,'' \emph{IEEE Transactions on Pattern
  Analysis and Machine Intelligence}, vol.~25, no.~10, pp. 1337--1342, 2003.

\bibitem{nadimi2004physical}
S.~Nadimi and B.~Bhanu, ``Physical models for moving shadow and object
  detection in video,'' \emph{IEEE Transactions on Pattern Analysis and Machine
  Intelligence}, vol.~26, no.~8, pp. 1079--1087, 2004.

\bibitem{salvador2004cast}
E.~Salvador, A.~Cavallaro, and T.~Ebrahimi, ``Cast shadow segmentation using
  invariant color features,'' \emph{Computer Vision and Image Understanding},
  vol.~95, no.~2, pp. 238--259, 2004.

\bibitem{panagopoulos2011illumination}
A.~Panagopoulos, C.~Wang, D.~Samaras, and N.~Paragios, ``Illumination
  estimation and cast shadow detection through a higher-order graphical
  model,'' in \emph{IEEE Conference on Computer Vision and Pattern
  Recognition}, 2011, pp. 673--680.

\bibitem{tian2016new}
J.~Tian, X.~Qi, L.~Qu, and Y.~Tang, ``New spectrum ratio properties and
  features for shadow detection,'' \emph{Pattern Recognition}, vol.~51, pp.
  85--96, 2016.

\bibitem{finlayson2006removal}
G.~D. Finlayson, S.~D. Hordley, C.~Lu, and M.~S. Drew, ``On the removal of
  shadows from images,'' \emph{IEEE Transactions on Pattern Analysis and
  Machine Intelligence}, vol.~28, no.~1, pp. 59--68, 2006.

\bibitem{finlayson2002removing}
G.~D. Finlayson, S.~D. Hordley, and M.~S. Drew, ``Removing shadows from
  images,'' in \emph{European Conference on Computer Vision}, 2002, pp.
  823--836.

\bibitem{liu2008texture}
F.~Liu and M.~Gleicher, ``Texture-consistent shadow removal,'' in
  \emph{European Conference on Computer Vision}, 2008, pp. 437--450.

\bibitem{finlayson2009entropy}
G.~D. Finlayson, M.~S. Drew, and C.~Lu, ``Entropy minimization for shadow
  removal,'' \emph{International Journal of Computer Vision}, vol.~85, no.~1,
  pp. 35--57, 2009.

\bibitem{wu2007natural}
T.-P. Wu, C.-K. Tang, M.~S. Brown, and H.-Y. Shum, ``Natural shadow matting,''
  \emph{ACM Transactions on Graphics (TOG)}, vol.~26, no.~2, p.~8, 2007.

\bibitem{khan2016automatic}
S.~H. Khan, M.~Bennamoun, F.~Sohel, and R.~Togneri, ``Automatic shadow
  detection and removal from a single image,'' \emph{IEEE Transactions on
  Pattern Analysis and Machine Intelligence}, vol.~38, no.~3, pp. 431--446,
  2016.

\bibitem{huang2011characterizes}
X.~Huang, G.~Hua, J.~Tumblin, and L.~Williams, ``What characterizes a shadow
  boundary under the sun and sky?'' in \emph{IEEE International Conference on
  Computer Vision}, 2011, pp. 898--905.

\bibitem{lalonde2010detecting}
J.-F. Lalonde, A.~A. Efros, and S.~G. Narasimhan, ``Detecting ground shadows in
  outdoor consumer photographs,'' in \emph{European Conference on Computer
  Vision}, 2010, pp. 322--335.

\bibitem{zhu2010learning}
J.~Zhu, K.~G. Samuel, S.~Z. Masood, and M.~F. Tappen, ``Learning to recognize
  shadows in monochromatic natural images,'' in \emph{IEEE Conference on
  Computer Vision and Pattern Recognition}, 2010, pp. 223--230.

\bibitem{guo2013paired}
R.~Guo, Q.~Dai, and D.~Hoiem, ``Paired regions for shadow detection and
  removal,'' \emph{IEEE Transactions on Pattern Analysis and Machine
  Intelligence}, vol.~35, no.~12, pp. 2956--2967, 2013.

\bibitem{gryka2015learning}
M.~Gryka, M.~Terry, and G.~J. Brostow, ``Learning to remove soft shadows,''
  \emph{ACM Transactions on Graphics (TOG)}, vol.~34, no.~5, p. 153, 2015.

\bibitem{khan2014automatic}
S.~H. Khan, M.~Bennamoun, F.~Sohel, and R.~Togneri, ``Automatic feature
  learning for robust shadow detection,'' in \emph{IEEE Conference on Computer
  Vision and Pattern Recognition}, 2014, pp. 1939--1946.

\bibitem{vicente2016large}
T.~F.~Y. Vicente, L.~Hou, C.-P. Yu, M.~Hoai, and D.~Samaras, ``Large-scale
  training of shadow detectors with noisily-annotated shadow examples,'' in
  \emph{European Conference on Computer Vision}, 2016, pp. 816--832.

\bibitem{wang2018stacked}
J.~Wang, X.~Li, and J.~Yang, ``Stacked conditional generative adversarial
  networks for jointly learning shadow detection and shadow removal,'' in
  \emph{IEEE Conference on Computer Vision and Pattern Recognition}, 2018, pp.
  1788--1797.

\bibitem{nguyen2017shadow}
V.~Nguyen, T.~F.~Y. Vicente, M.~Zhao, M.~Hoai, and D.~Samaras, ``Shadow
  detection with conditional generative adversarial networks,'' in \emph{IEEE
  International Conference on Computer Vision}, 2017, pp. 4510--4518.

\bibitem{qu2017deshadownet}
L.~Qu, J.~Tian, S.~He, Y.~Tang, and R.~W. Lau, ``Deshadow{N}et: A multi-context
  embedding deep network for shadow removal,'' in \emph{IEEE Conference on
  Computer Vision and Pattern Recognition}, 2017, pp. 4067--4075.

\bibitem{vicente2015leave}
T.~F.~Y. Vicente, M.~Hoai, and D.~Samaras, ``Leave-one-out kernel optimization
  for shadow detection,'' in \emph{IEEE International Conference on Computer
  Vision}, 2015, pp. 3388--3396.

\bibitem{guo2011single}
R.~Guo, Q.~Dai, and D.~Hoiem, ``Single-image shadow detection and removal using
  paired regions,'' in \emph{IEEE Conference on Computer Vision and Pattern
  Recognition}, 2011, pp. 2033--2040.

\bibitem{vicente2018leave}
T.~F.~Y. Vicente, M.~Hoai, and D.~Samaras, ``Leave-one-out kernel optimization
  for shadow detection and removal,'' \emph{IEEE Transactions on Pattern
  Analysis and Machine Intelligence}, vol.~40, no.~3, pp. 682--695, 2018.

\bibitem{lee2015deeply}
C.-Y. Lee, S.~Xie, P.~Gallagher, Z.~Zhang, and Z.~Tu, ``Deeply-supervised
  nets,'' in \emph{Artificial Intelligence and Statistics}, 2015, pp. 562--570.

\bibitem{krahenbuhl2011efficient}
P.~Kr{\"a}henb{\"u}hl and V.~Koltun, ``Efficient inference in fully connected
  {CRFs} with {Gaussian} edge potentials,'' in \emph{Advances in Neural
  Information Processing Systems}, 2011, pp. 109--117.

\bibitem{shen2015shadow}
L.~Shen, T.~Wee~Chua, and K.~Leman, ``Shadow optimization from structured deep
  edge detection,'' in \emph{IEEE Conference on Computer Vision and Pattern
  Recognition}, 2015, pp. 2067--2074.

\bibitem{hosseinzadeh2017fast}
S.~Hosseinzadeh, M.~Shakeri, and H.~Zhang, ``Fast shadow detection from a
  single image using a patched convolutional neural network,'' \emph{arXiv
  preprint arXiv:1709.09283}, 2017.

\bibitem{baba2004shadow}
M.~Baba, M.~Mukunoki, and N.~Asada, ``Shadow removal from a real image based on
  shadow density,'' in \emph{ACM SIGGRAPH 2004 Posters}.\hskip 1em plus 0.5em
  minus 0.4em\relax ACM, 2004, p.~60.

\bibitem{gong2014interactive}
H.~Gong and D.~P. Cosker, ``Interactive shadow removal and ground truth for
  variable scene categories,'' in \emph{British Machine Vision Conference},
  2014, pp. 1--11.

\bibitem{shen2008intrinsic}
L.~Shen, P.~Tan, and S.~Lin, ``Intrinsic image decomposition with non-local
  texture cues,'' in \emph{IEEE Conference on Computer Vision and Pattern
  Recognition}, 2008, pp. 1--7.

\bibitem{bousseau2009user}
A.~Bousseau, S.~Paris, and F.~Durand, ``User-assisted intrinsic images,'' in
  \emph{ACM Transactions on Graphics (SIGGRAPH Asia)}, vol.~28, no.~5, 2009, p.
  130.

\bibitem{shen2011intrinsic}
L.~Shen and C.~Yeo, ``Intrinsic images decomposition using a local and global
  sparse representation of reflectance,'' in \emph{IEEE Conference on Computer
  Vision and Pattern Recognition}, 2011, pp. 697--704.

\bibitem{zhao2012closed}
Q.~Zhao, P.~Tan, Q.~Dai, L.~Shen, E.~Wu, and S.~Lin, ``A closed-form solution
  to retinex with nonlocal texture constraints,'' \emph{IEEE Transactions on
  Pattern Analysis and Machine Intelligence}, vol.~34, no.~7, pp. 1437--1444,
  2012.

\bibitem{chen2013simple}
Q.~Chen and V.~Koltun, ``A simple model for intrinsic image decomposition with
  depth cues,'' in \emph{IEEE International Conference on Computer Vision},
  2013, pp. 241--248.

\bibitem{bell2014intrinsic}
S.~Bell, K.~Bala, and N.~Snavely, ``Intrinsic images in the wild,'' \emph{ACM
  Transactions on Graphics (TOG)}, vol.~33, no.~4, p. 159, 2014.

\bibitem{barron2015shape}
J.~T. Barron and J.~Malik, ``Shape, illumination, and reflectance from
  shading,'' \emph{IEEE Transactions on Pattern Analysis and Machine
  Intelligence}, vol.~37, no.~8, pp. 1670--1687, 2015.

\bibitem{narihira2015direct}
T.~Narihira, M.~Maire, and S.~X. Yu, ``Direct intrinsics: Learning
  albedo-shading decomposition by convolutional regression,'' in \emph{IEEE
  International Conference on Computer Vision}, 2015, pp. 2992--2992.

\bibitem{kim2016unified}
S.~Kim, K.~Park, K.~Sohn, and S.~Lin, ``Unified depth prediction and intrinsic
  image decomposition from a single image via joint convolutional neural
  fields,'' in \emph{European Conference on Computer Vision}, 2016, pp.
  143--159.

\bibitem{lettry2018darn}
L.~Lettry, K.~Vanhoey, and L.~Van~Gool, ``Darn: a deep adversarial residual
  network for intrinsic image decomposition,'' in \emph{IEEE Winter Conference
  on Applications of Computer Vision}, 2018, pp. 1359--1367.

\bibitem{cheng2018intrinsic}
L.~Cheng, C.~Zhang, and Z.~Liao, ``Intrinsic image transformation via scale
  space decomposition,'' in \emph{IEEE Conference on Computer Vision and
  Pattern Recognition}, 2018, pp. 656--665.

\bibitem{xie2015holistically}
S.~Xie and Z.~Tu, ``Holistically-nested edge detection,'' in \emph{IEEE
  International Conference on Computer Vision}, 2015, pp. 1395--1403.

\bibitem{lecun2015deep}
Y.~LeCun, Y.~Bengio, and G.~Hinton, ``Deep learning,'' \emph{Nature}, vol. 521,
  no. 7553, pp. 436--444, 2015.

\bibitem{bell2016inside}
S.~Bell, C.~L. Zitnick, K.~Bala, and R.~Girshick, ``Inside-outside net:
  Detecting objects in context with skip pooling and recurrent neural
  networks,'' in \emph{IEEE Conference on Computer Vision and Pattern
  Recognition}, 2016, pp. 2874--2883.

\bibitem{le2015simple}
Q.~V. Le, N.~Jaitly, and G.~E. Hinton, ``A simple way to initialize recurrent
  networks of rectified linear units,'' \emph{arXiv preprint arXiv:1504.00941},
  2015.

\bibitem{krizhevsky2012imagenet}
A.~Krizhevsky, I.~Sutskever, and G.~E. Hinton, ``{ImageNet} classification with
  deep convolutional neural networks,'' in \emph{Advances in Neural Information
  Processing Systems}, 2012, pp. 1097--1105.

\bibitem{simonyan2014very}
K.~Simonyan and A.~Zisserman, ``Very deep convolutional networks for
  large-scale image recognition,'' \emph{arXiv preprint arXiv:1409.1556}, 2014.

\bibitem{shrivastava2016training}
A.~Shrivastava, A.~Gupta, and R.~Girshick, ``Training region-based object
  detectors with online hard example mining,'' in \emph{IEEE Conference on
  Computer Vision and Pattern Recognition}, 2016, pp. 761--769.

\bibitem{deng2009imagenet}
J.~Deng, W.~Dong, R.~Socher, L.-J. Li, K.~Li, and L.~Fei-Fei, ``Image{N}et: A
  large-scale hierarchical image database,'' in \emph{IEEE Conference on
  Computer Vision and Pattern Recognition}, 2009, pp. 248--255.

\bibitem{jia2014caffe}
Y.~Jia, E.~Shelhamer, J.~Donahue, S.~Karayev, J.~Long, R.~Girshick,
  S.~Guadarrama, and T.~Darrell, ``Caffe: Convolutional architecture for fast
  feature embedding,'' in \emph{Proceedings of the 22nd ACM international
  conference on Multimedia}, 2014, pp. 675--678.

\bibitem{kingma2014adam}
D.~P. Kingma and J.~Ba, ``Adam: A method for stochastic optimization,''
  \emph{arXiv preprint arXiv:1412.6980}, 2014.

\bibitem{santhanam2017generalized}
V.~Santhanam, V.~I. Morariu, and L.~S. Davis, ``Generalized deep image to image
  regression,'' in \emph{IEEE Conference on Computer Vision and Pattern
  Recognition}, 2017, pp. 5609--5619.

\bibitem{wang2017stagewise}
T.~Wang, A.~Borji, L.~Zhang, P.~Zhang, and H.~Lu, ``A stagewise refinement
  model for detecting salient objects in images,'' in \emph{IEEE International
  Conference on Computer Vision}, 2017, pp. 4019--4028.

\bibitem{zhang2017amulet}
P.~Zhang, D.~Wang, H.~Lu, H.~Wang, and X.~Ruan, ``Amulet: Aggregating
  multi-level convolutional features for salient object detection,'' in
  \emph{IEEE International Conference on Computer Vision}, 2017, pp. 202--211.

\bibitem{Zhao_2017_CVPR}
H.~Zhao, J.~Shi, X.~Qi, X.~Wang, and J.~Jia, ``Pyramid scene parsing network,''
  in \emph{IEEE Conference on Computer Vision and Pattern Recognition}, 2017,
  pp. 2881--2890.

\bibitem{vicente2016noisy}
T.~F.~Y. Vicente, M.~Hoai, and D.~Samaras, ``Noisy label recovery for shadow
  detection in unfamiliar domains,'' in \emph{IEEE Conference on Computer
  Vision and Pattern Recognition}, 2016, pp. 3783--3792.

\bibitem{he2016deep}
K.~He, X.~Zhang, S.~Ren, and J.~Sun, ``Deep residual learning for image
  recognition,'' in \emph{IEEE Conference on Computer Vision and Pattern
  Recognition}, 2016, pp. 770--778.

\bibitem{yang2012shadow}
Q.~Yang, K.-H. Tan, and N.~Ahuja, ``Shadow removal using bilateral filtering,''
  \emph{IEEE Transactions on Image Processing}, vol.~21, no.~10, pp.
  4361--4368, 2012.

\end{thebibliography}
}



%

\begin{IEEEbiography}[{\includegraphics[width=1in,height=1.25in,clip,keepaspectratio]{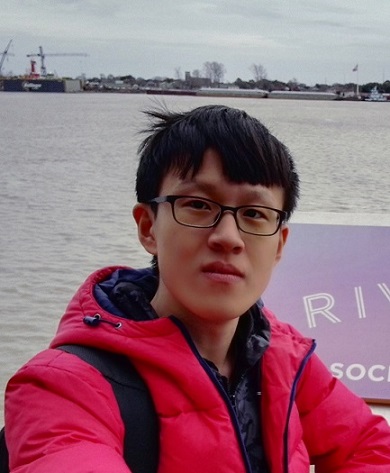}}]{Xiaowei Hu}
	
	received the B.Eng. degree in the Computer Science and Technology from South China University of Technology, China, in 2016. He is currently working toward the Ph.D. degree with the Department of Computer Science and Engineering, The Chinese University of Hong Kong. His research interests include computer vision and deep learning.
	
\end{IEEEbiography}


\begin{IEEEbiography}[{\includegraphics[width=1.0in,height=1.25in,clip,keepaspectratio]{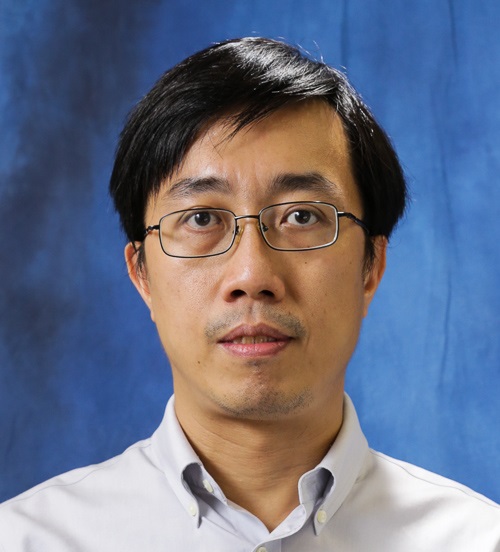}}]{Chi-Wing Fu} joined the Chinese University of Hong Kong as an associate professor from 2016.  He obtained his Ph.D. in Computer Science from Indiana University Bloomington, USA.  He served as the program co-chair of SIGGRAPH ASIA 2016 technical brief and poster, associate editor of Computer Graphics Forum, and program committee members in various conferences including IEEE Visualization.  His research interests include computer graphics, visualization, and user interaction.
\end{IEEEbiography}


\begin{IEEEbiography}[{\includegraphics[width=1in,height=1.25in,clip,keepaspectratio]{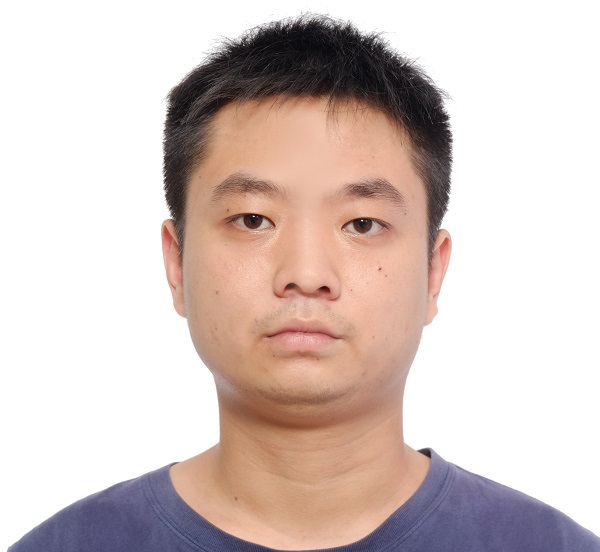}}]{Lei Zhu}
	received his Ph.D. degree in the Department of Computer Science and Engineering from the Chinese University of Hong Kong in 2017.
He is working as a postdoctoral fellow at the Chinese University of Hong Kong. His research interests include computer graphics, computer vision, medical image processing, and deep learning.
\end{IEEEbiography}


\begin{IEEEbiography}[{\includegraphics[width=1in,height=1.25in,clip,keepaspectratio]{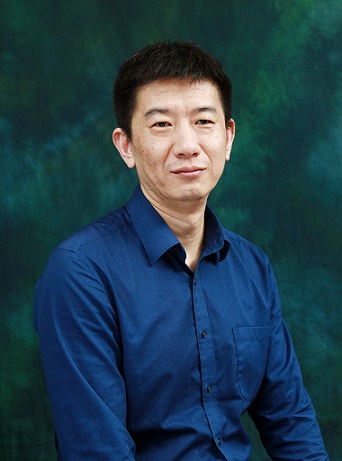}}]{Jing Qin} received his Ph.D. degree in Computer Science and Engineering from the Chinese University of Hong Kong in 2009. He is currently an assistant professor in School of Nursing, The Hong Kong Polytechnic University. He is also a key member in the Centre for Smart Health, SN, PolyU, HK. His research interests include innovations for healthcare and medicine applications, medical image processing, deep learning, visualization and human-computer interaction and health informatics.
\end{IEEEbiography}


\begin{IEEEbiography}[{\includegraphics[width=1in,height=1.25in,clip,keepaspectratio]{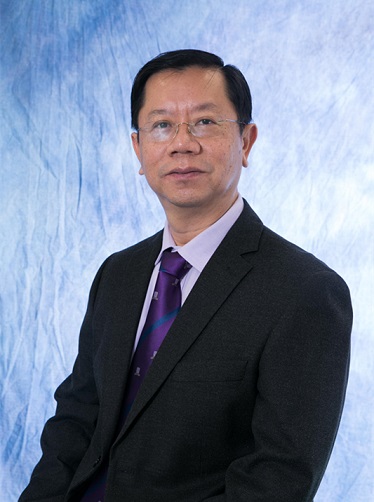}}]{Pheng-Ann Heng} received his B.Sc. (Computer Science) from the National University of Singapore in 1985. 
He received his M.Sc. (Computer Science), M. Art (Applied Math) and Ph.D. (Computer Science) all from the Indiana University in 1987, 1988, 1992 respectively.
He is a professor at the Department of Computer Science and Engineering at The Chinese University of Hong Kong. He has served as the Department Chairman from 2014 to 2017 and as the Head of Graduate Division from 2005 to 2008 and then again from 2011 to 2016.
He has served as the Director of Virtual Reality, Visualization and Imaging Research Center at CUHK since 1999. He has served as the Director of Center for Human-Computer Interaction at Shenzhen Institutes of Advanced Technology, Chinese Academy of Sciences since 2006. He has been appointed by China Ministry of Education as a Cheung Kong Scholar Chair Professor in 2007. 
His research interests include AI and VR for medical applications, surgical simulation, visualization, graphics and human-computer interaction.
\end{IEEEbiography}




\end{document}